%% file: MAIN/main.tex
\documentclass[letterpaper]{MAIN/nature}

\usepackage{color}
\usepackage{url}
\usepackage{verbatim}
\usepackage{multirow}
\usepackage{xspace}
\usepackage{graphicx}
\usepackage{amsmath} 
\usepackage{bbm}
\usepackage{amssymb}
\usepackage{booktabs} 
\usepackage{times} 
\usepackage{lscape}
\usepackage[left=1in,right=1in,bottom=1.1in,top=1in]{geometry}
\usepackage{enumitem}
\usepackage[table,xcdraw]{xcolor}
\usepackage[normalem]{ulem}
\usepackage{xcolor}
\usepackage[innercaption]{sidecap}
\usepackage{lineno}
\usepackage{stmaryrd}
\usepackage{xurl}
\usepackage[htt]{hyphenat}
\usepackage{makecell}
\usepackage{longtable} 
\usepackage{amsmath}   

\usepackage{float}
\usepackage{array}
\usepackage[most]{tcolorbox}

\definecolor{figblue}{HTML}{9EBCE3}
\definecolor{frameblue}{HTML}{2F5B8F}
\definecolor{backblue}{HTML}{EEF4FB}

\makeatletter
\newcommand*{\addFileDependency}[1]{
  \typeout{(#1)}
  \@addtofilelist{#1}
  \IfFileExists{#1}{}{\typeout{No file #1.}}
}
\newcommand{\xhdr}[1]{\vspace{1em}\noindent{{\bf #1}}}

\makeatother

\usepackage[colorlinks,citecolor=blue,urlcolor=magenta]{hyperref}

\usepackage[margin=-1cm, labelfont=bf, font=footnotesize]{caption}
\usepackage{algorithm}
\usepackage{algorithmic}
\usepackage{cleveref}
\newcommand{\hide}[1]{}

\let\oldnl\nl
\newcommand{\nonl}{\renewcommand{\nl}{\let\nl\oldnl}}  
\usepackage[normalem]{ulem}
\usepackage{comment}
\usepackage{bm}
\graphicspath{{../}{./}{../FIG/}{./FIG/}}

\title{\begin{center}A Living Benchmark for Information Retrieval from Electronic Health Records\end{center}}

\author  
{\begin{center}   
Jordan L. Cahoon$^{1,2}$, Chloe O. Stanwyck$^{1,3}$, Sulaiman Somani$^{4}$, Philip Chung$^{3}$, Kevin R Keet$^{4}$, Kameron C. Black$^{4}$,  Andrea T. Fisher$^{5}$, Sarita Khemani$^{4}$, Jerry Liu$^{4}$, Stephen Ma$^{4}$, Saloni K. Maharaj$^{4}$, Rita M. Pandya$^{4}$, Eduardo Perez-Guerrero$^{4}$, Priyanka Pillai$^{4}$, Lisa Shieh$^{4}$, David J.H. Wu$^{6}$, James Xie$^{3}$, James C. McAvoy$^{3}$, Teresa Nguyen$^{3}$, Jessica Tran$^{4}$, Lucy Yin$^{1}$, Bridget Lin$^{1}$, Alison Callahan$^{4}$, Jason A. Fries $^{1,4,7}$, Nigam H. Shah $^{1,4,8}$, and Emily Alsentzer$^{1,7,9,\ddag}$ \\[1mm]  
\small{$^{1}$Department of Biomedical Data Science, Stanford University, Stanford, CA} \\
\small{$^{2}$Department of Pathology, Stanford University, Stanford,
CA} \\
\small{$^{3}$Department of Anesthesiology, Perioperative and Pain Medicine, Stanford University, Stanford, CA} \\
\small{$^{4}$Department of Medicine, Stanford University, Stanford, CA} \\
\small{$^{5}$Department of Surgery, Stanford University, Stanford, CA}\\
\small{$^{6}$Department of Radiation Oncology, Stanford Cancer Center, Palo Alto, CA, USA} \\
\small{$^{7}$Weill Cancer Hub West} \\
\small{$^{8}$Center for Clinical Excellence Research, Stanford School of Medicine, Stanford, CA, USA}\\
\small{$^{9}$Department of Computer Science, Stanford University, Stanford, CA} \\
\small{$\ddag$Corresponding author. Email: ealsentzer@stanford.edu}\\
\end{center}
}

\begin{document}

\maketitle

{\spacing{1.4}

\section*{Abstract}

\begin{abstract}
 \input{MAIN/000abstract}
\end{abstract}
}

\clearpage

\spacing{1.38}

\input{MAIN/010intro}

\section*{Results}

\input{MAIN/020results}

\section*{Discussion}

\input{MAIN/030discuss}

\section*{Online Methods}

\input{MAIN/040methods}

\clearpage

\spacing{1.4}

\section*{Data availability}
All data were de-identified using a ``hiding in plain sight'' protocol
where protected health information (PHI) is replaced by coherent synthetic alternatives\cite{Carrell2013Hiding}. The dataset will be hosted on the university-approved, secure data portal, Redivis, under controlled access for clinical AI evaluation. Access will require signing a data usage agreement and completing CITI ethics trainings.
 
\section*{Code availability}
The complete code base used to generate the dataset and perform analyses is available at \url{https://github.com/alsentzerlab/brie}. 

\section*{Acknowledgments} 
 J.L.C is supported by the Warren Alpert Computational Biology \& Artificial Intelligence Fellowship. C.S. is supported by the National Institute of General Medical Sciences of the National Institutes of Health under award number T32GM089626. P.C. is supported by the Foundation for Anesthesia Education and Research. N.H.S acknowledges support from the Debra and Mark Leslie Endowment for AI in Healthcare as well as Stanford Healthcare's support for GUIDE-AI. A.T.F. is supported by T32HL166155 from the National Institutes of Health, Stanford Cardiovascular Institute, Computational Medicine in the Heart: Integrated Training Program. E.A. is supported by Weill Cancer Hub West, an Accenture HAI grant, and the Chan Zuckerberg Biohub. 
 
 Some of the computing for this project was performed on the Stanford Carina cluster. We would like to thank Stanford University and Stanford Research Computing for providing computational resources and support that contributed to these research results, as well as the Technology \& Digital Solutions Team for assisting with data access.

\section*{Competing interests} 
The authors declare the following competing interests: J.A.F. reports consulting fees from Snorkel AI. N.H.S. is a co-founder of Prealize Health and Atropos Health, serves on the board of directors of BrightSpring Health Services, and serves as an advisor to J\&J Innovative Medicines and AbbVie. E.A. reports consulting fees from Fourier Health. 

\section*{Authors contributions} 
C.O.S, S.S., P.C., K.R.K., K.C.B., A.T.F., S.K., J.L., S.M., S.K.M., R.M.P., E.P.G., P.P., L.S., P.C., and D.J.H.W. performed clinical review of the BRIE generator. C.O.S., J.X., J.C.A., and T.N. reviewed multiple answer generation. C.O.S, S.S., J.L.C., and B.L. reviewed the automated evaluation of LLM responses. J.L.C. and L.Y. ran experiments and performed analyses. J.L.C., J.A.F, and A.C. curated data. J.L.C., E.A., C.O.S., P.C, N.H.S, and J.A.F. aided in conceptualization and methods development. J.L.C. and E.A. prepared the original draft. All authors aided in review and edits.

\section*{References}

{
\spacing{0.85}

\bibliographystyle{MAIN/naturemag}
\bibliography{MAIN/refs}
}

\clearpage
\newgeometry{left=0.9in,right=0.9in,top=1in,bottom=1in}

\setcounter{figure}{0}
\setcounter{table}{0}
\renewcommand{\figurename}{Supplementary Figure}
\renewcommand{\tablename}{Supplementary Table}

\renewcommand{\theHfigure}{supp.\arabic{figure}}
\renewcommand{\theHtable}{supp.\arabic{table}}

\captionsetup{margin=0pt,labelfont=bf,font=footnotesize}
\setlength{\tabcolsep}{3pt}
\spacing{1}

\begin{center}
{\Large Supplementary Information}
\end{center}

\clearpage
\section*{Supplementary Notes}
\input{SI/SI-notes}

\clearpage
\section*{Supplementary Figures and Tables}
\input{SI/SI-figures}

\clearpage 

\end{document}

%% file: MAIN/000abstract.tex

\begin{abstract}
\normalfont
Large language model (LLM)-based clinical assistants are increasingly being integrated into electronic health record (EHR) systems, transforming how clinicians retrieve and synthesize information from patient records. Their safety and utility depend on rigorous evaluation, yet existing benchmarks are manually curated, costly to update, and rapidly become obsolete with evolving technological advancements. We present a scalable framework that automatically generates question--answer pairs from longitudinal EHR notes. Nineteen clinicians validate the benchmark generator, producing the Benchmark for Retrieving Information in EHRs (BRIE), a continuously maintainable evaluation dataset. Across nine LLMs and five inference strategies, state-of-the-art systems frequently omit clinically important information, particularly for questions requiring synthesis across multiple documents and encounters. Because the generator itself is validated, BRIE supports evaluations that static benchmarks cannot, including the generation of multiple answers that reflect variation in clinician reasoning for robust performance assessment and continuously refreshing benchmark content to guard against leakage. Our results demonstrate that scalable benchmark generation enables rigorous, up-to-date evaluation of clinical LLMs as they are deployed in rapidly evolving healthcare settings.
\end{abstract}

%% file: MAIN/010intro.tex

Large language models (LLMs) are rapidly transforming clinical workflows throughout hospitals. At a growing number of institutions, clinicians can now send requests about specific patient medical records to secure LLM deployments embedded within electronic health record (EHR) systems \cite{armitage_clinicians_2025,noauthor_chop_nodate,lynn_deep_2025,ucsf_ars_brim_2026, griot2025implementation,gao2026scout,shah2026adoption,openai2026healthcare}. These systems augment the chart review process by enabling clinicians to interact with patient records through a unified chat interface, shifting away from traditional search workflows. Through these interactions, clinicians may ask questions about a patient’s medical history, locate records from prior encounters, and generate summaries. Such use cases require models to navigate and synthesize information spread across long and complex patient records.

Monitoring these LLM deployments requires evaluation frameworks that can characterize model performance, identify failure modes, and reflect clinically realistic settings \cite{kanithi2024medic, ravichandran2025healthbench, artsi2025workflows, pan2025beyond}. However, new methodological advancements are introduced faster than traditional benchmarks can be created \cite{dsouza2025automating, ruder2024evolving, akhtar2026plateau}. As a result, despite growing interest in deployment, robust evaluation of clinical information retrieval remains an open challenge \cite{shool2025systematic,artsi2025challenges}. 
Reasoning across hundreds of clinical notes poses distinctive challenges for LLMs. Relevant information is often dispersed across multiple encounters, while clinical observations are frequently duplicated through copy-forward documentation \cite{hirschtick_copy_paste_2006,weis_copy_2014}. Further, notes may interleave findings from the current and past encounters, making it difficult to understanding when observations happened \cite{ebbers2022impact}. Models must extract meaningful signal from this noise, which can result in misinterpreting the record and missing important information \cite{cahoon2026clinicalnotebloatreduction}. Such errors can propagate into model responses and ultimately affect clinical care. As these LLM systems become increasingly embedded in clinical workflows, healthcare organizations must continuously evaluate their reliability. 

Recent work has enabled retrieval evaluation by creating benchmarks that include questions and answers based on individual patient notes. Often times, clinicians are asked to author these question and answers directly to ensure clinical utility \cite{fleming_medalign_2024}; however, the scale is limited by available annotations resources. Synthetic generation with LLMs enables a way to augment this process, where clinicians focus on validation rather than authoring from scratch \cite{kweon_ehrnoteqa_2024, cui_timer_2025}. Despite this progress, existing benchmarks do not fully capture breadth of information retrieval questions that are necessary to evaluate deployed chart review systems\cite{singhal2023llmclinical,yang2025ehrstruct}. More fundamentally, these benchmarks are treated as static artifacts that gradually lose effectiveness as their contents are absorbed into training corpora, documentation practices change overtime, and utilization evolves, such that measured performance may increasingly reflect memorization rather than robust clinical reasoning \cite{livebench,yan2026livemedbench, ronaghi2026clinicallygroundedprivacyevaluation}. Combined with the substantial clinician effort required to construct them, this makes existing paradigms insufficient for the robust evaluation of large language model deployments in chart review \cite{ahsan2024retrieving, nahum2025llms}.

Rather than asking clinicians to dedicate time to manually update the benchmark, we develop a scalable LLM-based generator to produce question--answer pairs from real, de-identified patient records. These pairs are structured along evaluation axes spanning reasoning, topics, and time to enable evaluation on specific clinical retrieval challenges. Nineteen physicians who regularly use these clinical LLM tools perform a one-time validation of the generator, supporting a ``living'' benchmark paradigm \cite{livebench}. After this initial validation, the generator can be rerun on new patient data to produce scalable, continuously updated benchmark datasets.

We leverage the scalable generator to create the Benchmark for Retrieving Information in EHRs (BRIE), a dataset containing over five hundred clinical information retrieval questions with verified answers and de-identified clinical notes. BRIE enables both the evaluation \emph{and} continuous monitoring of deployed clinical retrieval systems. Using BRIE, we identify failure modes across nine state-of-the-art models \cite{comanici2025gemini25,google2026gemini25family,openai2026gpt54blog,openai2026gpt54nanoblog,singh2026openaigpt5card,anthropic2025claudehaiku45blog,anthropic2026claudeopus47blog,qwen3.5,kimiTeam2025kimik2,moonshot2026kimik26} and five inference configurations \cite{robertson1994bm25}, including retrieval-augmented generation (RAG) \cite{lewis2020retrieval,liu_lost_2023,lopez2025clinical} and agentic approaches \cite{zhang2024agentic, qu2026trace}. We find that while hallucinations are rare, models frequently omit relevant facts, particularly when the answer is dispersed across multiple parts of the clinical record.

We demonstrate that our evaluation framework enables robust analyses that fixed benchmarks cannot easily support. Using the validated generator, we create multiple reference answers that represent different clinical interpretations and find that single-reference evaluation systematically underestimates model capability. Additionally, to show that BRIE can easily be updated, we fully regenerate BRIE from admissions collected two years later and show that it retains its difficulty without clinician intervention. Altogether, these contributions demonstrate how BRIE enables continuous, robust clinical retrieval evaluation.

%% file: MAIN/020results.tex
\subsection*{A scalable framework for generating clinical retrieval benchmarks}

\begin{figure}
    \hspace*{-.75cm}
    \includegraphics{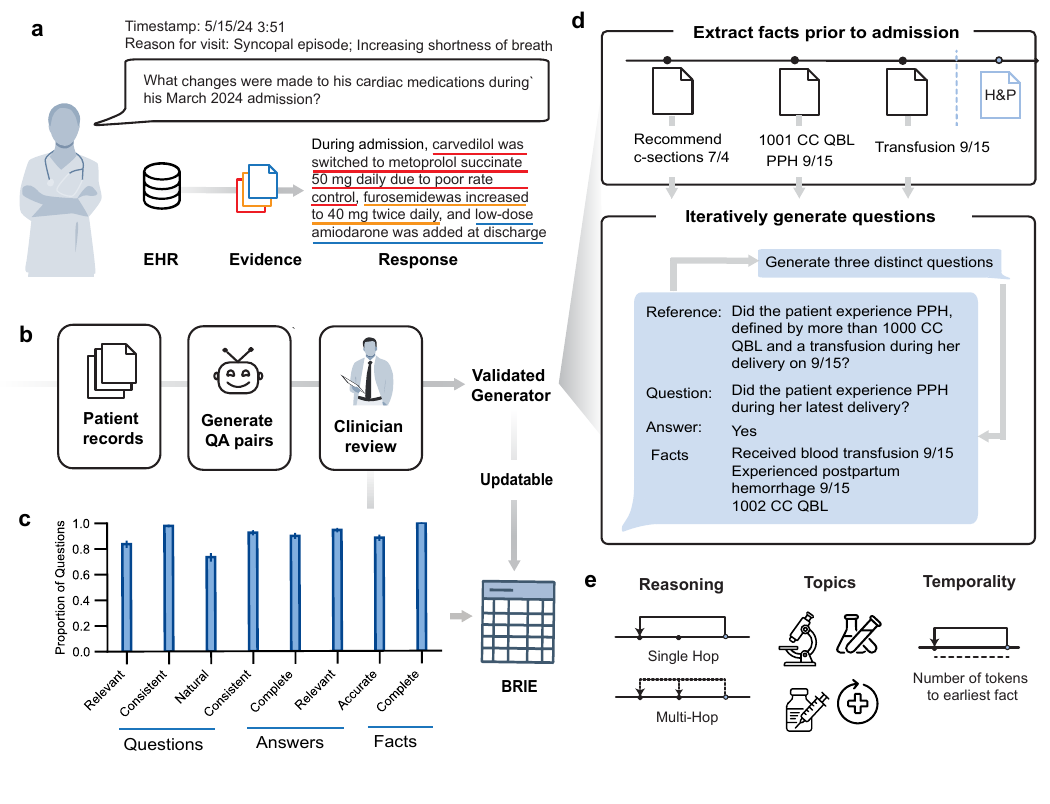}
    \caption{\textbf{A living benchmark for retrieval over EHRs}. (a) We formulate clinical information retrieval as a question--answering task  for a specific inpatient encounter. Answering each question requires identifying the relevant supporting sources and synthesizing a free-text response supported by those sources. (b) To construct a living benchmark, we generate question--answer pairs from de-identified clinical notes and have clinicians evaluate their quality to validate the benchmark generator. Once validated, the generator can be applied to new clinical notes to create updated evaluation datasets over time. (c) Generated question–answer pairs are of high quality. The the x axis denotes the criterion, spanning question, answer, and fact dimensions, and the y axis shows the proportion of the 675 question–answer pairs meeting each quality criterion. Error bars indicate 95\% confidence intervals. (d) Question-–answer generation comprises a fact-extraction stage, in which facts are drawn directly from longitudinal clinical notes, and a question-generation stage. (e) We define a clinical retrieval taxonomy along three axes: the number of reasoning hops, clinical topic, and temporality, measured by the number of tokens to the earliest supporting fact.}.
    \label{fig:overview}
\end{figure}

We focus evaluation on a common chart review task in which a clinician must quickly understand a patient's medical history when admitting a patient to the hospital (Figure \ref{fig:overview}a). We formalize this task as answering a clinical question for a specific patient at a specific point in time, using all information available in the medical record up to that point---here, when the admission History \& Physical (H\&P) note was written. The output is a free-text response supported by evidence from the patient's prior clinical documentation  (Figure \ref{fig:overview}a). To enable evaluation at scale, clinicians verify the question--answer pairs automatically generated from de-identified clinical notes (Figure \ref{fig:overview}b). To ensure that these pairs reflect questions a clinician might ask when admitting a patient, we ground their generation in the patient’s History \& Physical (H\&P) note, which captures the clinical information considered relevant at admission (Figure \ref{fig:overview}d).

Generation proceeds in two stages (Figure \ref{fig:overview}d). First, an LLM extracts and de-duplicates candidate facts from all clinical notes documented before the H\&P, removing redundancy introduced by copied or imported text. Next, the extracted facts and the H\&P note are provided jointly to an LLM (Gemini Pro 2.5) in a single prompt to generate question--answer pairs, where answers are directly linked to a subset of patient facts. The H\&P provides encounter-specific clinical context, while the extracted facts define the permissible evidence for each answer. Consequently, every answer must be supported exclusively by the extracted facts and not by the H\&P itself, ensuring that the questions evaluate retrieval from the prior medical record rather than information explicitly summarized in the H\&P.

To systematically identify where existing retrieval approaches succeed and fail, we organize question-answer generation around an evaluation-oriented taxonomy whose axes correspond to distinct retrieval challenges (Figure \ref{fig:overview}e). Each question is generated conditioned on two axes: Reasoning, which distinguishes questions answerable from a single source (single-hop) from those that require combining evidence across multiple sources (multi-hop) \cite{yang2018hotpotqa}, and Topics, the clinical concept(s) targeted by the question (Figure \ref{fig:overview}e). A third axis, Temporality, captures where the supporting evidence resides in the longitudinal record and is determined after question generation (see Online Methods Section \ref{meth:temporality}). 

Using the generator, we produced 675 chart-review questions from 63{,}878 de-identified clinical notes spanning a representative sample of 68 patients at Stanford Health Care (median 400 notes per patient, range 102--9{,}319; Supplementary Figure \ref{sfig:patient-note} and Supplementary Table \ref{stab:demographics}). 

\subsection*{Clinicians validate benchmark generator for clinical relevance and accuracy}

To assess the quality of the automatically generated question--answer pairs, we recruited fifteen physicians, with two independent reviewers assigned to each pair. Reviewers rigorously evaluated each generated entry by separately assessing the question, answer, and fact set across 14 criteria for clinical utility and accuracy. Overall, 83.7\% of questions (95\% CI, 80.9--86.3) were judged clinically relevant and 97.8\% (95\% CI, 96.6--98.8) were consistent with the patient chart (Figure \ref{fig:overview}c). Natural phrasing was the primary area for improvement: 73.3\% of questions (95\% CI, 70.1--76.6) had realistic phrasing (Figure \ref{fig:overview}c), and reviewers supplied alternative phrasing where appropriate. Generated answers were of similarly high quality, with 92.0\% (95\% CI, 90.2--94.2) judged consistent with the chart, 89.8\% (95\% CI, 87.4--92.0) complete, and 94.2\% (95\% CI, 92.4--95.9) relevant (Figure \ref{fig:overview}c). Supporting fact lists were comparably strong, with 88.0\% (95\% CI, 85.6--90.5) supporting the corresponding answer and 99.7\% (95\% CI, 99.3--100.0) judged complete (Figure \ref{fig:overview}c).


\begin{figure}
    \centering
    \includegraphics[width=\linewidth]{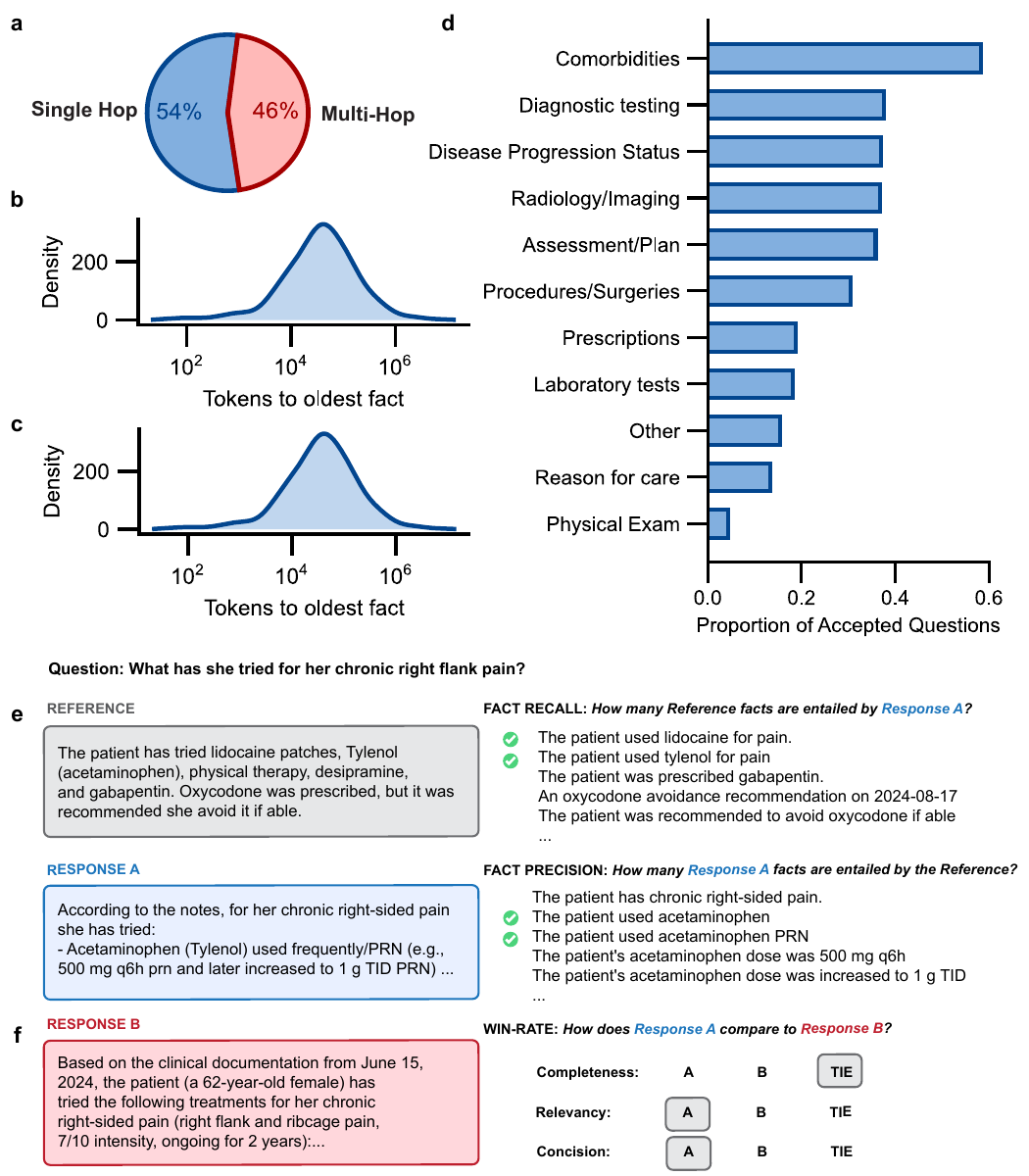}
    \caption{\textbf{Benchmark for Retrieval in EHRs (BRIE)}. BRIE represents a variety of retrieval settings, including (a) different reasoning types (single- and multi-hop), timeframes as measured by (b) the number of tokens between the query and the earliest supporting fact and (c) the number of tokens number of tokens between the earliest and latest supporting facts, and (d) question topics. We evaluate model responses using two metrics, illustrated with an example. (e) Fact recall is the proportion of facts in the Reference answer that appear in Response A, and fact precision is the proportion of facts in Response A that are supported by the Reference. (f) To compare relative performance, two responses are assessed in terms of completeness, relevancy and concision.}
    \label{fig:dataset}
\end{figure}

\subsection*{BRIE: benchmark for retrieving information in electronic health records}

Applying our scalable generation framework, we created BRIE, a benchmark designed to probe the challenges of retrieving information across longitudinal electronic health records. To establish an initial high-confidence reference dataset, disagreements about clinical relevance were adjudicated by a third physician. We retained question–answer pairs judged clinically relevant and for which at least one reviewer identified no missing, hallucinated, or irrelevant information, yielding 508 questions (Supplementary Table \ref{stab:filtering}).

BRIE represents diverse retrieval settings found in clinical practice. 276 questions require single-hop reasoning, and 232 synthesize information across different encounters through multi-hop reasoning (Figure \ref{fig:dataset}a; Supplementary Table \ref{tab:question-type-counts}). Answers contain a median of 5 supporting facts (range, 1--21). To account for copy-forward documentation, only the most recent mention of each supporting fact was retained. Using these mentions, the earliest evidence supporting an answer occurs a median of $3.8 \times 10^{4}$ tokens (range, $0$--$7.2 \times 10^{6}$) before the clinical query time, and the supporting facts span a median of $2.3 \times 10^{4}$ tokens (range, $0$--$7.0 \times 10^{6}$; Figure \ref{fig:dataset}b,c). Notably, for 63 questions, the earliest supporting evidence occurs more than $180{,}000$ tokens before the query time, requiring retrieval across extensive longitudinal portions of the medical record rather than localized chart review (Supplementary Table \ref{tab:question-type-counts}).

BRIE covers a broad range of clinical topics, with individual questions often spanning multiple categories. The most common topics include comorbidities (N=296), diagnostic testing (N=276), disease progression or status (N=188), radiology and imaging (N=187), assessment and plan (N=182), procedures and surgeries (N=155), prescriptions (N=95), laboratory tests (N=92), and reasons for care (N=68) (Figure \ref{fig:dataset}d; Supplementary Table \ref{tab:question-type-counts}).

We used BRIE to evaluate question-answering performance across nine models spanning proprietary frontier systems and open-weight models of varying sizes. Each model was evaluated under five inference configurations. We first considered two long-context baselines: \textit{Recent}, which fills the context window with the most recent clinical notes, and \textit{Recent-180K}, which includes up to a maximum of 180{,}000 tokens, a cut off selected to fall well within the smallest available context window. Because patient records often far exceed even the largest available context windows, and processing such long contexts can be prohibitively expensive at scale, we also evaluated retrieval-based alternatives. Specifically, we considered two retrieval-augmented generation (RAG) methods: \textit{BM25}, a sparse retriever that ranks passages by lexical overlap with the question, and \textit{Dense}, which ranks document chunks by semantic similarity. Finally, we evaluated an \textit{Agentic} configuration that performs LLM-guided retrieval through iterative tool use with conditional stopping.

The large number of model and retrieval configurations makes exhaustive manual evaluation infeasible. We therefore developed and validated two automated LLM-as-a-judge methods\cite{Bedi2026}. The first uses fact entailment to compare each model response against the clinician-validated reference answer (Figure \ref{fig:dataset}e). Fact recall measures the fraction of reference facts that the model response surfaces, and fact precision measures the fraction of facts in the model response that appear in the reference answer. Because two responses with similar fact recall and precision can still differ in clinical utility, we add a complementary win-rate metric that directly compares two responses for completeness (coverage of reference facts), relevance (focus on question-relevant information), and concision (conveys the relevant information concisely) (Figure \ref{fig:dataset}f). Further details and validation procedures are provided in Online Methods Section \ref{meth:inference}.

\subsection*{State-of-the-art models frequently omit clinically relevant information}

\begin{figure}
    \vspace*{-1cm}
    \centering
    \includegraphics[width=\linewidth]{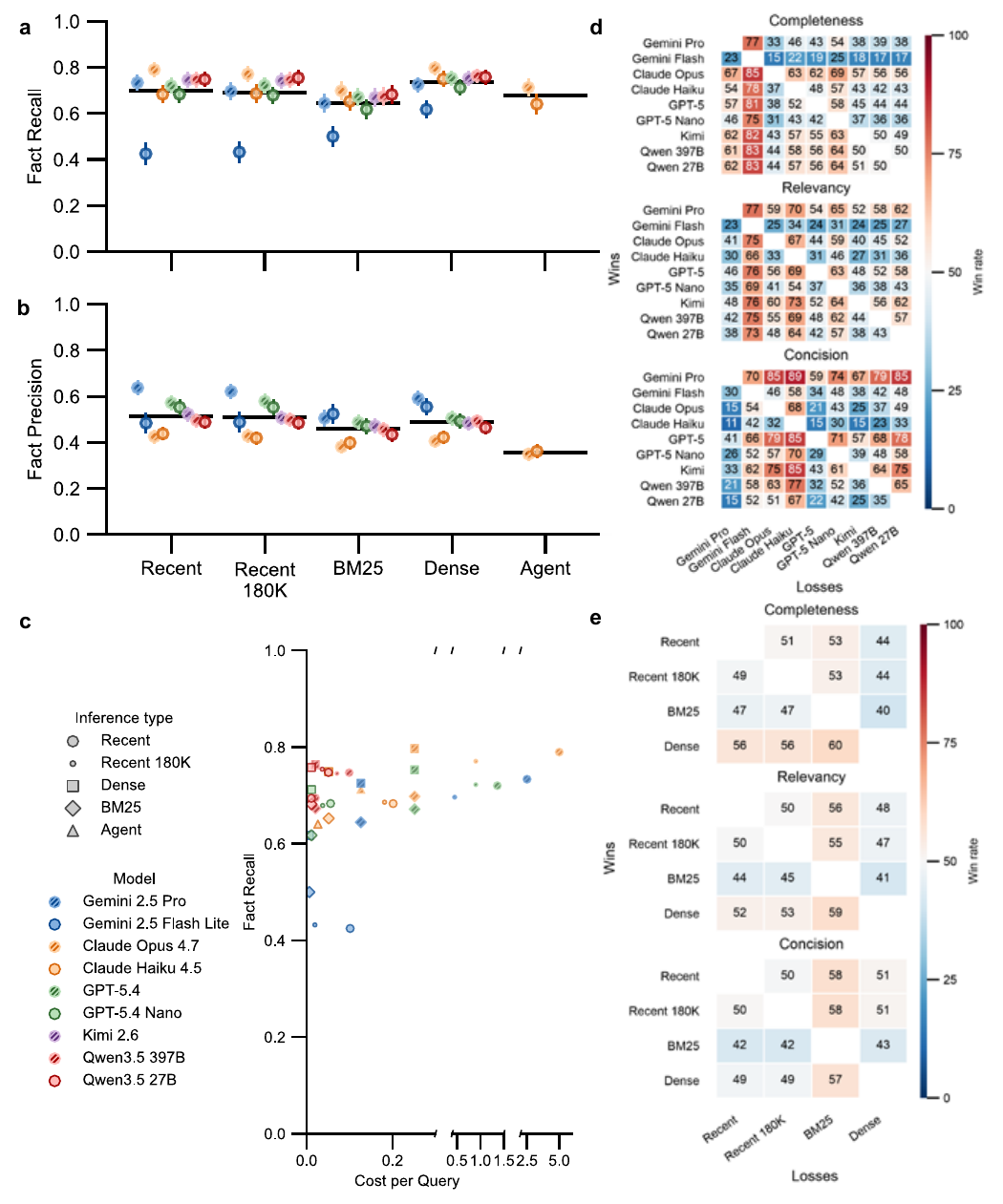}
    \caption{\textbf{Performance varies across model and inference configurations}. Average (a) fact recall and (b) fact precision are reported for each model--inference combination. Marker shape denotes the inference type, color denotes the model, and solid marker outlines identify the smaller model within each family. Error bars represent max-$t$ adjusted 95\% confidence intervals around the mean. (c) Average fact recall is plotted against inference cost for each model--inference combination. (d-e) Pairwise win rates summarize relative performance in terms of completeness, relevance, and concision: (d) each cell shows the percentage of wins of the y-axis model over the x-axis model, pooled across inference configurations, and (e) each cell shows the percentage of wins of the y-axis inference type over the x-axis inference type, pooled across models.}
    \label{fig:performance}
\end{figure}

We evaluate BRIE across model and inference configuration types to surface trends in state of the art information retrieval. Fact recall was moderate across all configurations ranging from 0.28 to 0.78 (Figure \ref{fig:performance}a; Supplementary Table \ref{tab:fact-recall-bootstrap-model-inference}). Even the strongest model-inference combination left roughly a quarter of the clinician-verified supporting facts unsurfaced. Under \textit{Recent} inference, frontier models achieved the highest average recall: Claude Opus 4.7 (0.78; 95\% CI, 0.76–0.82), Gemini 2.5 Pro (0.73; 95\% CI, 0.70–0.77) and GPT 5.4 (0.72; 95\% CI, 0.79–0.76) (Figure \ref{fig:performance}a; Supplementary Table \ref{tab:fact-recall-bootstrap-model-inference}).

Average fact recall fell among proprietary, cost-efficient models, most sharply for Gemini 2.5 Flash Lite (0.43; 95\% CI, 0380–0.47), with more modest declines for Claude Haiku 4.5 (0.68; 95\% CI, 0.64–0.72) and GPT 5.4 Nano (0.68; 95\% CI, 0.5–0.72). Open-weight models remained competitive with the frontier models despite their smaller scale, with Kimi K2.6, Qwen 3.5 397B, and Qwen 3.5 27B each reaching approximately 0.74 recall (95\% CI, 0.71-0.78, 0.71-0.78, 0.73-0.78) (Figure \ref{fig:performance}a; Supplementary Table \ref{tab:fact-recall-bootstrap-model-inference}).

Fact precision was consistently lower than fact recall, ranging from 0.17 to 0.63, indicating that model responses were more verbose than the reference response and included information beyond what the question required (Figure \ref{fig:performance}b; Supplementary Table \ref{tab:fact-precision-bootstrap-model-inference}). This tendency was most pronounced for the Claude models under \textit{Recent} inference (Claude Opus 4.7, 0.43; 95\% CI, 0.40–0.46; Claude Haiku 4.5, 0.44; 95\% CI, 0.41–0.47; ; Supplementary Table \ref{tab:fact-precision-bootstrap-model-inference}), reflecting a preference for more comprehensive responses at the expense of concision. Consistent with these results, pairwise preference evaluation most frequently favored Claude Opus 4.7 for completeness and relevance, whereas Gemini 2.5 Pro was most frequently favored for concision (Figure \ref{fig:performance}d).

Because facts absent from the reference answer may nonetheless be supported by the clinical record, we assessed whether the additional content reflected hallucination.  In a random sample of 50 queries spanning models and configurations, 99.2\% of extracted model response facts (95\% CI, 98.5–99.6) were faithful to the clinical record. Thus, the lower fact precision primarily reflected additional chart-grounded content rather than fabrication (Supplementary Figure \ref{sfig:hallucination} and Supplementary Table \ref{stab:hallucination}). These findings indicate that the dominant failure mode is therefore omission, not hallucination. 

\subsection*{Retrieval augmented generation matches long-context recall at lower cost}

Retrieval methods such as retrieval augmented generation (RAG) and agentic systems show promise to improve information retrieval and lower deployment costs, in exchange for added implementation overhead. We evaluated responses across inference configurations to assess whether these systems should be deployed in practice. Across individual models, \textit{Dense} retrieval matched or exceeded \textit{Recent} inference on fact recall while processing an average of 329{,}697 (69\%) fewer tokens, and it consistently outperformed keyword-based \textit{BM25} retrieval (Figure 3a–b, Appendix). The gains were most pronounced for cost-efficient models, where dense retrieval raised average fact recall by 0.18, 0.06, and 0.03 for Gemini 2.5 Flash Lite, Claude Haiku 4.5, and GPT 5.4 Nano, respectively, relative to recent-context inference (Figure 3a). \textit{BM25} retrieval, by contrast, yielded inconsistent results, changing fact recall by +0.07, –0.03, and –0.07, respectively, for the same models (Figure \ref{fig:performance}a).

\textit{Agentic} retrieval has been proposed as a promising approach for chart review, but did not improve fact recall over \textit{Dense} retrieval. For Claude Opus 4.7 and Claude Haiku 4.5, it even underperformed \textit{Recent} inference, reducing fact recall by 0.08 and 0.04, respectively (Figure \ref{fig:performance}a).

Pairwise comparisons reinforced the advantage of \textit{Dense} retrieval. Across models, responses produced with \textit{Dense} retrieval were preferred in 61\% and 56\% of comparisons for completeness and relevance, respectively. They were preferred in 47\% of comparisons for concision, indicating comparable brevity (Figure \ref{fig:performance}e).  \textit{Dense} retrieval also sharply reduced inference cost. Limiting retrieval to the top 50 documents holds the input context near 50K tokens per patient, producing a nearly constant cost across the cohort. Because 65 of the 68 BRIE patients carry more than 50K tokens of source notes (Supplementary Figure \ref{sfig:patient-token}), this cap would reduce the inference cost of processing the full benchmark by approximately \$837.43, or 86\%, at Claude Opus 4.7 pricing, assuming a fixed response length (Figure \ref{fig:performance}c, Supplementary Table \ref{stab:inference-cost}).

Beyond reducing inference cost, \textit{Dense} retrieval can surface question-relevant documents that may otherwise remain buried in the record. Several illustrative cases emerged when Claude Opus 4.7 was equipped with \textit{Dense} retrieval. For a patient presenting with cough and dyspnea, \textit{Dense} retrieval surfaced the most recent pulmonary function test, a 2019 study documenting severe airflow obstruction and substantial decline, which \textit{Recent} inference missed entirely. In another case, when asked for the most recent bone marrow biopsy in a patient with a history of myelofibrosis, \textit{Dense} retrieval accurately recovered the most recent report that documented normal bone marrow function, helping shift the diagnostic focus away from potential relapse. In contrast, \textit{Recent} inference returned contradictory claims about whether the biopsy existed. Finally, when identifying details of the most recent urinary tract infection in a patient presenting with pyelonephritis, \textit{Dense} retrieval identified the \textit{Klebsiella aerogenes} susceptibility profile and prior treatment course with ertapenem followed by ciprofloxacin, whereas \textit{Recent} conflated separate admissions and missed the culture report, omitting the causative organism and its susceptibilities entirely. Loss of such details can distort diagnostic reasoning and lead to ineffective or potentially harmful treatment. In each example, the relevant evidence was present in the context but difficult to identify because of duplicated text and temporal ambiguity. By filtering the record to the most relevant documents, \textit{Dense} retrieval can preserve or improve recall while substantially reducing the computational cost of clinical question answering.


\subsection*{Performance declines substantially for multi-hop and longitudinal retrieval}

The BRIE taxonomy enables us to decompose aggregate fact recall across reasoning complexity, clinical topic, and temporal distance, revealing failure modes obscured by aggregated performance. 

\begin{figure}
    \centering
    \includegraphics{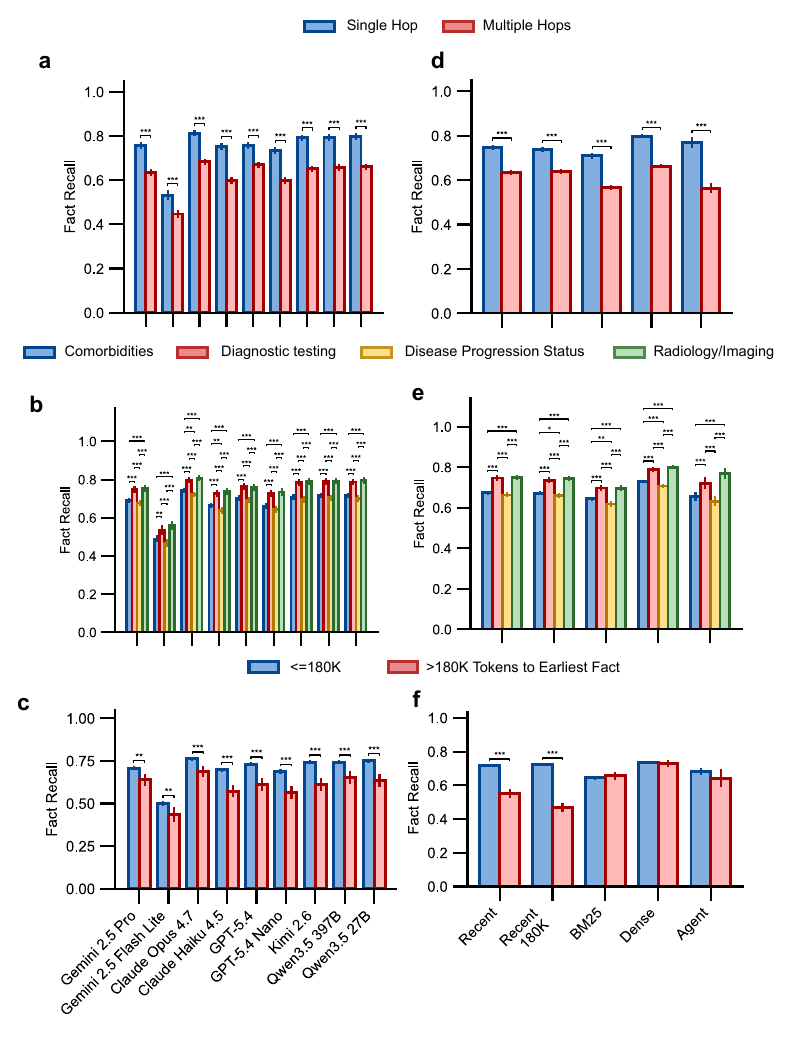}
    \caption{\textbf{State of the art models exhibit failure modes across reasoning, topics, and temporality.} Fact recall is aggregated across inference configurations and stratified by model for (a) reasoning complexity, (b) the four most frequent topics, and (c) answers with supporting facts that occur more than 180K tokens into the patient record. (d–f) Fact recall aggregated across models by inference type for the same question categories. Error bars represent the 95\% confidence intervals around the mean. Differences between groups were assessed using two-sided Mann--Whitney U tests, with Benjamini--Hochberg correction for multiple comparisons. We use the following significance thresholds: *** p $<$ 0.001; ** p $<$ 0.01; * p $<$ 0.05; ns = not significant (p $\geq$ 0.05).}
    \label{fig:failure}
\end{figure}

Reasoning complexity had the largest effect on fact recall (Figure \ref{fig:failure}a,d). Across all models and inference strategies, multi-hop questions consistently yielded lower fact recall than single-hop questions. These questions require synthesizing across multiple encounters and subsequently may cause the model to omit clinically relevant details. In one example, the question asked about prior rashes and treatments in an allogenic stem cell transplant patient presenting with worsening facial rash concerning for disseminated varicella zoster. \textit{Recent} inference by GPT-5.4, Claude Opus 4.7, and Gemini Pro 2.5 omitted key information about the effectiveness of prior treatments, including improvement with clotrimazole and metronidazole. All three models also omitted the dose-dependent relationship between ponatinib and the patient’s chronic rash, while GPT-5.4 omitted the ponatinib association entirely. These omissions obscured clinically relevant distinctions between the patient’s chronic and acute symptoms.

Multi-hop questions require reasoning about which information is relevant for the clinical context. Notably, alternative retrieval approaches exacerbated rather than reduced omission.
The average decrease in fact recall (multi-hop minus single-hop) was –0.20 for \textit{Agentic}, –0.14 for \textit{BM25}, and –0.13 for \textit{Dense} retrieval, compared with –0.12 for \textit{Recent} and –0.10 for \textit{Recent-180K} (Supplementary Figures \ref{sfig:recall-precision-reasoning-recent}-\ref{sfig:recall-precision-reasoning-agent}). Thus, although retrieval reduced inference cost and performed well overall, existing retrieval strategies did not consistently surface all evidence required for questions requiring reasoning across multiple encounters.

Fact recall also varied by clinical topic (Figure \ref{fig:failure}b,e). Under \textit{Recent} inference, recall was lower for questions about comorbidities (mean recall, 0.68; 95\% CI: 0.67-0.69) and disease progression (mean recall, 0.67; 95\% CI: 0.66-0.68) than for questions about imaging (0.76; 0.75–0.77) and diagnostic testing (0.76; 0.74–0.77) (Supplementary Figures \ref{sfig:recall-precision-topic-recent}-\ref{sfig:recall-precision-topic-agent}). Because comorbidities and disease progression may be documented across multiple encounters and summarized in many notes, models may struggle to distinguish and identify separate events. For example, when asked to characterize the pancreatitis history of a patient presenting with epigastric discomfort, \textit{Recent} inference with Gemini-Pro 2.5 identified prior alcohol-associated episodes in December 2020 and May 2022 but omitted more recent admissions in June and August 2022 during which lipase levels were normal. By recalling only the confirmed episodes, the response could mislead the differential diagnosis. This pattern persists when the analysis is restricted to multi-hop questions, suggesting that trends in recall across topics is distinct from those in complex reasoning  (Supplementary Figure \ref{sfig:topic-multi}).

Finally, fact recall declined when supporting evidence occurred farther back in the patient record (Figure \ref{fig:failure}c,f). While most relevant facts are documented in the most recent 180{,}000 tokens, 63 questions require surfacing information beyond this cutoff(Supplementary Figure \ref{sfig:fact-token}). In many cases, these facts may not be accessible with \textit{Recent} inference because the information falls out of context. For example, when asked about the management plan for the patient’s initial onset atrial fibrillation, \textit{Recent} inference with GPT 5.4 was unable to identify the relevant historical note because it fell outside the context window. When the earliest supporting fact appeared more than 180K tokens before the query, fact recall fell by an average of -0.16 and -0.25 for \textit{Recent} and \textit{Recent-180K} across models, respectively, but no significant drop in performance was observed for the other inference methods (Supplementary Figures \ref{sfig:recall-precision-temporality-recent}-\ref{sfig:recall-precision-temporality-agent}).

Together, these findings show that retrieval failures are concentrated rather than uniform. Current systems struggle most when answers require integrating evidence across encounters, tracking clinical concepts that evolve over time, or recovering facts from distant portions of the record. These limitations persist across models and inference strategies.

\subsection*{Multiple reference answers improve robustness of retrieval evaluation}

\begin{figure}
    \vspace*{-1cm}
    \centering
    \includegraphics[width=\linewidth]{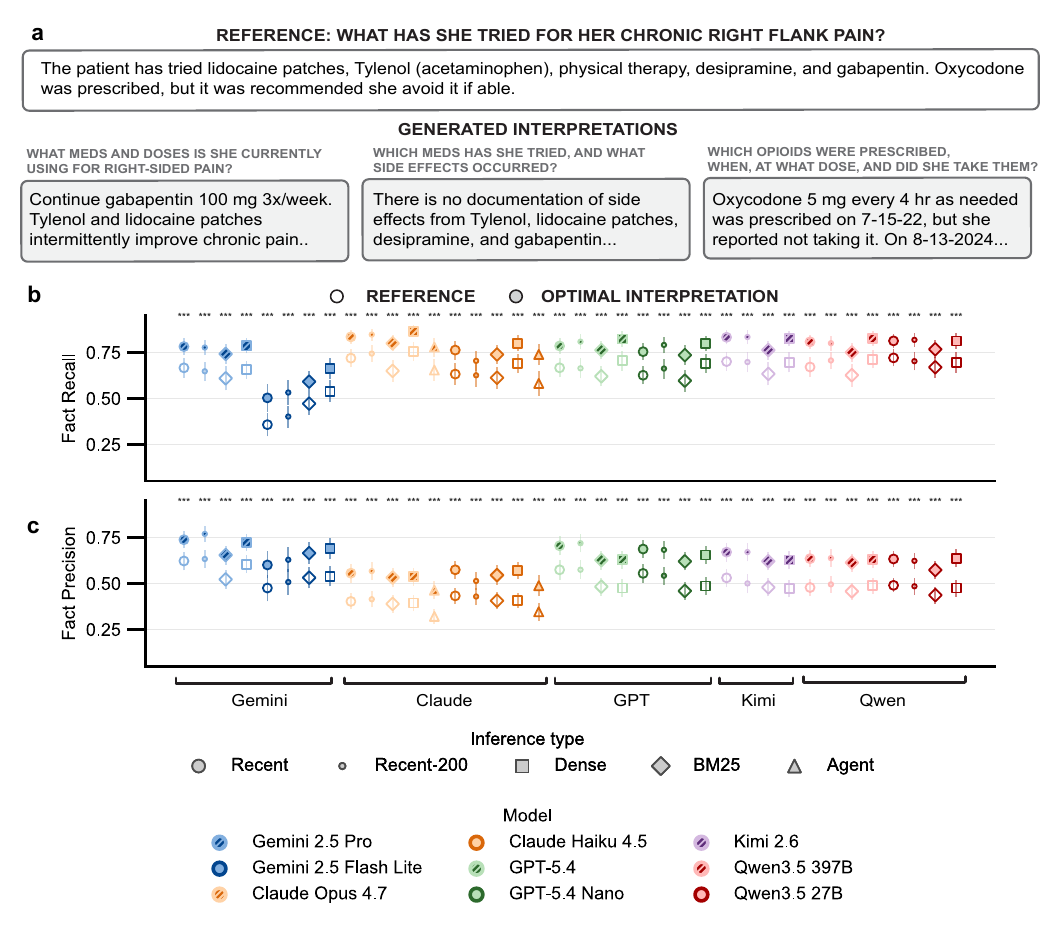}
    \caption{\textbf{Multiple generated interpretations of BRIE questions enable robust evaluation.} (a) Clinical information retrieval questions can have multiple valid answers based on the context. Using clinician approved interpretations for 87 questions, corresponding to 279 valid, generated interpretations  (b) fact recall and (c) fact precision is reported for each model--inference combination. Hollow points indicate scores obtained using the original reference answer, and filled points indicate the highest score obtained across all clinician-validated reference answers and alternative interpretations. The shape denotes the inference type, and the color denotes the model type, where markers with a solid outline are the smaller model in the family. Differences between reference and optimal interpretations were assessed using two-sided, paired Wilcoxan signed rank tests, with Benjamini--Hochberg correction for multiple comparisons. We use the following significance thresholds: *** p $<$ 0.001; ** p $<$ 0.01; * p $<$ 0.05; ns = not significant (p $\geq$ 0.05).}
    \label{fig:multiple}
\end{figure}

Standard benchmarks typically prescribe a single reference answer per question. However, in clinical settings, there may be multiple valid responses depending on context or clinician perspective (Figure \ref{fig:multiple}a). A response may therefore receive a low score because it does not align with one particular reference answer, even when it represents a valid clinical interpretation. As a result, evaluating models against a single reference answer may systematically underestimate true retrieval performance.  To quantify this effect, we sampled 100 BRIE questions and generated alternative interpretations of each, along with corresponding reference answers. Four board-certified physicians evaluated the resulting question–answer pairs for relevance and accuracy. An interpretation was considered valid if its answer also correctly addressed the original question and the modified question narrowed the scope by specifying additional details, such as the relevant time-frame, clinical focus, or desired response format.

Eighty-seven of the 100 sampled questions yielded at least one additional valid interpretation, with an average of 3.18 valid interpretations per question (range, 1–5). Consistent with our initial benchmark validation, the generated answers were of high quality and required minimal editing: 96.5\% (95\% CI, 94.1--98.6) of answers were judged accurate, 91.6\% (95\% CI, 88.5--94.8) complete, and 96.9\% (95\% CI, 94.8--98.6) relevant. We then treated the clinician-validated answers as alternative references and reevaluated the same model responses from model-inference configurations described in prior sections against all valid references.

Assuming the most optimistic evaluation scenario, in which the model response was scored against the interpretation that yielded the highest score (Figure \ref{fig:multiple}a), we found that fact recall increased by 0.12 (95\% CI: 0.12--0.13), corresponding to a relative improvement of 31.4\% (95\% CI: 28.9--33.93\%) (Figure \ref{fig:multiple}b). Fact precision increased by 0.14 (95\% CI: 0.14--0.15), corresponding to a relative improvement of 45.4\% (95\% CI: 42.7--48.0\%) across all model--inference combinations (Figure \ref{fig:multiple}c). Benjamini–Hochberg-corrected two-sided Wilcoxon signed-rank tests showed statistically significant differences using the most optimistic evaluation across model and inference configurations (Supplementary Tables \ref{tab:fact-recall-specification-wilcoxon} and \ref{tab:fact-precision-specification-wilcoxon}). When ranking all model and inference combinations, the top (e.g. Claude Opus 4.7 \emph{Recent} and \emph{Dense}) and bottom methods (e.g. inference with Gemini Flash Lite 2.5) are unchanged. However. optimistic interpretation can change the internal rankings, for example increasing Kimi with \emph{Recent-180K} inference from 10th to 4th highest recall. These results show that single-reference evaluation can substantially underestimate retrieval performance by penalizing responses aligned with alternative valid clinical interpretations. However, even when adjusting for clinician preferences and evaluating against the most optimistic interpretation, we observe that state-of-the-art retrieval approaches still have omission errors. 

\subsection*{BRIE can be automatically regenerated to mitigate contamination and drift}

A central goal of a living benchmark is to easily update the benchmark without clinician intervention. To assess whether the BRIE generation framework can produce updated evaluation cohorts without human-in-the-loop filtering, we compared three variants of BRIE: the clinician-validated $\text{BRIE}$ dataset (n=508), the unfiltered version prior to clinician review, $\text{BRIE}_{\text{unfiltered}}$ (n=675), and $\text{BRIE}_{\text{new}}$ (n=1{,}000), a dataset constructed from admissions in 2026, two years after the most recent admissions included in BRIE.

\begin{figure}
    \vspace{-.8cm}
    \centering
    \includegraphics[width=\linewidth]{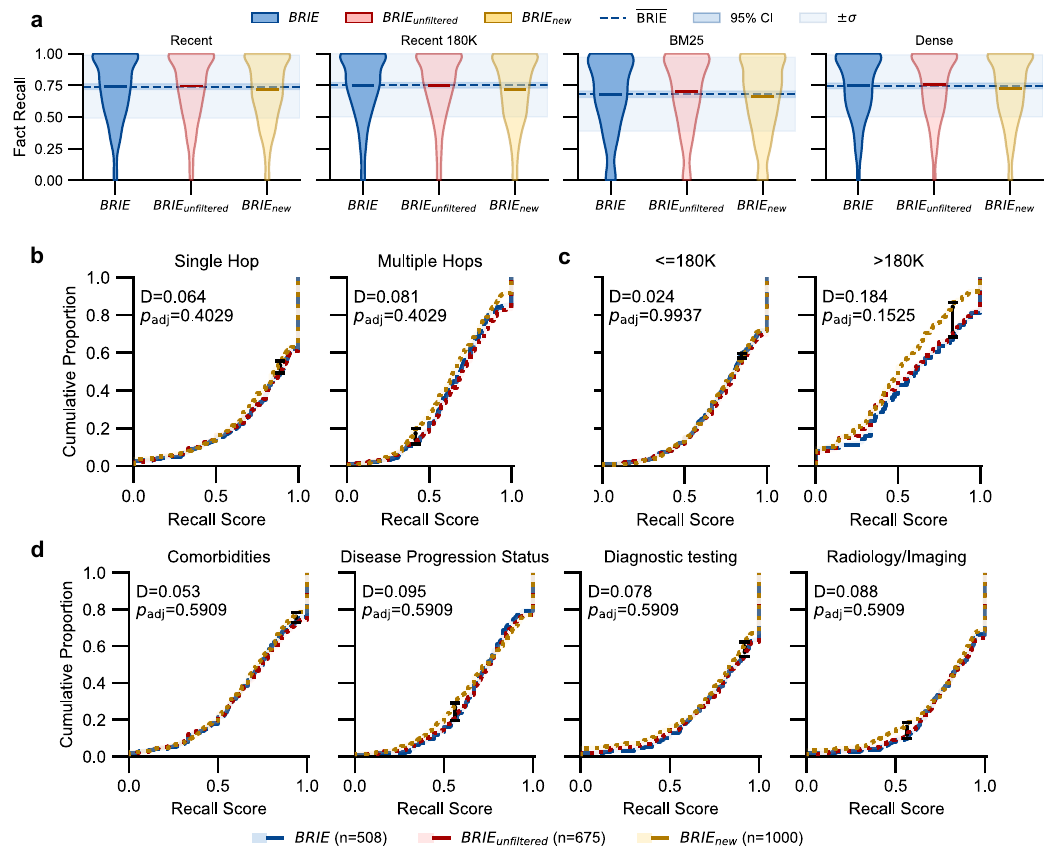}
    \caption{\textbf{BRIE can be automatically updated without compromising benchmark difficulty}. (a) Average fact recall across inference types for Qwen 27B for the BRIE (n=508, blue), $\text{BRIE}_\text{unfiltered}$ (n=675, red), and $\text{BRIE}_\text{new}$ (n=1000, yellow) datasets are reported. Solid horizontal lines denote cohort means, and the dashed line marks the BRIE mean. The widest light blue band indicates values one standard deviation away from the mean fact recall of BRIE whereas the darker band around the dashed line indicates the 95\% confidence interval. (b-d) Empirical cumulative distributions of fact recall are shown across (b) reasoning complexity, (c) temporal distance, and (d) clinical topic for all three cohorts. Brackets are drawn at the maximum distance in cumulative proportion between BRIE and the $\text{BRIE}_\text{new}$ Cohort. The Kolmogorov-Smirnov (KS) statistics are reported with significance thresholds as the following: *** p $<$ 0.001; ** p $<$ 0.01; * p $<$ 0.05; ns = not significant (p $\geq$ 0.05). }
    \label{fig:scale}
\end{figure}

We evaluated \textit{Recent}, \textit{Recent-180K}, \textit{BM25}, and \textit{Dense} retrieval with Qwen 3.5 27B across all three cohorts and found that performance was stable: mean fact recall was 0.73 (95\% CI: 0.72–-0.74), 0.73 (95\% CI: 0.73–-0.75), and 0.70 (95\% CI: 0.70-–0.71) for the BRIE, $\text{BRIE}_\text{unfiltered}$, and  $\text{BRIE}_{\text{new}}$ cohorts, respectively (Figure \ref{fig:scale}a). Corresponding fact precision values were 0.46 (95\% CI: 0.45-47), 0.46 (95\% CI: 0.45–0.46), and 0.42 (95\% CI, 0.41–0.43) (Figure \ref{fig:scale}a). One-sided Welch tests established non-inferiority within a 5-percentage-point margin for fact recall, indicating that neither cohort was meaningfully easier than BRIE (adjusted $p=8.97\times10^{-8}$ for $\mathrm{BRIE}{\mathrm{new}}$ and $p=2.89\times10^{-13}$ for $\mathrm{BRIE}{\mathrm{unfiltered}}$) (Supplementary Table \ref{tab:fact-recall-not-easier}). These findings suggest that automatically generated cohorts retain similar overall difficulty despite the absence of clinician filtering.

We next examined whether the regenerated datasets preserved the characteristic failure modes identified by the BRIE taxonomy.
Across reasoning complexity (Figure \ref{fig:scale}b), temporal characteristics (Figure \ref{fig:scale}c), and clinical topics (Figure \ref{fig:scale}d), the fact-recall distributions of  $\text{BRIE}_\text{unfiltered}$
and $\text{BRIE}_\text{new}$ did not differ significantly from those of BRIE. Using two-sample Kolmogorov--Smirnov tests, we did not detect significant distributional differences all comparisons (Supplementary Table \ref{tab:question-type-counts} and\ref{tab:fact-recall-specification-ks}). Importantly, the same failure modes, including reduced performance on multi-hops reasoning, topics that require longitudinal reasoning, and temporally distant evidence retrieval, persist across all cohorts. Together, these findings suggest that the benchmark generation framework preserves both overall task difficulty without human intervention and under temporal drift. 

%% file: MAIN/030discuss.tex
We present a scalable framework for building a living benchmark for information retrieval from longitudinal clinical notes. Rather than relying on clinicians to manually author benchmarks, our framework automatically generates chart-review questions grounded in real inpatient admissions, capturing the information needs that arise during routine clinical practice. To establish the validity of this approach, nineteen clinicians evaluated the generated questions and reference answers, judging them to be high quality and representative of real-world chart review. This single upfront annotation cost establishes the validity of the generator, enabling it to be rerun on newly collected clinical records to continuously refresh and expand the benchmark without requiring additional clinician annotation.

Using these validated annotations, we construct the Benchmark for Retrieving Information in EHRs (BRIE), which we release with verified reference answers, supporting evidence, and de-identified longitudinal clinical notes. BRIE consists of longitudinal records from medically complex patients, representing some of the most challenging information retrieval settings in clinical practice. We envision this resource serving as a foundation for developing and evaluating retrieval methods, while also supporting a broad range of clinical natural language processing research.

Using BRIE, we evaluated state-of-the-art retrieval across nine models and five inference configurations. Prior work has focused largely on model hallucinations as these errors can propagate into clinical practice and compromise care \cite{pandit2025medhallucomprehensivebenchmarkdetecting, asgari2025framework,shah2024accuracy}. We find that although model responses can be verbose, they contain few hallucinations. Instead, state-of-the-art retrieval frequently fails to include clinically relevant information. These omission errors are increasingly recognized as a key limitation of clinical AI systems \cite{niu2026aipatient}. Importantly, omissions pose a distinct challenge in clinical practice because they are difficult to detect. A response may appear plausible and accurate while silently excluding critical information that would only become apparent through a careful review of the patient chart. As a result, omission errors may be more dangerous than hallucinations, making them a critical target for future retrieval methods. 

These failures were not uniform across retrieval settings. Performance declined substantially when questions required synthesizing information across multiple encounters, tracking a condition over time, or recovering evidence buried deep in the record. The concentration of these failures in specific retrieval settings has implications for how clinical LLMs should be validated. Strong aggregate performance should not be interpreted as evidence that a system can reliably support ``EHR retrieval'' as a broad capability. Instead, evaluations should establish which retrieval settings a system can reliably support, distinguishing, for example, single-encounter retrieval from multi-encounter synthesis and longitudinal tracking of clinical conditions \cite{rajpurkar2025clinical,bressman2026software}.

Many clinical systems rely on long-context inference, where recent notes are concatenated and provided to the model (e.g., \textit{Recent}, \textit{Recent 180K}) \cite{wornow2025contextcluesevaluatinglong, chen2025buildingehrfoundationmodel}. Although straightforward to implement, this strategy can be computationally expensive for longitudinal records. Retrieval methods can reduce context length and inference cost, but their cost--performance tradeoffs in realistic clinical settings remain unclear \cite{griot2025implementation, lewis2020retrieval, ahsan2024retrieving}. We found that \textit{Dense} retrieval achieved performance comparable to long-context inference at substantially lower cost, suggesting that improving retrieval may provide greater practical benefit than simply expanding context windows. 

The scalable nature of our approach enables evaluations that are difficult to achieve with traditional benchmarks. First, because realistic questions rarely have a single valid answer, we developed a method to create multiple valid answers for a subset of BRIE queries \cite{taveekitworachai2026robustnessanswerformatsmedical, hosseini2024benchmarklongformmedicalquestion, saab2024capabilities}. Evaluating against these generated responses shows that single-reference scoring systematically underestimates model capability, motivating multi-reference evaluation for clinical information retrieval. Additionally, the results of this experiment provide a valuable asset for the community where the validated generator and annotations may be used to develop models that better align to different clinical contexts and user preferences.

Rather than treating BRIE as a static artifact, our framework supports continuous regeneration from newly collected records. Regenerated benchmarks preserve both overall difficulty and failure-mode structure, mitigating benchmark contamination while keeping evaluation aligned with evolving clinical practice \cite{li_r2med_2025, livebench}. By generating clinically relevant questions together with multiple validated answer interpretations at scale, the framework enables rigorous evaluation at scale. Because the generator can be reused, new versions of BRIE can be constructed from records collected in different years or at different institutions with minimal manual annotation. This can substantially reduce the cost of benchmark creation while enabling institution-specific evaluations that account for distribution shifts across diverse care settings.

This work has several limitations that point toward avenues for future research. First, BRIE demonstrates that realistic, longitudinal chart-review benchmarks can be generated automatically, and it exhibits encouraging temporal robustness, reproducing the benchmark on a cohort collected two years later. Establishing cohort generalizability, however, will require evaluation across institutions with differing documentation practices and patient populations. Second, BRIE is derived from a rich corpus of sixty thousand clinical notes that span long patient timelines, yet realism could be enhanced further by incorporating structured EHR data, including laboratory results, medications, and vital signs. Third, our evaluation covers five inference configurations, including an agentic retrieval approach that used keyword search. Future work should evaluate richer retrieval tools and reasoning strategies. Finally, the fact-based evaluation framework we introduce and validate measures information coverage without yet weighting facts by their clinical importance. We leave to future work the development of methods that weight facts by clinical importance to better align evaluation metrics with real-world clinical utility.

Together, our findings demonstrate that scalable, continuously updated benchmark generation can support rigorous evaluation of language models for clinical information retrieval. BRIE exposes systematic failures that aggregate evaluations can obscure, particularly the persistent omission of information distributed across longitudinal records. By re-framing evaluation as an ongoing process rather than a fixed artifact, BRIE provides a foundation for identifying when and why clinical retrieval systems fail as models, data, and deployment settings evolve.

%% file: MAIN/040methods.tex
The Online Methods  are organized as follows: (1) task and cohort definition; (2) benchmark generator; (3) clinician validation and BRIE curation; (4) model and inference configurations; (5) response evaluation and validation of the automated judges; (6) multiple-reference analysis; (7) benchmark regeneration analysis; and (8) statistical analyses.

\section{Task and Cohort Definition}
\label{meth:cohort}
We formulated clinical information retrieval as a question--answering task anchored to a specific inpatient admission. For each query, the reference time was the timestamp of the admission History \& Physical (H\&P) note and the retrieval corpus comprised clinical notes available before that time. The system was asked to produce a free-text answer supported by information in the prior patient record. The H\&P provided encounter-specific context for benchmark generation but was not itself a permissible source of evidence for the reference answer, ensuring that the task evaluated retrieval from the preceding longitudinal record.

\xhdr{Cohort Creation.} De-identified clinical notes were obtained from the STAnford Research Repository and included patients seen at Stanford Health Care and Lucile Packard Children's Hospital through August 25, 2024. Because this corpus is also used to develop foundation models and other artificial intelligence tools, patients were randomly sampled from the test split. Patients were eligible for inclusion if they had at least 100 notes before their most recent inpatient H\&P note and were not a pediatric patient. Supplementary Figure \ref{sfig:patient-note} shows the distribution of note counts, Supplementary Figure \ref{sfig:patient-token} shows the distribution of tokens across note timelines, and Supplementary Figure \ref{sfig:patient-daterange} shows the time elapsed between the index admission and oldest record across note timelines. Together, these demonstrate that our cohort has a sufficient longitudinal history for retrieval evaluation. We also used regular expressions to exclude H\&P notes matching any of the phrases ``surgical h\&p,'' ``preoperative history,'' ``pre-endoscopy procedure,'' ``pre-procedure,'' or ``surgery service consultation,'' limiting the overrepresentation of procedure-specific encounters while preserving a diverse set of admissions. In total, 75 patients were sampled. Cohort demographics are reported in Supplementary Table \ref{stab:demographics}.

\section{Benchmark Generator}
We developed a scalable framework to generate realistic questions that clinicians might ask when deciding to admit a patient. Using the index H\&P note for encounter-specific context, the pipeline jointly generates question–answer pairs grounded in facts extracted from the preceding notes. Specifically, generation is broken down in three stages. First, clinical notes preceding the reference H\&P are transformed into a de-duplicated patient timeline of timestamped atomic facts. Second, the patient timeline and reference H\&P are provided to an LLM to generate clinically relevant question--answer pairs. Third, the supporting facts are mapped back to their occurrences in the clinical record to characterize the temporal retrieval demands of each question. The resulting entries contain a naturally phrased clinical question, a reference answer, supporting facts, and metadata pertaining to the question's reasoning, clinical topic, and temporality. The following sections describe each stage in detail. All generative tasks used a PHI-compliant instance of Gemini-Pro 2.5 Chat, which achieved stellar performance on the LMArena long-query and overall text leader boards at the time of development \cite{chiang_chatbot_2024}.

\xhdr{Fact Extraction.}
Longitudinal clinical notes served as the source for question generation. Raw clinical notes are flawed documents, containing remnants of formatting, duplicated information due to copy-forward, clinical shorthand, and temporal ambiguity from interleaved information recorded in other visits. This ``note bloat'' from text adds little new information and can impede LLM reasoning by reducing the amount of information that fits within the model context \cite{cahoon2026clinicalnotebloatreduction}.

To reduce note bloat before question generation, we transformed the longitudinal record into a patient timeline of timestamped atomic claims. Atomic claims are logical propositions that cannot be further decomposed (e.g. ``Patient was diagnosed with Type 2 Diabetes (2021-09-13)'') \cite{russell_philosophy_2009}. Prior work has demonstrated that LLMs excel at extracting atomic claims from clinical notes, which in turn can be interpreted as patient \emph{facts} \cite{chung_verifact_2025,munnangi_factehr_2024}. Representing the record as facts removes much of the formatting, shorthand, and repeated text present in the original notes while providing a structured representation for downstream question generation.

Facts and their associated timestamps were extracted from clinical notes preceding the most recent H\&P note, which served as the index note. Supplementary Figure \ref{sfig:fact-extract} presents the prompt used for fact extraction. During fact extraction, timestamps were normalized by interpreting relative temporal expressions with respect to the source note date (e.g., ``last week'' was assigned a date one week before the note date). The resulting fact list was then de-duplicated to remove redundant entries referring to the same event. Supplementary Figure \ref{sfig:fact-dedup} presents the prompt used for de-duplication. This process reduced the mean token count to approximately one-third of that of the raw notes, as shown in Supplementary Table \ref{stab:fact_count} which compares token counts before and after processing.

To keep extracted fact lists within the model’s output token limit, clinical notes were processed in chunks of 20,000 characters. After fact extraction, the facts from all notes were merged in chronological order based on note sequence. Even after de-duplicating facts across notes, four patients still had facts that exceeded the Gemini 2.5 Pro context window. In these cases, we filtered out facts that were from inpatient encounters but not in the discharge summary.

\xhdr{Iterative Question Generation.} \label{meth:generation} Question generation used the patient timeline and the index H\&P to produce encounter-relevant question--answer pairs. The reference H\&P provided admission-specific context for identifying clinically relevant information needs, whereas the patient timeline supplied the facts available to answer each question. Reference answers were required to be fully supported by the timeline, and supporting facts were extracted from the atomic fact timeline.

Generation proceeded through sequential user--model exchanges. The prompt encouraged diversity in question type, reasoning requirements, and temporal scope \cite{raman_its_2024}. Previous exchanges were retained in context to discourage repetition and broaden topical coverage. For each user--model exchange, the generator produced a question, a naturally phrased version of the question, a reference answer, supporting facts, a question type, and a clinical relevance rationale. The rationale was included to encourage chain-of-thought reasoning about the question's clinical significance. Supplementary Figure \ref{fig:question-generation} presents the generation prompt, including the question-type definitions and output requirements. Generation was repeated until 15 candidate questions were obtained for each patient encounter.

To represent different retrieval demands and temporal scopes, questions were generated in three categories: single-hop (recent), single-hop (past), and multi-hop. Single-hop questions targeted one event occurring either less than two years (recent) or more than two years (past) before admission. Multi-hop questions required integrating information across facts or time points, such as linking related facts, grouping information by diagnosis, or comparing findings over time. The naturally phrased version was introduced because the initially generated questions were often more specific and detailed than questions clinicians would typically pose during chart review. Each naturally phrased question remained linked to the same reference answer and supporting facts and was used for retrieval evaluation.

Following question drafting, a separate model assigned up to three clinical topic labels to each question. These labels described clinical content and were distinct from the three question types. The accepted topics were allergies, appointments, assessment/plan, comorbidities, communications, demographics, devices/implants, diagnostic testing, disease progression status, family history, immunizations, laboratory tests, payment, physical exam, prescriptions, procedures/surgeries, radiology/imaging, reason for care, social determinants of health, and vitals. Supplementary Figure \ref{sfig:quesiton-classification} presents the topic-classification prompt. Assigned topic labels were subsequently verified by clinicians during review.

The 15 drafted questions were then filtered by the model to select 10 questions using the reference H\&P, question reasoning type, and question topics. Selection prioritized relevance to the patient's presenting concerns, answer verifiability, and diversity across clinical domains, while avoiding redundancy. The filtering prompt also instructed the model to exclude questions requiring information newly introduced in the reference H\&P. Supplementary Figure \ref{fig:question-filter} presents the filtering prompt and selection criteria. 

\xhdr{Temporality Estimation.}
\label{meth:temporality}
We sought to quantify how far back into the preceding clinical record a retrieval system would need to search to answer each BRIE question. 
We defined temporality as the look-back interval from the reference H\&P needed to retrieve all supporting facts needed to answer the question. Because supporting information may be repeated across notes through copy-forward documentation, the original source of a fact may overestimate how far back a system must search. To account for facts repeated through copy-forward, we identified each supporting fact’s most recent mention before the reference H\&P. Question-level temporality was defined by the supporting fact whose most recent mention occurred farthest back in the record. 

Fact mentions were identified through a three-stage retrieve-and-confirm pipeline, inspired by \cite{chung_verifact_2025}. Supporting facts were first decomposed into atomic clinical claims (Supplementary Figure \ref{sfig:fact-decomp}), and the corresponding evidence span was extracted from the source note (Supplementary Figure \ref{fig:fact-evidence}). We then searched the record for additional mentions of each claim using lexical retrieval and dense retrieval. Lexical retrieval used note-level BM25 \cite{robertson1994bm25,robertson2009bm25beyond,robertson1994okapitrec3} after lowercasing and removing stop words and generic clinical filler terms (Supplementary Table \ref{stab:stopwords}). Notes with positive BM25 scores for retrieval with fact and evidence spans were retained. Dense retrieval followed the same chunking and embedding procedure described in Online Methods Section \ref{meth:inference}; notes containing a chunk with cosine similarity $\geq0.40$ were ranked by their highest-scoring chunk, with the top 200 retained. Notes containing an exact match of the source evidence span were also included. Candidate notes were verified to determine whether they documented the same clinical event or observation as the corresponding claim. Exact evidence-span matches were automatically confirmed, and remaining candidates were adjudicated by Gemini 3.1 Flash Lite using the prompt shown in Supplementary Figure \ref{sfig:fact-dating}. This process results in a set of documents and evidence spans for each decomposed fact. Supplementary Figure \ref{sfig:fact-count} shows the distribution of decomposed fact counts per question. Supplementary Figure \ref{sfig:fact-token} shows the distributions of token distances to the earliest and latest confirmed fact mentions and the difference between these distances.

\section{Validation of Benchmark Generator and BRIE Curation}
\label{meth:validation}
We conducted a clinician annotation study to validate the quality of generated questions, answers, and supporting facts, and used these annotations to construct the final BRIE dataset.

\noindent\textbf{Generator Validation.}
Fifteen physicians evaluated the generated questions, answers, and supporting facts. During annotation, clinicians had access to the reference encounter information and the complete patient chart, allowing them to assess each generated entry against the underlying clinical record. Clinicians evaluated question quality using binary assessments of clinical relevance, verifiability, leakage, and consistency with the patient chart. Leakage was defined as inclusion of visit-specific information from the reference H\&P that would not have been available from the preceding record. Clinicians also verified the assigned reasoning and clinical topic labels and could optionally rewrite questions that were not naturally phrased. Answers were assessed for missing, irrelevant, or hallucinated information, and supporting facts were assessed for grounding in the patient chart. Each entry was reviewed by two clinicians.

\xhdr{Filtering and curation.} 
We used the clinician annotations to filter out low-quality entries and construct the curated BRIE dataset. Of the original 750 annotated entries, 75 were excluded because of incomplete annotations, yielding a pre-filtered dataset of 675 entries ($\text{BRIE}_\text{unfiltered}$). For the curated BRIE dataset, we retained questions judged clinically relevant and free of hallucinated content, together with answers that at least one annotator judged to contain no hallucinated, missing, or irrelevant information. This resulted in a final dataset of 508 questions. The full filtering breakdown is reported in Supplementary Table \ref{stab:filtering}.

Disagreements in clinical relevance or natural phrasing were adjudicated by a third clinician. For answer evaluation, annotators disagreed on consistency, relevance, and completeness for 41, 32, and 116 entries, respectively. We manually reviewed a random sample of 50 entries with answer-level disagreement to characterize the sources of disagreement and found that disagreement largely stemmed from clinician preferences about level of detail and was unlikely to impact downstream clinical action.

\section{Model and Inference Configurations}
\label{meth:inference}
We evaluated nine language models under five inference configurations representing long-context, retrieval-augmented, and agentic approaches. 
All models were evaluated with four common configurations: \textit{Recent}, \textit{Recent-180K}, \textit{BM25}, and \textit{Dense} while the \textit{Agentic} configuration was evaluated only with the two Claude models, yielding 38 model--inference configurations in total. Each retrieval inference configuration was used to generate one response per BRIE question.

\xhdr{Models.}
We evaluated models spanning multiple families, parameter scales, and access types. Using PHI-secure deployments, we generated responses with Gemini Pro 2.5 \cite{comanici2025gemini25}, Gemini Flash Lite 2.5 \cite{google2026gemini25family}, GPT 5.4 \cite{openai2026gpt54blog,singh2026openaigpt5card}, GPT 5.4 Nano \cite{openai2026gpt54nanoblog}, Claude Opus 4.7 \cite{anthropic2026claudeopus47blog}, Claude Haiku 4.5 \cite{anthropic2025claudehaiku45blog}, Kimi K2.6 \cite{kimiTeam2025kimik2,moonshot2026kimik26}, Qwen 3.5 397B \cite{qwen3.5}, and Qwen 3.5 27B \cite{qwen3.5}. Default inference settings were used unless otherwise specified.

\xhdr{Inference Configurations.} We evaluated models under five inference methods that represent commonly deployed information retrieval approaches. The two long-context configurations supplied models with the most recent notes directly, whereas the BM25 and Dense configurations first retrieved a subset of the record. The \textit{Agentic} configuration instead allowed the model to navigate the record iteratively using retrieval and summarization tools. Implementation details are described below.

\noindent\textbf{Recent and Recent-180K.}
For both long-context configurations, notes were ordered from most to least recent and greedily added until the token budget was reached. \textit{Recent} used each model's native context limit, as specified in Supplementary Table \ref{stab:context-limits}, whereas \textit{Recent-180K} used a fixed budget of 180{,}000 tokens across models. Token counts were estimated using tiktoken \cite{tiktoken}. Because tiktoken only approximates token counts for non-OpenAI models, notes were removed when necessary to ensure that inputs fit within the model's native context window. Each note was provided with its date and note type.

\noindent\textbf{Vector Retrieval.}
For the \textit{BM25} and \textit{Dense} configurations, each note was segmented into 4{,}000-character chunks with 400-character overlap using a recursive character splitter that preserved whitespace boundaries. Each chunk retained the source note title and timestamp. At inference time, the BRIE question was used to retrieve the top $K=50$ chunks, which were concatenated in rank order and provided to the language model.

\textit{BM25} ranked chunks using Okapi BM25 \cite{robertson1994bm25,robertson2009bm25beyond,robertson1994okapitrec3} with default parameters ($k_1=1.5$, $b=0.75$, $\epsilon=0.25$) over lowercased word-level tokens matched on word boundaries. \textit{Dense} retrieval encoded chunks and questions using Octen-Embedding-8B \cite{octenTeam2025rteb,octen2025embedding8b}, which was selected for its superior performance on the Retrieval Embedding Benchmark Healthcare tasks at the time of development \cite{rteb2025}. Passage embeddings were generated with the model's passage prefix and a maximum sequence length of 2{,}048 tokens, L2-normalized ($d=4096$), and ranked by cosine similarity to the question embedding. 
As an exploratory analysis, we also evaluated  a late interaction retrieval model for Kimi K2.6, Qwen 3.5 27B, and Qwen 3.5 397B. 
Late interaction preserves fine-grained signal by ranking chunks with MaxSim score, the sum of each query token's maximum similarity to any chunk token. Notes were segmented into 800-character chunks with 80-character overlap and ranked with LightOn's LateOn model \cite{sourty2026denseonlateon}. The top 50 chunks were provided to the language model. Because late-interaction retrieval did not outperform the other retrieval methods, it was excluded from the main analysis. Supplementary Figure \ref{sfig:late} reports fact recall and precision for these experiments.

\noindent\textbf{Agent.}
We implemented an agentic retrieval baseline in which a LLM identifies relevant information by autonomously navigating patient notes. Because agentic harnesses can widely vary, we equip the agent with tools that are commonly found in the literature to get a reasonable performance estimate \cite{moll2026agenticclinicalreasoninglongitudinal,cinarkoras2026configurableclinicalinformationextraction}. The agent has access to a set of four tools: (1) \texttt{get\_patient\_meta} called once at the start of inference to return the number of notes, counts by note type, and overall date range; (2) \texttt{search\_notes}: filters notes by case-insensitive keywords, note type, and inclusive date range, returning the note identifier, note type, date, and first 100 characters in reverse chronological order;
(3) \texttt{get\_note}:  returns the full text of a selected note; (4)  \texttt{summarize\_notes}:  generates a summary of a batch of notes.

After receiving the question, index time, and patient metadata, the agent first generated a retrieval plan specifying candidate note types, time windows, search terms, and anticipated tool calls. It then entered an iterative loop in which it issued a tool call, reasoned over the returned information, and updated its answer draft. The agent could terminate and return an answer after any turn. We set a maximum of 300 turns, although no query reached this limit. To limit context growth, raw tool outputs were discarded after each step, while the tool call, reasoning trace, and current answer draft were retained. The accumulated reasoning trace was additionally summarized every 10 turns. The \textit{Agentic} configuration was evaluated with Claude Opus 4.7 and Claude Haiku 4.5. Across BRIE queries, Opus required a mean of 4.14 turns (SD=4.97) and Haiku 6.18 turns (SD=3.31) before returning an answer. Supplementary Table \ref{stab:agent-steps} reports step-count statistics, and Supplementary Table \ref{stab:agent-usage} reports tool-use frequencies across queries.

\section{Retrieval Evaluation}
Due to the number of inference configurations and BRIE entries, it is intractable to manually review all responses. Additionally, existing heuristic scores are shown to be poorly correlated with human judgment \cite{fleming_medalign_2024, Bedi2026}. Therefore, we evaluated model responses using two complementary approaches: fact entailment, which measures factual coverage and precision relative to the reference answer, and pairwise comparison, which assesses relative response quality. Alignment to human annotators was assessed to verify the accuracy of automated evaluation. Hallucination rate and inference cost was also assessed. 

\xhdr{Fact Entailment.}
BRIE questions require recovering specific clinical facts from the patient timeline. To accommodate differences in wording and granularity, we decomposed reference answers and model responses into atomic clinical facts and compared them using semantic entailment \cite{grolleau2026medfacteval}.

Let $R$ denote the set of reference facts and $C$ the set of response facts. For a fact $f$ and a set of facts $S$, define $E(f,S)=1$ if $f$ is semantically entailed by $S$, and $0$ otherwise. Fact recall and precision were defined as
\[
\mathrm{Recall} =
\frac{\sum_{r \in R} E(r,C)}{|R|},
\qquad
\mathrm{Precision} =
\frac{\sum_{c \in C} E(c,R)}{|C|}.
\]
Recall measures the proportion of reference facts supported by the response, whereas precision measures the proportion of response facts supported by the reference. Entailment judgments were made using Gemini 3.1 Flash Lite. Supplementary Figure \ref{sfig:entailment} presents the entailment prompt and describes how the direction of comparison was reversed for precision.

\xhdr{Fact Entailment Tuning and Verification.}
We validated automated fact entailment against manual annotations to assess alignment with human judgment. Sixty responses were randomly sampled across model and inference configurations. Ten responses were annotated by one clinical informatics annotator and used to refine the entailment prompt; the remaining 50 were independently annotated by two clinical informatics annotators and held out for evaluation.

For the 50 held-out responses, pairwise Cohen's kappa ($\kappa$) was calculated among the two human annotators and the optimized LLM evaluator using individual fact-entailment judgments. Agreement between the automated evaluator and annotators 1 and 2 was moderate for recall ($\kappa=0.51$ and $0.60$, respectively) and substantial for precision ($\kappa=0.79$ and $0.74$, respectively). The corresponding inter-annotator agreement was $\kappa=0.61$ for recall and $\kappa=0.78$ for precision. Supplementary Figure \ref{sfig:heatmap-kappa-fact} shows pairwise $\kappa$ among the two annotators and the LLM evaluator. Overall, agreement between the automated evaluator and human annotators was similar to the agreement observed between human annotators.

\xhdr{Win-rate.}
Relative comparisons can distinguish two responses that score similarly in objective metrics but have different downstream clinical impact. Pairs of responses were evaluated for completeness, relevance, and concision using the naturally phrased query and reference answer as context \cite{Bedi2026}. Supplementary Figure \ref{sfig:winrate} presents the prompt used for pairwise evaluation. To reduce positional bias, each response pair was evaluated twice with the presentation order reversed \cite{zheng2023judging}. Comparisons that changed outcome after order reversal were treated as ties. Win-rate was calculated as $\frac{WINS + \frac{1}{2} TIES}{TOTAL}$. To reduce the number of comparisons, pairwise evaluation was conducted within model types and inference types. Gemini Flash Lite 3.1 was used for pairwise evaluation because of its efficiency and cost. 

\xhdr{Win-rate Verification.}
We validated automated pairwise evaluation against preferences from two physicians. Each physician independently evaluated 50 randomly sampled response pairs spanning model and inference configurations for completeness, relevance, and concision using the same criteria as the automated evaluator.
Cohen's kappa ($\kappa$) was calculated between the human raters and the automated evaluation for each comparison axis. Agreement between the automated evaluator and annotators 1 and 2 was $\kappa=0.44$ and $0.61$ for completeness, $\kappa=0.39$ and $0.03$ for relevance, and $\kappa=0.48$ and $0.47$ for concision, respectively. The corresponding inter-annotator agreement was $\kappa=0.37$ for completeness, $\kappa=0.29$ for relevance, and $\kappa=0.71$ for concision. Supplementary Figure \ref{sfig:heatmap-kappa-elo} shows pairwise $\kappa$ among the two physicians and the automated evaluator. These results indicate that alignment between automated pairwise evaluation and physician preferences were comparable to inter-annotator agreement for clinically relevant axes such as completeness and relevancy. These findings support automated win-rate evaluation, particularly for completeness, where agreement with both physicians exceeded inter-annotator agreement. 

\xhdr{Hallucination Detection.}
Model responses frequently contained facts that were not found in the reference answers. To assess if those facts were grounded in the patient records, unentailed response facts were assessed with the same procedure outlined in Online Methods Section  \ref{meth:temporality}. Facts were faithful if at least one supporting evidence span could be located. Supplementary Figure \ref{sfig:hallucination} shows the prompt used to detect hallucinations. We conducted this procedure on 50 BRIE questions across all model and inference configurations. Across inference methods, nearly every unentailed fact could be supported by the record. The full results are reported in Supplementary Table \ref{stab:hallucination}.

\xhdr{Cost calculations.}
We estimated inference cost using token counts from tiktoken \cite{tiktoken} and per-token pricing from the corresponding model providers. Pricing was obtained from Google Cloud Agent Platform for Gemini and Claude models\footnote{\url{https://cloud.google.com/gemini-enterprise-agent-platform/generative-ai/pricing}}, Microsoft Azure OpenAI for GPT models\footnote{\url{https://azure.microsoft.com/en-us/pricing/details/azure-openai/}}, the Kimi pricing page for Kimi models\footnote{\url{https://www.kimi.com/resources/kimi-k2-6-pricing}}, and OpenRouter for Qwen models\footnote{\url{https://openrouter.ai/qwen}} (accessed June 20, 2026).

Input token counts were estimated based on inference configuration: \textit{Recent} used the naive context window for each model; \textit{Recent-180K} used 180,000 tokens; RAG-based (\textit{Dense}, \textit{BM25}) used 50K tokens (top-50  documents); and \textit{Agentic} used token counts from empirical measurements recorded during experiments. Output length was fixed at 250 tokens per query across all settings. Per-query costs for each model and inference strategy are reported in Supplementary Table \ref{stab:inference-cost}.

\section{Multiple answer analysis.}
\label{meth:regeneration}
A natural clinical question may admit multiple reasonable answers depending on its interpretation, including differences in temporal scope, topic focus, or level of detail. A single reference answer may therefore not capture the full range of acceptable responses. To assess the sensitivity of BRIE evaluation to this ambiguity, we generated and clinically validated alternative reference answers for 100 sampled queries, then re-evaluated model responses against the resulting multi-reference sets.

\xhdr{Alternative Answer generation.}
For each query, we first generated 8--10 candidate interpretations, or ``seeds,'' using the prompt in Supplementary Figure \ref{sfig:answer-seed}. Seeds varied along three axes: time frame, topic focus, and  level of detail. The original query and reference question--answer pair were provided as inputs,  and each seed was required to differ meaningfully from both the original reference and the other candidates. 

Because seed generation did not access the patient record, we next grounded each interpretation in the available clinical history. Each seed's time frame was resolved to an inclusive date range using dates from the patient fact timeline, and facts within that range were retained. Seeds with no supporting facts were discarded. Supplementary Figure \ref{sfig:answer-timeframe} presents the time-frame resolution prompt.
For each remaining seed, we generated a specific question, reference answer, and verbatim supporting facts from the retained patient facts. Generation was additionally conditioned on a randomly sampled clinician persona---emergency department physician, admitting resident, attending physician, pharmacist, registered nurse, subspecialty consultant, or social worker---to encourage variation in clinical perspective. The supplied facts constrained the generated answer. Supplementary Figure \ref{sfig:answer-grounding} presents the prompt for generating these grounded pairs.
Finally, an LLM filtered and de-duplicated the resulting pairs, removing interpretations that lacked a well-defined answer or substantially overlapped with another candidate or the original reference. The model selected 2--5 distinct interpretations per query for clinician validation using the prompt in Supplementary Figure \ref{sfig:answer-dedup}.

\xhdr{Answer verification.}
Four board-certified physicians evaluated the generated question--answer pairs to identify additional acceptable reference answers for the original BRIE queries. Generated questions were considered reasonable interpretations if they clarified the topic, time window, or expected answer details (e.g., dates or dosages) while remaining consistent with the original query. Generated answers were evaluated for missing, hallucinated, and irrelevant information, and clinicians could edit both the question and answer for accuracy.

Supplementary Figure \ref{sfig:answer-validation} shows the proportion of answers associated with accepted interpretations that were judged accurate, complete, and relevant. The validated answers corresponding to reasonable interpretations, incorporating any clinician edits, were grouped by their original BRIE query to form a set of reference answers for evaluation. Supplementary Figure \ref{sfig:answer-count} shows the number of validated reference answers per query.

\xhdr{Multi-reference evaluation.}
Model responses generated using the inference configurations described in Online Methods Section \ref{meth:inference} were evaluated against each clinician-validated reference answer using fact entailment.  For each response, we reported the highest recall and precision obtained across the available reference answers, reflecting performance under the best-supported interpretation of the original query.

\section{Benchmark Regeneration Analysis.}
We assessed whether the BRIE generation framework could be applied to more recent clinical data without additional clinician curation. To do so, we regenerated the benchmark on a temporally held-out cohort and compared retrieval performance between the original automatically generated cohort, $\text{BRIE}_{\text{unfiltered}}$, and the regenerated cohort, $\text{BRIE}_{\text{new}}$.

\xhdr{Temporal Cohort.}
Because documentation practices may change over time, we constructed an additional cohort of recent encounters to assess the temporal generalization of the benchmark generation framework. Using the sampling and generation procedures described in Online Methods Section \ref{meth:cohort}, we sampled 100 additional patients admitted in 2026 and generated 1,000 question--answer pairs. We refer to this cohort as $\text{BRIE}_{\text{new}}$. No filtering or clinician editing was performed.

\xhdr{Performance Assessment.}
To assess whether retrieval performance patterns persisted across cohorts, we evaluated the same model and inference configurations on $\text{BRIE}_{\text{unfiltered}}$ and $\text{BRIE}_{\text{new}}$. We generated responses using Qwen 3.5 27B with the \emph{Recent}, \emph{Recent-180K}, \emph{BM25}, and \emph{Dense} inference methods, limiting evaluation to one model to keep inference costs manageable. Descriptions of the inference configurations are detailed in the Online Methods Section \ref{meth:inference}. For each query, fact recall and precision were calculated against its single generated reference answer using semantic fact entailment. Performance was reported for the automatically assigned topics and reasoning categories. Both cohorts were evaluated without clinician editing or filtering.

\section{Statistical Analyses.}
We used bootstrap confidence intervals and nonparametric hypothesis tests to quantify uncertainty and assess differences across experimental conditions. We constructed 95\% confidence intervals from 1{,}000 bootstrap resamples. For fact entailment, we used max-$t$ correction to control multiplicity across 38 tests. Supplementary Tables \ref{tab:fact-recall-bootstrap-model-inference} and \ref{tab:fact-precision-bootstrap-model-inference} show the adjusted confidence intervals for fact recall and precision across all inference configurations, respectively. We also reported confidence intervals for all clinician metrics used to evaluate the generator.

Statistical tests were selected according to the structure of each analysis. To compare fact recall between question types, we used two-sided Mann--Whitney U tests within each question category, with the counts specified in  Supplementary Table \ref{tab:question-type-counts}.  Benjamini–Hochberg correction was conducted for models and inference methods for each reasoning, temporality, and topic. The complete results are presented in Supplementary Tables \ref{tab:question-type-reasoning}--\ref{tab:question-type-time}. For paired analyses, including comparison of single- and multiple-reference evaluation for the same questions, we used two-sided Wilcoxon signed-rank tests, corrected across inference configurations separately for fact recall and precision (Supplementary Tables \ref{tab:fact-recall-specification-wilcoxon} and \ref{tab:fact-precision-specification-wilcoxon}, respectively).
Finally, we assessed the impact of temporal drift on BRIE difficulty. We conducted one-sided Welch tests to test for non-inferiority at an absolute margin of $\delta=0.05$ across each question category and overall for both $\text{BRIE}_{\text{new}}$ and $\text{BRIE}_{\text{unfiltered}}$, correcting within question category using the Benjamini--Hochberg procedure. Supplementary Table \ref{tab:fact-recall-not-easier} presents the complete results, where fact recall of both cohorts was found not inferior to that of BRIE. We also conducted Kolmogorov--Smirnov tests between $\text{BRIE}$ and $\text{BRIE}_{\text{new}}$ to evaluate whether the fact recall cumulative distribution functions differed within question categories, adjusting within question category with the Benjamini--Hochberg procedure. Complete results are presented in Supplementary Table \ref{tab:fact-recall-specification-ks}.

%% file: SI/SI-notes.tex
\vspace{1em}
\section*{Annotation guidelines}
\label{apd:annotation}
Complete annotation guidelines for validating the generator guidelines are provided \href{https://docs.google.com/document/d/1NRLf_ibjQk8GC1clnA5L8CfgkjJoe2Vgc30AWIbMZvk/edit?usp=sharing}{here}. Complete guidelines for validating multiple answer generation are provided \href{https://docs.google.com/document/d/1f1Bniwh_HIUWyeg4iaIKw6fXFABDX515f4ItPJ5eebU/edit?usp=sharing}{here}.

%% file: SI/SI-figures.tex

\begin{figure}[H]
\centering
\begin{tcolorbox}[
  unbreakable, enhanced, colback=backblue, colframe=frameblue,
  title=\textbf{Prompt: Fact Extraction},
  fonttitle=\bfseries, fontupper=\small, left=4pt, right=4pt,
  top=3pt, bottom=3pt
]

\textbf{Role.} Act as a clinician performing chart review on a patient just
admitted to the hospital. Generate a list of atomic claims from a given excerpt
of a clinical note.

\smallskip
\textbf{Atomic claim definition.} An atomic claim is a phrase or sentence making
a single assertion --- factual or a hypothesis posed by the text. Atomic claims
are indivisible (cannot be decomposed into more fundamental claims) and must have
a subject, predicate, and object, where the predicate relates the subject to the
object. More complex statements can be composed from atomic claims.

\smallskip
\textbf{Do.}
\begin{itemize}[leftmargin=1.2em, itemsep=1pt, topsep=2pt]
  \item Extract discrete atomic claims from the \texttt{"text"} field. Each must
    include a subject, predicate, and object and stand alone without ambiguity.
  \item Include only clinically relevant claims (symptoms, procedures, tests,
    medications, diagnoses, clinical locations).
  \item Use only the provided text; add no outside knowledge or assumptions.
    Preserve the full context of each claim.
  \item Write each claim in the shortest unambiguous form; avoid pronouns or
    vague references.
  \item Always refer to the subject as ``patient,'' even if the text uses a name
    or identifier.
  \item Append a date (\texttt{YYYY-MM-DD}) to every claim:
  \begin{itemize}[leftmargin=1.2em, itemsep=0pt, topsep=1pt]
    \item If an absolute date is given, use it.
    \item If a relative reference is used (e.g., ``yesterday,'' ``last week''),
      resolve it against \texttt{note\_date}.
    \item If no event date is given, use \texttt{note\_date}.
    \item For vague ranges (e.g., ``last month''), default to the first day of
      that period unless otherwise specified.
  \end{itemize}
  \item If there are no valid clinically relevant claims, return \texttt{"claims"}
    as an empty list \texttt{[]}.
\end{itemize}

\textbf{Do not.}
\begin{itemize}[leftmargin=1.2em, itemsep=1pt, topsep=2pt]
  \item Do not include claims not directly about the patient's clinical care
    (e.g., provider names, note authors, addenda, phone numbers, clinic
    addresses, administrative details).
  \item Do not invent or infer claims beyond what is explicitly stated.
\end{itemize}

\end{tcolorbox}
\caption{Condensed fact Extraction System Prompt with Time Normalization. Prompt adjusted from$^{64}$} 
\label{sfig:fact-extract}
\end{figure}

\begin{figure}
\centering
\begin{tcolorbox}[
  unbreakable, enhanced, colback=backblue, colframe=frameblue,
  title=\textbf{Prompt: Fact De-duplication},
  fonttitle=\bfseries, fontupper=\small, left=4pt, right=4pt,
  top=3pt, bottom=3pt
]

\textbf{Role.} Act as an expert clinician reviewing a list of patient facts.
Some facts may be duplicates or semantically redundant. Identify which facts
should be removed so the final list is concise, non-redundant, and retains all
unique clinical information. Do \emph{not} regenerate the list --- return only
the \emph{indices} of facts to remove.

\smallskip
\textbf{Redundancy rules.}
\begin{itemize}[leftmargin=1.2em, itemsep=1pt, topsep=2pt]
  \item A fact is redundant if it asserts the same claim as another, even if
    phrased differently (e.g., ``Patient has hypertension'' vs.\ ``History of
    high blood pressure'').
  \item If two facts are identical except for timestamps, keep the most complete
    one (more timestamps).
  \item If one fact is a subset of another (e.g., ``Admitted to hospital'' vs.\
    ``Admitted to hospital (2014-08-01)''), mark the subset for removal.
  \item \textbf{Conflicting facts:} if two facts make contradictory claims, keep
    \emph{both} --- mark neither as redundant.
  \item \textbf{Unique timestamps:} facts differing only by distinct timestamps
    are \emph{not} redundant and must both be kept (e.g., ``Admitted
    (2014-08-01)'' and ``Admitted (2014-09-01)'').
\end{itemize}

\textbf{Output (strict).} Return \emph{only} a JSON object with one key; no facts,
no prose, no explanation. If no redundancies are found, return an empty list.

\begin{verbatim}
{ "redundant_fact_indices": [ <0-based indices to remove> ] }
\end{verbatim}

\smallskip
\textbf{Example.} Input facts (indexed):
\begin{itemize}[leftmargin=1.2em, itemsep=0pt, topsep=2pt]
  \item[\texttt{0}:] Patient has hypertension
  \item[\texttt{1}:] History of high blood pressure
  \item[\texttt{2}:] Admitted to hospital (2014-08-01)
  \item[\texttt{3}:] Admitted to hospital (2014-09-01)
  \item[\texttt{4}:] Admitted to hospital
\end{itemize}

Output: \texttt{\{ "redundant\_fact\_indices": [1, 4] \}}

\emph{Rationale:} fact 1 duplicates fact 0; fact 4 is a subset of facts 2 and 3
(removed), while 2 and 3 are kept for their unique timestamps.
\end{tcolorbox}
\caption{Condensed prompt for removing duplicated facts due to copy-forward or imported structured data} \label{sfig:fact-dedup}
\end{figure}
\begin{table}[H]
\centering
\begin{tabular}{lcc}
\toprule
Stage & Count & Token Count \\
\midrule
Raw Note          & $(1.6 \pm 1.3)\times 10^{3}$ & $(1.0 \pm 0.9)\times 10^{5}$ \\
Fact              & $(3.3 \pm 3.0)\times 10^{3}$ & $(5.6 \pm 5.0)\times 10^{4}$ \\
\cmidrule(lr){1-3}
\makecell[l]{\textbf{Refined} \\ \textbf{Facts}} & $\mathbf{(2.1 \pm 1.9)\times 10^{3}}$ & $\mathbf{(3.3 \pm 2.2)\times 10^{4}}$ \\
\bottomrule
\end{tabular}
\caption{Count and number of tokens for each stage of fact extraction for 25 randomly sampled patients.  The \textbf{Refined Facts}, pruned for redundancy, was used for question generation.}
\label{stab:fact_count}
\end{table}

\begin{figure}[p]
    \centering
    
\begin{tcolorbox}[
  unbreakable, enhanced, colback=backblue, colframe=frameblue,
  title=\textbf{Prompt: Question Generation},
  fonttitle=\bfseries, fontupper=\small, left=4pt, right=4pt,
  top=3pt, bottom=3pt
]

\textbf{Role.} You are a clinician who is seeing a new patient and you are performing a comprehensive review of the patient.
Your job is to use the list of patient facts to generate retrieval-only questions, each with (a) a concise answer and (b) the exact supporting facts copied verbatim from the fact list. 

\smallskip
\textbf{Core rules.}
\begin{itemize}[leftmargin=1.2em, itemsep=1pt, topsep=2pt]
  \item Use only the provided facts; do not invent, summarize, interpret,
    calculate, or speculate.
  \item Ensure temporal diversity (early events, recent events, multi-timepoint
    trends).
  \item Prefer questions requiring reasoning or integration over trivial recall.
  \item Copy supporting facts exactly into \texttt{fact\_subset} (no paraphrase,
    trimming, or merging; dates verbatim). Include all directly supporting facts;
    for duplicates, keep the most relevant.
  \item Answers must be fully supported by \texttt{fact\_subset} and include all
    associated dates.
  \item Phrase questions in natural clinical language; avoid test-like wording.
    Omit explicit dates unless needed to distinguish similar events; otherwise use
    relative phrasing (``most recent,'' ``last 3 years'').
  \item Avoid subjective modifiers (``old,'' ``a while back'') and vague prompts
    (``tell me about,'' ``details''); anchor with terms like \emph{outcome},
    \emph{result}, \emph{findings}, \emph{assessment}.
  \item Each question must be clinically useful for admission and justified in
    \texttt{clinical\_relevance\_rationale}. If no valid question exists, return
    \texttt{[]}.
\end{itemize}

\textbf{\texttt{question\_rewrite}.} Rephrase \texttt{question} as a clinician
would naturally ask or think in context: remove redundant specificity, modifiers,
and date clutter; use relative/contextual time references when relevant; avoid
narrative or summarizing phrasing; preserve objective, unambiguous scope.

\smallskip
\textbf{Allowed topics (non-exhaustive).} Comorbidities; procedures/surgeries;
devices/implants; imaging; diagnostic/genetic/pathology testing; demographics;
prescriptions (type, interactions, side effects, contraindications); labs;
disease progression (severity, complications, staging, function); social
determinants; assessment \& plan; vitals; appointments; physical exams; family
history; communications; payment; reason for care; immunizations; allergies.

\smallskip
\textbf{Question types.}
\begin{itemize}[leftmargin=1.2em, itemsep=1pt, topsep=2pt]
  \item \texttt{single\_hop\_recent}: one event $<$2 years before admission.
  \item \texttt{single\_hop\_past}: one event $>$2 years before admission.
  \item \texttt{multi\_hop}: synthesis across timepoints/facts --- reasoning
    chains, clustering by diagnosis/label, or comparison over time.
\end{itemize}

\textbf{Output (strict).} Return \emph{only} a JSON array (no prose, no code fences). 

\end{tcolorbox}
    \caption{Question generation prompt}
    \label{fig:question-generation}
\end{figure}

\begin{figure}[p]
\vspace{-3.4em}
\centering

\begin{tcolorbox}[
  unbreakable, enhanced, colback=backblue, colframe=frameblue,
  title=\textbf{Prompt: Question Topic Classification},
  fonttitle=\bfseries, fontupper=\small, left=4pt, right=4pt,
  top=3pt, bottom=3pt
]

\textbf{Task.} Identify the top 3 most relevant topics in a given
patient-specific clinical question.

\smallskip
\textbf{Rules.}
\begin{itemize}[leftmargin=1.2em, itemsep=1pt, topsep=2pt]
  \item Choose only from the provided list of accepted topics.
  \item Select up to 3 topics most directly relevant to the question.
  \item Always return the result in valid JSON format.
\end{itemize}

\textbf{Accepted topics.} Allergies; Appointments; Assessment/Plan;
Comorbidities; Communications; Demographics; Devices/Implants; Diagnostic
testing; Disease Progression Status; Family History; Immunizations; Laboratory
tests; Payment; Physical Exam; Prescriptions; Procedures/Surgeries;
Radiology/Imaging; Reason for care; Social Determinants of Health; Vitals.

\end{tcolorbox}
\vspace{-1em}
\caption{Condensed prompt for topic classification.} 
\label{sfig:quesiton-classification}
\end{figure}

\begin{figure}[p]
\begin{tcolorbox}[
  unbreakable, enhanced, colback=backblue, colframe=frameblue,
  title=\textbf{Prompt: Question Filter},
  fonttitle=\bfseries, fontupper=\small, left=4pt, right=4pt,
  top=3pt, bottom=3pt
]

\textbf{Role.} Act as a clinical reasoning assistant. Select the 10 best
questions from the input JSON, using the admission H\&P note and question
metadata.

\smallskip
\textbf{Selection criteria.}
\begin{itemize}[leftmargin=1.2em, itemsep=1pt, topsep=2pt]
  \item \textbf{Most relevant:} must address the patient's presenting concerns,
    comorbidities, or contextual factors in the H\&P.
  \item \textbf{Most defined:} clear, specific, and objectively answerable; avoid
    vague, opinion-based, or overly broad questions.
  \item \textbf{Most diverse:} prioritize questions spanning different clinical
    domains (e.g., comorbidities, baseline function, risk factors, social
    history, medications, prior care); no redundancy.
  \item \textbf{No leakage:} exclude questions that require surfacing new
    information from the H\&P (e.g., newly reported symptoms, new disease
    progression details, or new medication changes).
\end{itemize}

\end{tcolorbox}
    \caption{Condensed prompt for filtering redundant questions.}
    \label{fig:question-filter}
\end{figure}


\begin{figure}[p]
    \centering
\begin{tcolorbox}[
  unbreakable, enhanced, colback=backblue, colframe=frameblue, coltitle=white,
  title=\textbf{Prompt: Atomic Fact Decomposition},
  fonttitle=\bfseries, fontupper=\small, left=4pt, right=4pt, top=3pt, bottom=3pt
]
\textbf{System.} You are a clinician performing chart review. Respond only with a
valid JSON object.

\smallskip
\textbf{Instruction.} Extract every atomic clinical claim from the answer below.

\smallskip
\textbf{Atomic claim definition.} An atomic claim makes a single assertion with a
subject, predicate, and object. It must stand alone without ambiguity and cannot
be decomposed into more fundamental claims.

\smallskip
\textbf{Do.}
\begin{enumerate}[leftmargin=1.4em, itemsep=1pt, topsep=2pt]
  \item Extract discrete atomic claims. Each must include a subject, predicate,
    and object.
  \item Include only clinically relevant claims (symptoms, procedures, tests,
    medications, diagnoses, clinical locations).
  \item Use only the provided text. Do not add outside knowledge or assumptions.
  \item Write each claim in the shortest unambiguous form. Avoid pronouns or vague
    references.
  \item Always refer to the subject as ``patient,'' even if the text uses a name
    or identifier.
  \item When a date is mentioned in the text, create one atomic fact that captures
    the date and the high-level event type (e.g., ``A TSH measurement was
    performed on May 30.''). All other sub-facts from that same event MUST NOT
    include the date.
  \item If there are no valid clinically relevant claims, return \texttt{"facts"}
    as an empty list \texttt{[]}.
\end{enumerate}

\textbf{Do not.}
\begin{enumerate}[leftmargin=1.4em, itemsep=1pt, topsep=2pt]
  \item Do not include claims unrelated to the patient's clinical care (e.g.,
    provider names, administrative details).
  \item Do not invent or infer claims beyond what is explicitly stated.
  \item Do not combine multiple events into a single claim.
  \item Do not append dates to sub-facts --- a measurement or finding is a separate
    atomic fact from the date on which it occurred.
\end{enumerate}

\textbf{Examples.} [EGD example and TSH example with date-anchor splitting.]

\tcblower
\footnotesize\itshape Sibling variant \texttt{ATOMIZE\_PROMPT}: same definition
and rules, but takes a numbered list of existing facts and returns
\texttt{\{"facts": [\{"text": ..., "source\_idx": N\}]\}} so each atomic claim
keeps provenance back to its source fact.

\end{tcolorbox}
\caption{Condensed prompt for fact extraction (input into fact entailment and fact dating)}
\label{sfig:fact-decomp}
\end{figure}

\begin{figure}[p]
    \centering
    \begin{tcolorbox}[
  unbreakable, enhanced, colback=backblue, colframe=frameblue, coltitle=white,
  title=\textbf{Prompt: Fact Evidence Identification},
  fonttitle=\bfseries, fontupper=\small, left=4pt, right=4pt,
  top=3pt, bottom=3pt
]

\textbf{System.} You are a clinical information extraction assistant. Extract the
exact substring from the note that most directly supports the given fact.

\smallskip
\textbf{User.}
\begin{verbatim}
Fact: {fact}

Note:
{note_text}
\end{verbatim}

\textbf{Instruction.} Return only the shortest exact substring from the note that
directly documents the SAME specific event or observation described by the fact
(same procedure, finding, or measurement --- not a related or similar one). No
explanation or surrounding text.
\end{tcolorbox}
    \caption{Condense prompt to identify substring evidence for fact}
    \label{fig:fact-evidence}
\end{figure}

\begin{figure}[p]
    \centering
    \begin{tcolorbox}[
  unbreakable, enhanced, colback=backblue, colframe=frameblue, coltitle=white,
  title=\textbf{Prompt: Cross-Note Confirmation},
  fonttitle=\bfseries, fontupper=\small, left=4pt, right=4pt,
  top=3pt, bottom=3pt
]

\textbf{System.} You are a clinical information extraction assistant. Given a
patient note and a list of clinical facts, identify which facts are independently
documented in the note.

\smallskip
\textbf{User.}
\begin{verbatim}
Note (Date: {note_date}, Title: {note_title}):
{note_text}

Facts to check:
1. Fact: {fact}
   Reference evidence: "{evidence_span}"
... (batched, 1-indexed)
\end{verbatim}

\textbf{Instruction.} For each fact, determine whether this note independently
documents EXACTLY the same procedure, finding, medication, or measurement.

\smallskip
\textbf{Strict rules --- reject the fact if any apply.}
\begin{itemize}[leftmargin=1.2em, itemsep=1pt, topsep=2pt]
  \item The note names a DIFFERENT procedure even if it occurred on the same date
    or involves the same anatomical region.
  \item The note only mentions a related concept without documenting the specific
    event.
  \item The note records a different medication, dose, or route.
  \item You are not highly confident the note refers to the identical event.
\end{itemize}

\textbf{Output.} Return a JSON array --- only for facts with clear, unambiguous
evidence in this note:
\begin{verbatim}
[{"fact_idx": <1-N>, "evidence": "<shortest exact supporting substring>"}]
\end{verbatim}

\end{tcolorbox}
    \caption{Condense prompt to identify all mentions of facts}
    \label{sfig:fact-dating}
\end{figure}
\begin{figure}
    \centering
    \includegraphics[width=\linewidth]{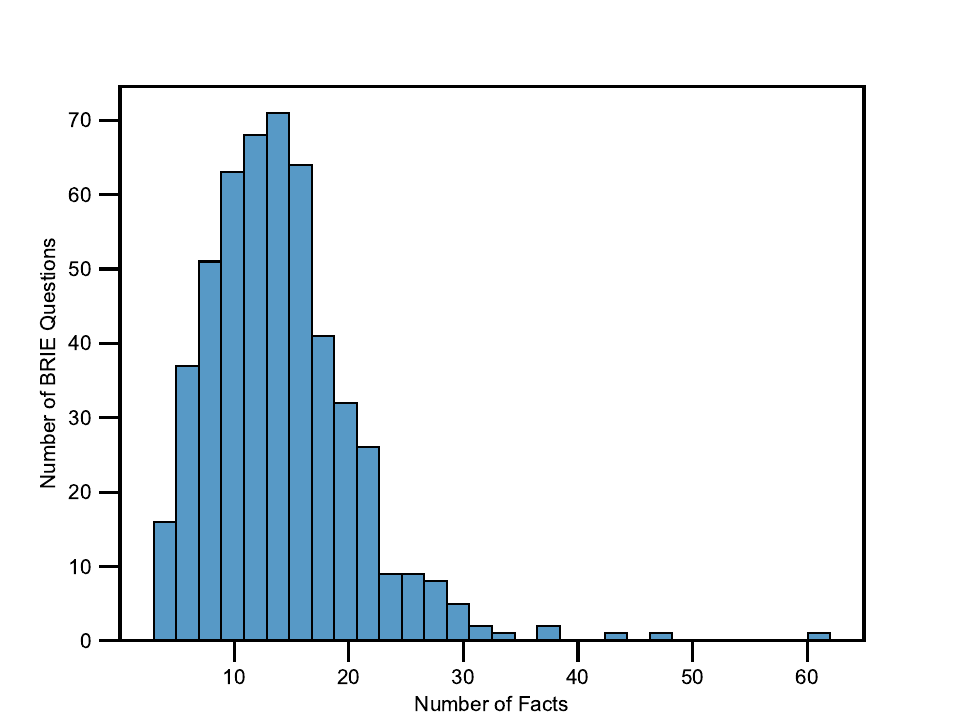}
    \caption{Number of facts per BRIE question after further decomposition}
    \label{sfig:fact-count}
\end{figure}

\begin{figure}
    \centering
    \includegraphics[width=\linewidth]{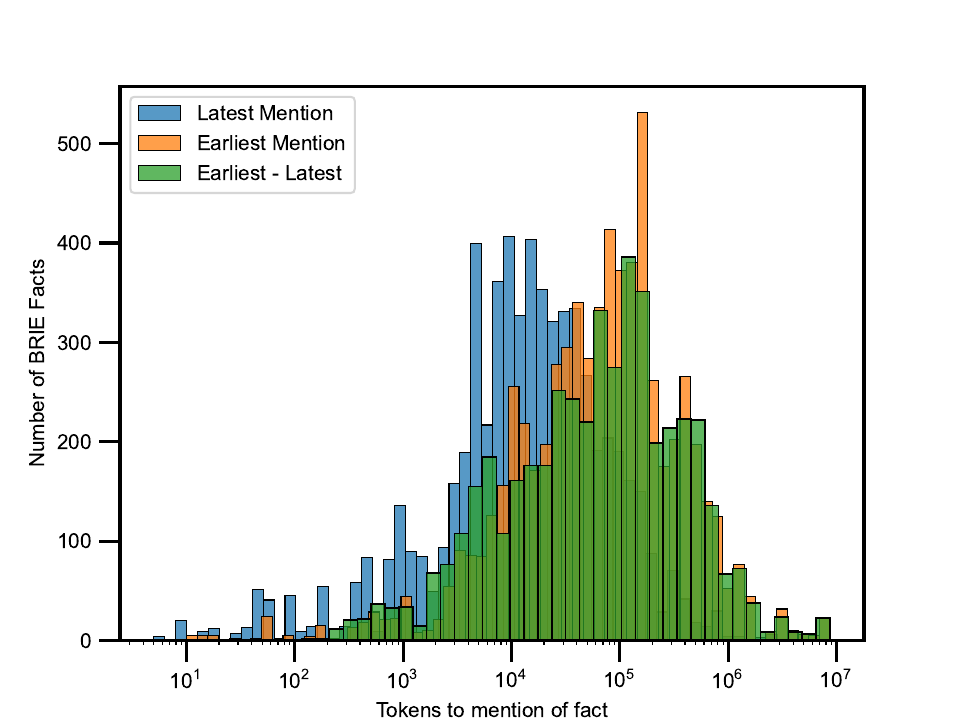}
    \caption{Number of tokens to fact mentions}
    \label{sfig:fact-token}
\end{figure}


\begin{table}[t]
\centering
\small
\renewcommand{\arraystretch}{1.3}
\begin{tabular}{@{}>{\raggedright\arraybackslash}p{0.22\textwidth} c >{\raggedright\arraybackslash}p{0.62\textwidth}@{}}
\toprule
\textbf{Category} & \textbf{n} & \textbf{Tokens} \\
\midrule
General function words & 113 &
a, an, the, and, or, but, in, on, at, to, for, of, with, by, from, up, about,
into, through, during, before, after, above, below, between, out, off, over,
under, then, here, there, when, where, how, all, both, each, more, most, other,
some, such, no, nor, not, only, same, so, than, too, very, can, will, just, now,
this, that, these, those, is, are, was, were, be, been, being, have, has, had,
do, does, did, would, could, should, may, might, must, shall, also, its, it, he,
she, they, we, i, my, his, her, their, our, your, which, who, whom, as, if,
while, although, however, therefore, thus, since, because, well, within,
without, including, via, per, see \\
\addlinespace
Clinical narrative \& charting & 45 &
patient, pt, px, history, hx, assessment, plan, reported, reports, noted, notes,
follow, followup, discharge, admission, admitted, presents, presenting,
presented, new, old, previous, prior, current, recent, stable, unchanged,
medical, surgical, social, family, review, reviewed, discussed, visit,
appointment, clinic, office, normal, abnormal, negative, positive, right, left,
bilateral \\
\addlinespace
Medication \& dosing & 22 &
tablet, tablets, tab, capsule, capsules, cap, oral, daily, prn, mg, ml, mcg,
mgs, mls, kg, dose, doses, dosing, prescribed, unit, units, cc \\
\addlinespace
Temporal \& measurement & 11 &
year, years, month, months, week, weeks, day, days, date, time, status \\
\bottomrule
\end{tabular}
\caption{Clinical stopword list, grouped by category.}
\label{stab:stopwords}
\end{table}

\begin{table}[t]
\centering
\begin{tabular}{llr r r}
\toprule
\multicolumn{2}{l}{\textbf{Characteristic}} & \textbf{n} & \textbf{\%} & \textbf{Ref.\ \%} \\
\midrule
\multicolumn{5}{l}{\textit{Gender}} \\
\quad Male   & & 44 & 58.7 & 49.9 \\
\quad Female & & 31 & 41.3 & 50.1 \\
\midrule
\multicolumn{5}{l}{\textit{Ethnicity and race}} \\
\quad Not Hispanic or Latino & White                     & 40 & 53.3 & 49.7 \\
\quad Hispanic or Latino     & No matching concept       & 14 & 18.7 & 13.6 \\
\quad Not Hispanic or Latino & Asian                     &  9 & 12.0 & 18.0 \\
\quad Not Hispanic or Latino & Black or African American &  4 &  5.3 &  5.6 \\
\quad Hispanic or Latino     & White                     &  4 &  5.3 &  4.4 \\
\quad Not Hispanic or Latino & No matching concept       &  3 &  4.0 &  5.2 \\
\quad Hispanic or Latino     & Asian                     &  1 &  1.3 &  0.1 \\
\midrule
\multicolumn{2}{l}{\textit{Age (years)}} & \multicolumn{3}{r}{} \\
\quad Mean (SD)    & & \multicolumn{2}{r}{64.8 (17.4)}       & 66.2 (17.2) \\
\quad Median [IQR] & & \multicolumn{2}{r}{66.0 [54.0, 78.5]} & 69.0 [56.0, 80.0] \\
\quad Range        & & \multicolumn{2}{r}{27 -- 91}          & 0 -- 94 \\
\bottomrule
\end{tabular}
\caption{Patient cohort demographics ($N = 75$). Ref.\ \% denotes the corresponding percentage in the full source population ($N = 25{,}443$).}
\label{stab:demographics}
\end{table}

\begin{figure}
    \centering
    \includegraphics[width=\linewidth]{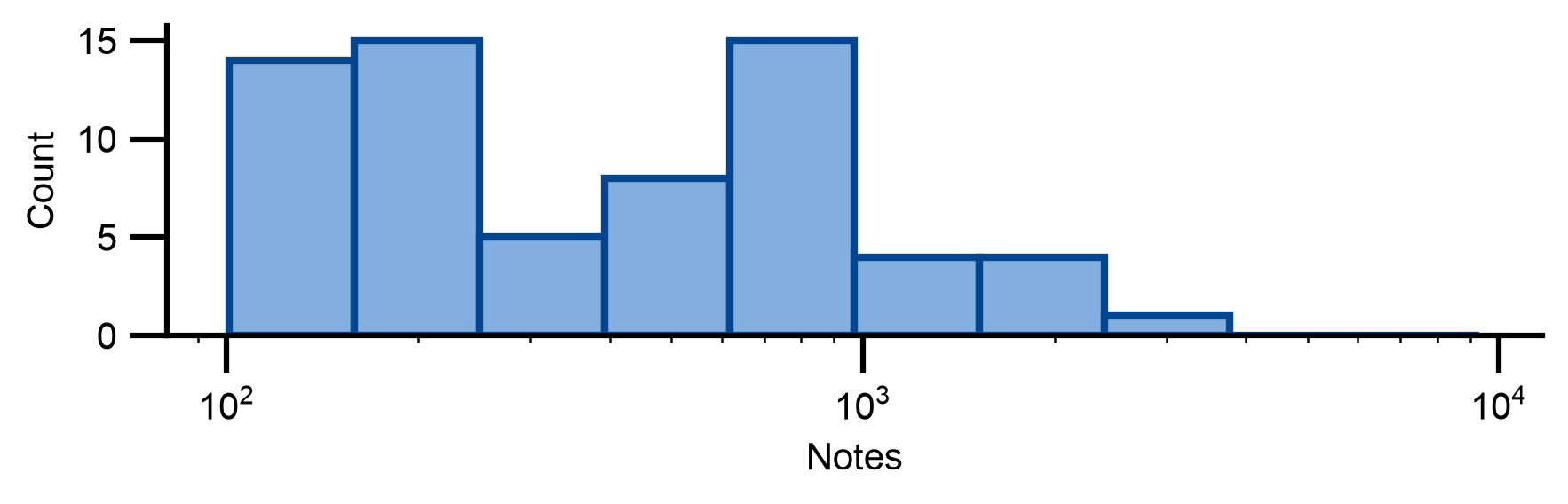}
    \caption{Number of notes per patient}
    \label{sfig:patient-note}
\end{figure}
\begin{figure}
    \centering
    \includegraphics[width=\linewidth]{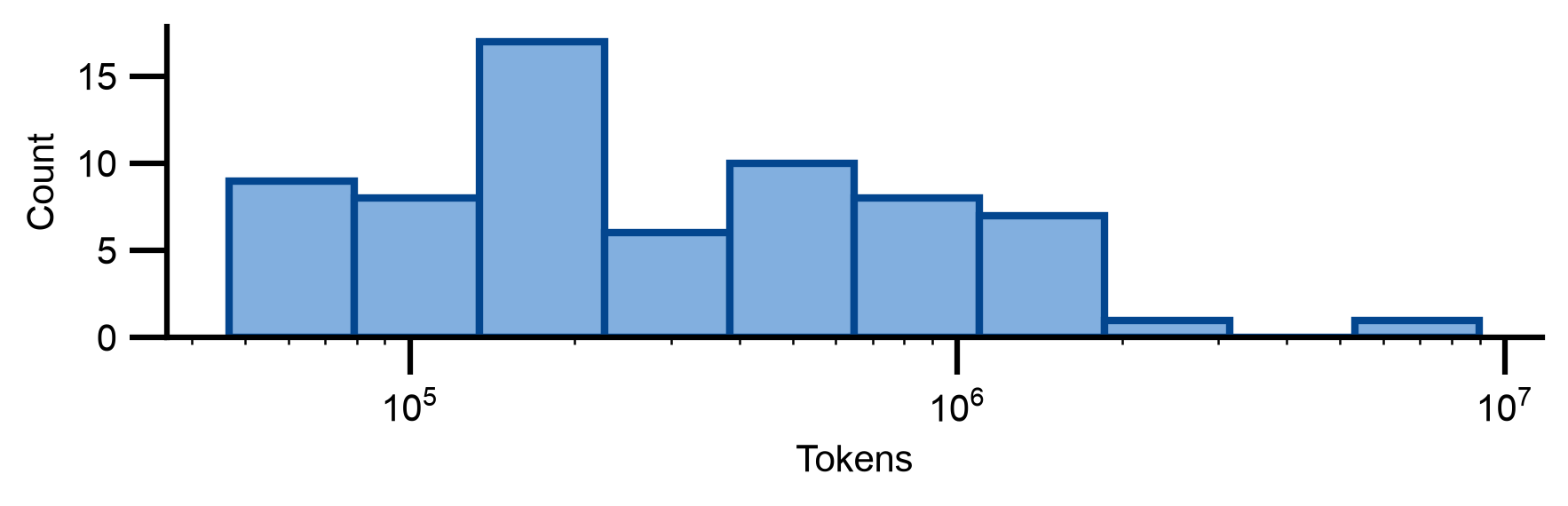}
    \caption{Number of tokens per patient}
    \label{sfig:patient-token}
\end{figure}
\begin{figure}
    \centering
    \includegraphics[width=\linewidth]{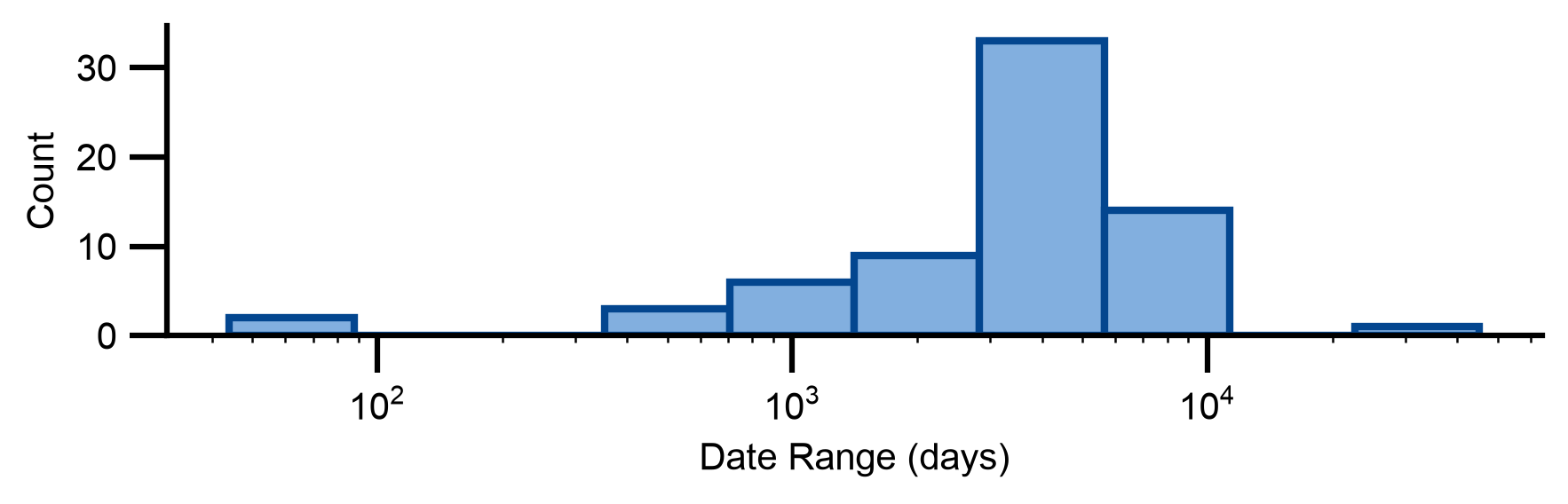}
    \caption{Number of days spanning longitudinal records}
    \label{sfig:patient-daterange}
\end{figure}


\begin{table}[t]
\centering
\begin{tabular}{l r r r}
\toprule
\textbf{Filtering stage} & \textbf{Remaining} & \textbf{Removed} & \textbf{\% of original} \\
\midrule
Original                          & 750 & --- & 100.0 \\
Two annotators                    & 675 & 75  & 90.0 \\
Consistent                        & 660 & 15  & 88.0 \\
Relevant questions                & 550 & 110 & 73.3 \\
\textbf{One annotator accepted the answer} & \textbf{508} & \textbf{42} & \textbf{67.7} \\
\midrule
Both annotators accepted the answer & 308 & 200 & 41.1 \\
\bottomrule
\end{tabular}
\caption{Consecutive filtering of generated questions. BRIE is
indexed on the one-annotator-accepted set (\textbf{508}); the number of questons where both annotators accepted the answer is reported for reference.}
\label{stab:filtering}
\end{table}

\begin{table}[t]
\centering

\begin{tabular}{l r r r}
\toprule
\textbf{Model} & \textbf{Official context} & \textbf{Buffer} & \textbf{Tiktoken limit} \\
\midrule
Gemini 2.5 Pro        & 1{,}000{,}000 & 50{,}000 & 950{,}000 \\
Gemini 2.5 Flash Lite & 1{,}000{,}000 & 50{,}000 & 950{,}000 \\
Claude Opus 4.7       & 1{,}000{,}000 & 75{,}000 & 925{,}000 \\
Claude Haiku 4.5      &   200{,}000   & 50{,}000 & 150{,}000 \\
GPT 5.4               &   272{,}000   & 10{,}000 & 262{,}000 \\
GPT 5.4 Nano          &   272{,}000   & 10{,}000 & 262{,}000 \\
Kimi K2.6             &   262{,}144   & 50{,}000 & 212{,}144 \\
Qwen 3.5 397B         &   253{,}953   & 30{,}000 & 223{,}953 \\
Qwen 3.5 27B          &   253{,}953   & 30{,}000 & 223{,}953 \\
\bottomrule
\end{tabular}
\caption{Model context-window limits. The effective limit subtracts a buffer from
the official context window to account for tokenizer (tiktoken) counting
discrepancies and prompt boiler plates.}
\label{stab:context-limits}
\end{table}

\begin{table}[t]
\centering

\begin{tabular}{l r r r r r r r}
\toprule
\textbf{Model} & \textbf{Mean} & \textbf{SD} & \textbf{Min} & \textbf{25\%} & \textbf{Median} & \textbf{75\%} & \textbf{Max} \\
\midrule
Claude Opus  & 4.17 & 3.31 & 0 & 2 & 3 & 5 & 25 \\
Claude Haiku & 6.18 & 4.97 & 0 & 2 & 4 & 9 & 35 \\
\bottomrule
\end{tabular}
\caption{Number of agent steps per query, by model}
\label{stab:agent-steps}
\end{table}

\begin{table}[t]
\centering
\begin{tabular}{l l r r r r r r r}
\toprule
\textbf{Model} & \textbf{Tool} & \textbf{Mean} & \textbf{SD} & \textbf{Min} & \textbf{25\%} & \textbf{Median} & \textbf{75\%} & \textbf{Max} \\
\midrule
\multirow{3}{*}{Claude Opus}
  & \texttt{search\_notes}    & 0.57 & 0.13 & 0.00 & 0.50 & 0.50 & 0.67 & 0.93 \\
  & \texttt{get\_note}        & 0.34 & 0.20 & 0.00 & 0.17 & 0.50 & 0.50 & 0.83 \\
  & \texttt{summarize\_notes} & 0.09 & 0.17 & 0.00 & 0.00 & 0.00 & 0.00 & 0.67 \\
\cmidrule(lr){1-9}
\multirow{3}{*}{Claude Haiku}
  & \texttt{search\_notes}    & 0.57 & 0.17 & 0.00 & 0.50 & 0.50 & 0.67 & 1.00 \\
  & \texttt{get\_note}        & 0.42 & 0.17 & 0.00 & 0.33 & 0.50 & 0.50 & 0.86 \\
  & \texttt{summarize\_notes} & 0.01 & 0.05 & 0.00 & 0.00 & 0.00 & 0.00 & 0.50 \\
\bottomrule
\end{tabular}
\caption{Tool-usage fraction per query, by model. Values are the fraction of each query's tool calls allocated to a given tool.}
\label{stab:agent-usage}
\end{table}

\begin{figure}[p]
    \centering
    \begin{tcolorbox}[
enhanced, colback=backblue, colframe=frameblue, coltitle=white,
  title=\textbf{Prompt: Fact Entailment (Recall)},
  fonttitle=\bfseries, fontupper=\small,
]
\textbf{System.} You are a clinician performing chart review. Respond only with a
valid JSON array.

\smallskip
\textbf{Recall direction} (which reference facts are covered).
Given a list of REFERENCE facts and a list of CANDIDATE facts, identify which REFERENCE facts are semantically entailed by any of the CANDIDATE facts. Return [] if none are entailed. Judge as a clinician reviewing the chart would, not as a literal string matcher.

\smallskip
\textbf{Rules.}
\begin{enumerate}[leftmargin=*, itemsep=0pt, topsep=0pt, parsep=0pt]
\item A REFERENCE fact is entailed if the CANDIDATE facts assert it --- it need not be restated by a single CANDIDATE fact.
\item Both lists are atomized, so a dated event is split into a bare event anchor (``A hemodynamic measurement was performed on 2021-10-16.'') plus separate date-free facts giving that event's details, findings, or results. Judge an anchor by its details, not by its date: the anchor is entailed whenever the CANDIDATE facts assert those details.
\item Two events belong to the same episode of care when their dates match, fall within about two weeks of each other, or one is given only as a month or an approximate date. Do not require an exact date match.
\item Within one episode of care, a REFERENCE fact describing a component, step, or routine part of a larger event is entailed by a CANDIDATE fact describing that larger event: an encounter entails the medications, fluids, and assessments given during it, and a procedure entails the measurements and specimens it ordinarily involves.
\item Wording and granularity need not match. A fact may be more specific in one respect (naming the drug, device, or site) and less specific in another (a month rather than a day); neither difference blocks entailment, in either direction.
\item Entail on semantic equivalence or logical implication, not only on
restatement. If a CANDIDATE fact means the same thing in different words, or logically implies the REFERENCE fact, mark it entailed --- a stated consequence of a symptom implies the symptom, resuming or restarting a treatment implies that it was initiated, and a documented trial of a treatment implies that it was given.
\item One supporting CANDIDATE fact is enough. Judge each REFERENCE fact on its own, and do not withhold entailment because other CANDIDATE facts describe related events pointing a different way: a separate or later event involving the same medication, problem, or procedure does not cancel an earlier one.
\item Only two things block entailment: the CANDIDATE facts describe no related event or encounter at all, or a CANDIDATE fact directly denies the REFERENCE fact itself. A treatment tried without benefit still entails that it was given, but does not entail that it helped.
\end{enumerate}
\tcblower
\footnotesize\itshape Facts are supplied as 0-indexed numbered lists, optionally preceded by direction-matched few-shot examples. The precision direction uses the same rules with the roles of the two lists exchanged, returning indices of the entailed CANDIDATE facts. Recall $=$ \mbox{$|$entailed ref$|/|$ref$|$}; precision
$=$ \mbox{$|$entailed cand$|/|$cand$|$}.
\end{tcolorbox}
    \caption{Condense, optimized fact entailment prompt for recall. The precision prompt uses similar rules but the REFERENCE and CANDIDATE facts lists are flipped.}
    \label{sfig:entailment}
\end{figure}

\begin{figure}
    \centering
    \includegraphics[width=\linewidth]{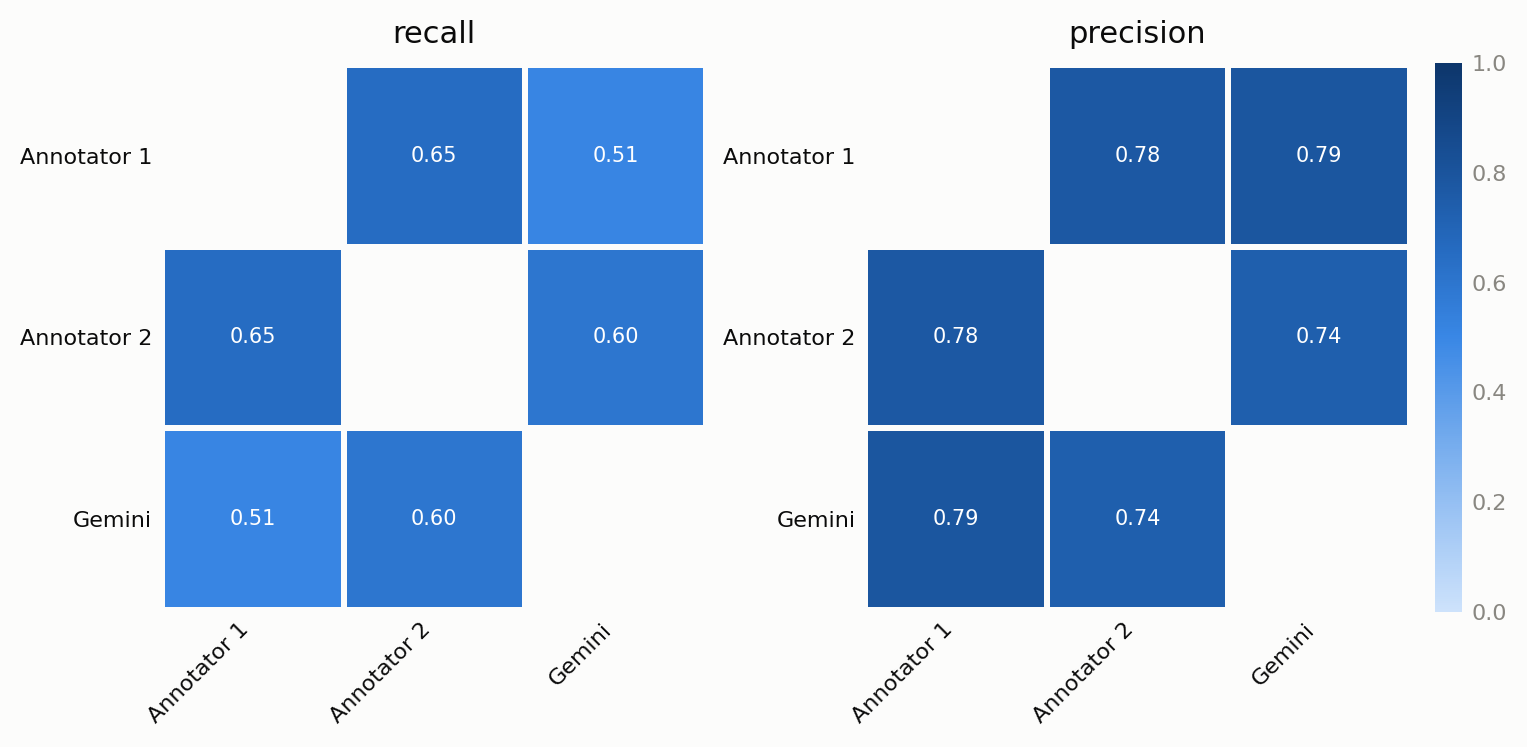}
    \caption{Cohen's Kappa {$\kappa$} between optimized prompt and human raters for 50 randomly sampled responses for fact entailment. Gemini-Flash-Lite 3.1 was used entailment.}
    \label{sfig:heatmap-kappa-fact}
\end{figure}

\begin{figure}[p]
    \centering
   \begin{tcolorbox}[
  unbreakable, enhanced, colback=backblue, colframe=frameblue, coltitle=white,
  title=\textbf{Prompt: Win-rate (Pairwise Judge)},
  fonttitle=\bfseries, fontupper=\small, left=4pt, right=4pt,
  top=3pt, bottom=3pt
]

\textbf{System.} You are a medical expert. Respond only with a valid JSON object.

\smallskip
\textbf{Instruction.} You are a medical expert comparing two responses to a
clinical information retrieval query. Given a reference answer (gold standard) and
two candidate responses (A and B), decide which response is better on each
dimension, or declare a tie.

\smallskip
\textbf{Inputs.}
\begin{verbatim}
Question:
<question>{QUESTION}</question>

Reference answer:
<reference>{REFERENCE}</reference>

Response A:
<response_a>{RESPONSE_A}</response_a>

Response B:
<response_b>{RESPONSE_B}</response_b>
\end{verbatim}

\textbf{Dimensions.}
\begin{itemize}[leftmargin=1.2em, itemsep=1pt, topsep=2pt]
  \item \textbf{Completeness:} which response includes more of the important
    clinical details present in the reference answer? Prefer the response that
    omits fewer key facts.
  \item \textbf{Relevancy:} which response stays closer to what the question asks
    and the reference answer covers, without introducing unnecessary or tangential
    details?
  \item \textbf{Concision:} which response communicates the necessary information
    more concisely, without excessive verbosity or redundant phrasing?
\end{itemize}

For each dimension, output \texttt{"A"}, \texttt{"B"}, or \texttt{"tie"}.

\smallskip
\textbf{Output format.}
\begin{verbatim}
{
    "completeness": {"winner": "A" | "B" | "tie", "explanation": "..."},
    "relevancy":    {"winner": "A" | "B" | "tie", "explanation": "..."},
    "concision":    {"winner": "A" | "B" | "tie", "explanation": "..."}
}
\end{verbatim}
Ensure the output is valid JSON with double quotes for all keys and string values.

\end{tcolorbox}
    \caption{Condensed win-rate prompt. The win rate is computed twice for each pair, with the ordering reversed in the second inference.}
    \label{sfig:winrate}
\end{figure}

\begin{figure}
    \centering
    \includegraphics[width=\linewidth]{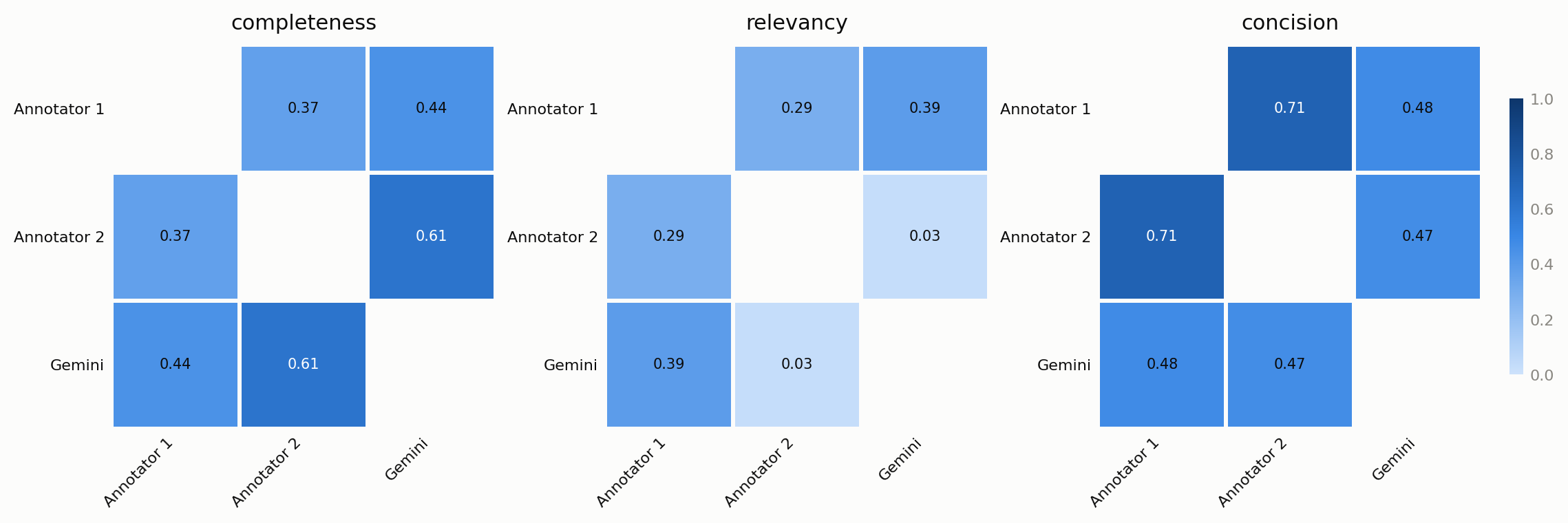}
    \caption{Cohen's Kappa {$\kappa$} between automated pairwise evaluation and human raters for 50 randomly sampled pairs. Gemini-Flash-Lite 3.1 was used automatd evaluation.}
    \label{sfig:heatmap-kappa-elo}
\end{figure}

\begin{figure}[p]
    \centering
    \begin{tcolorbox}[
  unbreakable, enhanced, colback=backblue, colframe=frameblue, coltitle=white,
  title=\textbf{Prompt: Hallucination Detection (Fact Grounding)},
  fonttitle=\bfseries, fontupper=\small, left=4pt, right=4pt,
  top=3pt, bottom=3pt
]

\textbf{System.} You are a clinician performing chart review. Respond only with a
valid JSON array.

\smallskip
\textbf{Inputs} (prompt built per note).
\begin{verbatim}
Note (Date: {note_date}, Title: {note_title}):
{note_text}

Facts to check:
1. {fact}
2. {fact}
... (1-indexed)
\end{verbatim}

\textbf{Instruction.} For each fact, determine whether this note:
\begin{itemize}[leftmargin=1.2em, itemsep=1pt, topsep=2pt]
  \item \texttt{"supported"}: explicitly documents the same finding, event, or
    measurement.
  \item \texttt{"contradicted"}: explicitly states something that conflicts with
    the fact (e.g., denies the finding, records a conflicting value or different
    date).
\end{itemize}

Omit facts where the note provides no relevant evidence (\texttt{not\_supported}).

\smallskip
\textbf{Output.} Return a JSON array --- only for facts with clear supported or
contradicted evidence:
\begin{verbatim}
[{"fact_idx": <1-N>, "verdict": "supported"|"contradicted",
  "evidence": "<shortest exact substring from the note>"}]
\end{verbatim}

\tcblower
\footnotesize\itshape Each atomic fact of an answer is checked against the
patient's notes. A fact never supported by any note is the hallucination signal.
\end{tcolorbox}
    \caption{Condense hallucination detection prompt}
    \label{sfig:hallucination}
\end{figure}


\begin{table}[t]
\centering

\begin{tabular}{l c c}
\toprule
\textbf{Inference type} & \textbf{Rate} & \textbf{95\% CI} \\
\midrule
Recent       & 0.99 & [0.99, 1.00] \\
Recent-200K  & 0.99 & [0.98, 1.00] \\
Dense        & 0.99 & [0.97, 1.00] \\
BM25         & 0.99 & [0.99, 1.00] \\
Agent        & 0.99 & [0.94, 1.00] \\
\bottomrule
\end{tabular}
\caption{Percentage of grounded facts by inference type, with 95\% confidence intervals.}
\label{stab:hallucination}
\end{table}


\begin{table}[t]
\centering
\small
\begin{tabular}{l r r r r r r}
\toprule
& \multicolumn{2}{c}{\textbf{Price (\$/1M)}} & \multicolumn{4}{c}{\textbf{Cost per run (\$)}} \\
\cmidrule(lr){2-3} \cmidrule(lr){4-7}
\textbf{Model} & \textbf{In} & \textbf{Out} & \textbf{Full ctx} & \textbf{180K} & \textbf{50K} & \textbf{25K} \\
\midrule
Claude Haiku 4.5      & 1.00  & 5.00  & 0.2013 & 0.1813 & 0.0513 & 0.0263 \\
Claude Opus 4.7       & 5.00  & 25.00 & 5.0013 & 0.9013 & 0.2513 & 0.1263 \\
\addlinespace
Gemini Flash Lite 2.5 & 0.10  & 0.30  & 0.1013 & 0.0193 & 0.0063 & --- \\
Gemini Pro 2.5        & 2.50  & 15.00 & 2.5013 & 0.4513 & 0.1263 & --- \\
\addlinespace
GPT-5.4               & 5.00  & 22.50 & 1.3613 & 0.9013 & 0.2513 & --- \\
GPT-nano 5.4          & 0.20  & 1.25  & 0.0557 & 0.0373 & 0.0113 & --- \\
\addlinespace
Kimi K2.6             & 0.95  & 4.00  & 0.2503 & 0.1723 & 0.0488 & --- \\
Qwen 3.5 397B         & 0.385 & 2.45  & 0.0990 & 0.0706 & 0.0205 & --- \\
Qwen 3.5 27B          & 0.195 & 1.56  & 0.0508 & 0.0364 & 0.0110 & --- \\
\bottomrule
\end{tabular}
\caption{Estimated per-run inference cost (USD) by input context size. Price columns are per 1M tokens. The \emph{Full ctx} column uses each model's effective context window (Haiku 200K; Opus, Gemini Flash Lite, Gemini Pro 1M; GPT-5.4 /
nano 272K; Kimi 262K; Qwen 254K); the remaining columns use fixed input sizes. Each cost includes a fixed \$0.00125 output component; the 25K (agent) setting was run only for the Claude models.}
\label{stab:inference-cost}
\end{table}

\begin{figure}[p]
    \centering
   \begin{tcolorbox}[
  unbreakable, enhanced, colback=backblue, colframe=frameblue, coltitle=white,
  title=\textbf{Stage 1 --- Seed Generation},
  fonttitle=\bfseries, fontupper=\small, left=4pt, right=4pt, top=3pt, bottom=3pt
]

\textbf{Role.} Clinical informatics expert generating diverse question seeds for
a medical QA dataset.

\smallskip
\textbf{Task.} Given a natural clinical query and a reference QA pair, produce
8--10 distinct seeds, each a different plausible interpretation varying in
timeframe, topic focus, and detail level. Use the reference only as an
\emph{exclusion} reference --- do not replicate it.

\smallskip
\textbf{Inputs.} \texttt{\{natural\_query\}}; reference
\texttt{\{reference\_question\}} / \texttt{\{reference\_answer\}}.

\smallskip
\textbf{Axes of variation.}
\begin{itemize}[leftmargin=1.2em, itemsep=1pt, topsep=2pt]
  \item \textbf{Timeframe:} full history / a named episode or admission / most
    recent only / between two explicit dates or events.
  \item \textbf{Topic focus:} free-text clinical sub-topic (e.g.\ antibiotic
    susceptibilities, symptom onset, regimen and doses, labs, imaging). Do not
    force predefined categories.
  \item \textbf{Detail level:} \emph{minimum} (single fact/value) / \emph{concise}
    (key facts, no elaboration) / \emph{thorough} (full picture with context,
    modifiers, relevant negatives).
\end{itemize}

\textbf{Constraints.} Each seed must differ meaningfully from the reference and
from every other seed; maximize diversity across all three axes; exclude
implausible interpretations even if answerable.

\smallskip
\textbf{Output.} JSON array; each seed: \texttt{seed\_id}, \texttt{query\_focus},
\texttt{timeframe}, \texttt{topic}, \texttt{detail\_level}
(\texttt{"minimum"|"concise"|"thorough"}), \texttt{differs\_from\_reference\_by}.
\end{tcolorbox}
    \caption{Multiple Answer Generation (Stage 1/4). Condensed seed generation prompt.}
    \label{sfig:answer-seed}
\end{figure}

\begin{figure}[p]
    \centering
    \begin{tcolorbox}[
  unbreakable, enhanced, colback=backblue, colframe=frameblue, coltitle=white,
  title=\textbf{Stage 2 --- Timeframe Resolution},
  fonttitle=\bfseries, fontupper=\small, left=4pt, right=4pt, top=3pt, bottom=3pt
]

\textbf{Role.} Clinical informatics assistant resolving a seed's free-text
timeframe into a concrete date window, anchored to the patient's fact dates.

\smallskip
\textbf{Task.} Return inclusive \texttt{start\_date}/\texttt{end\_date}
(\texttt{YYYY-MM-DD}) bounding the timeframe. Query focus is for disambiguating
boundaries only --- do not filter by topic.

\smallskip
\textbf{Inputs.} \texttt{\{seed\_query\_focus\}}, \texttt{\{seed\_timeframe\}},
\texttt{\{fact\_list\}} (date context only).

\smallskip
\textbf{Interpretation rules.}
\begin{enumerate}[leftmargin=1.4em, itemsep=1pt, topsep=2pt]
  \item \textbf{Most recent only:} most recent fact relevant to the topic; set
    start $=$ end to that date. Anchor to the topic, not the global latest date.
  \item \textbf{Named episode/event:} earliest and latest dates of facts in that
    episode; use focus to disambiguate.
  \item \textbf{Relative window} (e.g.\ ``last 2 years''): anchor to the most
    recent fact date as end.
  \item \textbf{Between two events/dates:} the boundary dates (inclusive).
  \item \textbf{Full history:} earliest $\rightarrow$ most recent fact date.
\end{enumerate}

\textbf{Edge case.} If no facts exist or the timeframe is unresolvable, return
\texttt{null} for both dates and explain.

\smallskip
\textbf{Output.} \texttt{\{ "resolved\_timeframe": "...", "start\_date":
"YYYY-MM-DD or null", "end\_date": "YYYY-MM-DD or null" \}}
\end{tcolorbox}
    \caption{Multiple Answer Generation (Stage 2/4). Condensed time-frame identification prompt.}
    \label{sfig:answer-timeframe}
\end{figure}

\begin{figure}[p]
    \centering
    \begin{tcolorbox}[
  unbreakable, enhanced, colback=backblue, colframe=frameblue, coltitle=white,
  title=\textbf{Stage 3 --- QA Generation},
  fonttitle=\bfseries, fontupper=\small, left=4pt, right=4pt, top=3pt, bottom=3pt
]

\textbf{Role.} Clinical documentation specialist generating one QA pair grounded
strictly in the provided facts, returning the verbatim facts used.

\smallskip
\textbf{Inputs.} \texttt{\{natural\_query\}} (context only); seed (Focus /
Timeframe / Topic / Detail); \texttt{\{persona\}}; \texttt{\{fact\_list\}}.

\smallskip
\textbf{Instructions.}
\begin{enumerate}[leftmargin=1.4em, itemsep=1pt, topsep=2pt]
  \item Write a specific question this persona would plausibly ask, consistent
    with the seed and natural query --- precise about timeframe and topical scope.
  \item The question must specify all expected return items (clear what would be
    ``missing'').
  \item Write a precise answer grounded only in the facts: do not infer or
    generalize; reflect incompleteness; note absence of information when relevant;
    stay within scope.
  \item Return the verbatim facts used.
  \item If the seed cannot be fulfilled (facts absent/ambiguous/contradictory),
    return empty output.
\end{enumerate}

\textbf{Output.} Fulfillable: \texttt{\{ "question": "...", "answer": "...",
"supporting\_facts": [...] \}}; \quad not fulfillable: \texttt{\{ "question":
null, "answer": null, "supporting\_facts": [] \}}
\end{tcolorbox}
    \caption{Multiple Answer Generation (Stage 3/4). Condensed fact grounding prompt.}
    \label{sfig:answer-grounding}
\end{figure}

\begin{figure}[p]
    \centering
    \begin{tcolorbox}[
  unbreakable, enhanced, colback=backblue, colframe=frameblue, coltitle=white,
  title=\textbf{Stage 4 --- De-duplication},
  fonttitle=\bfseries, fontupper=\small, left=4pt, right=4pt, top=3pt, bottom=3pt
]

\textbf{Role.} Clinical QA dataset curator collapsing generated QA pairs to the
2--5 most distinct, coherent interpretations of the natural query.

\smallskip
\textbf{Inputs.} \texttt{\{natural\_query\}}; reference
\texttt{\{reference\_question\}} / \texttt{\{reference\_answer\}};
\texttt{\{qa\_pairs\}}.

\smallskip
\textbf{Filtering criteria.}
\begin{enumerate}[leftmargin=1.4em, itemsep=1pt, topsep=2pt]
  \item \textbf{Coherence:} a plausible interpretation of the query? Reject if not
    what the clinician could reasonably have meant.
  \item \textbf{Redundancy:} substantial overlap with another pair or the
    reference (same facts/info, even if reworded). Drop rephrasings of the
    reference; prefer the more natural question with the precise answer.
  \item \textbf{Specificity:} specific enough to have a determinate answer? Reject
    if multiple different answers could be correct.
\end{enumerate}

\textbf{Output.} \texttt{\{ "retained": [\{ "qa\_id", "question", "answer",
"supporting\_facts" \}], "removed": [\{ "qa\_id",
"reason": "coherence|redundancy|specificity", "explanation" \}] \}}
\end{tcolorbox}
    \caption{Multiple answer generation (Stage 4/4). Condensed de-duplication prompt.}
    \label{sfig:answer-dedup}
\end{figure}

\begin{figure}
    \centering
    \includegraphics[width=\linewidth]{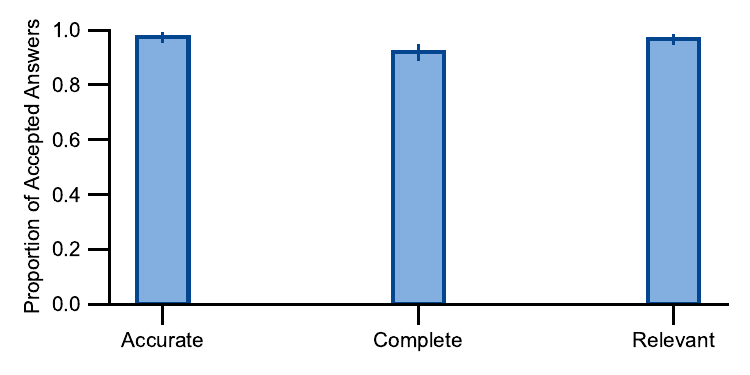}
    \caption{Multiple answer validation results}
    \label{sfig:answer-validation}
\end{figure}
\begin{figure}
    \centering
    \includegraphics[width=\linewidth]{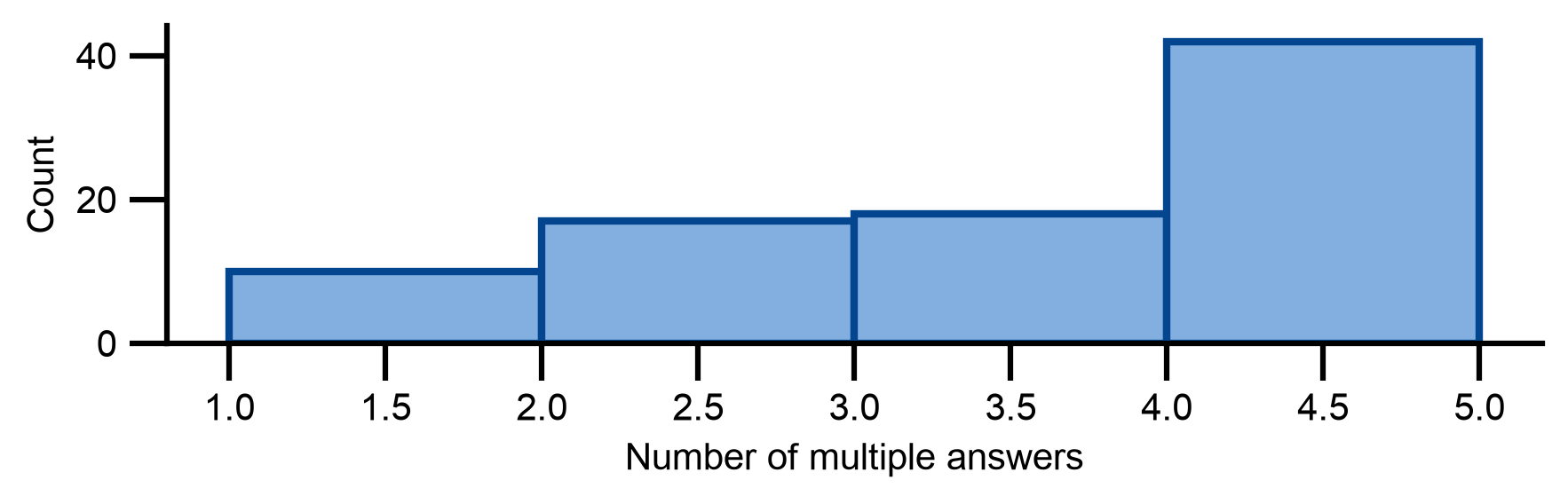}
    \caption{Number of accepted answers per accepted BRIE query}
    \label{sfig:answer-count}
\end{figure}

\input{SI/FIG/tables/questiontype_count}
\input{SI/FIG/tables/fact_performance}

\input{SI/FIG/tables/questiontype_mwu}

\begin{figure}
    \centering
    \includegraphics[width=0.49\linewidth]{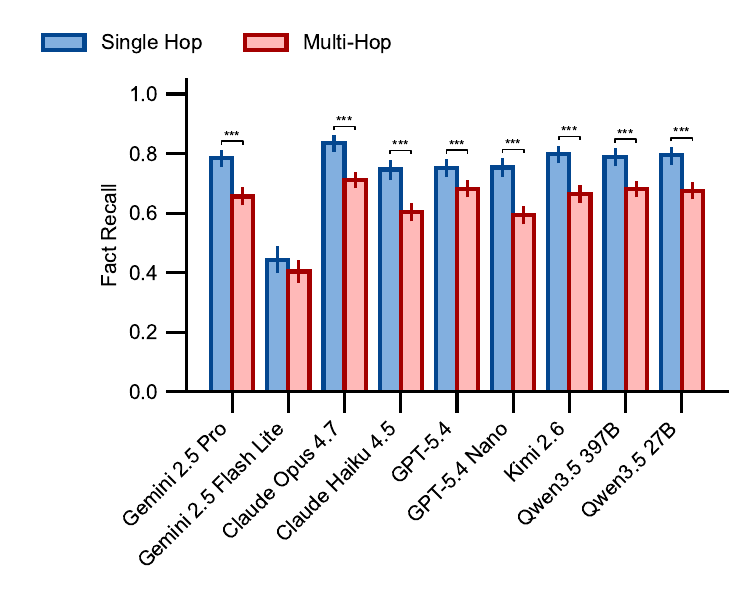}\hfill
    \includegraphics[width=0.49\linewidth]{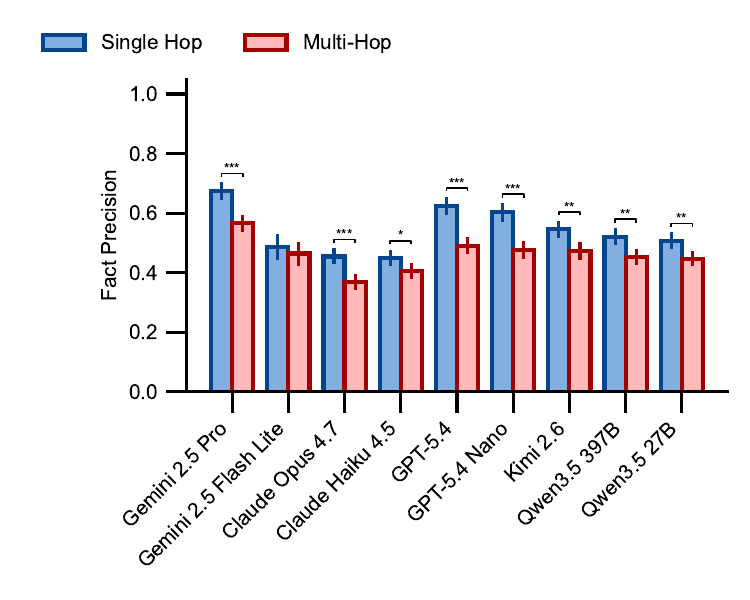}
    \caption{Fact Recall and Precision for Multi- and Single Hop questions for \textit{Recent} Inference. Significance thresholds: *** p $<$ 0.001; ** p $<$ 0.01; * p $<$ 0.05; ns = not significant (p $\geq$ 0.05)}
    \label{sfig:recall-precision-reasoning-recent}
\end{figure}
\begin{figure}
    \centering
    \includegraphics[width=0.49\linewidth]{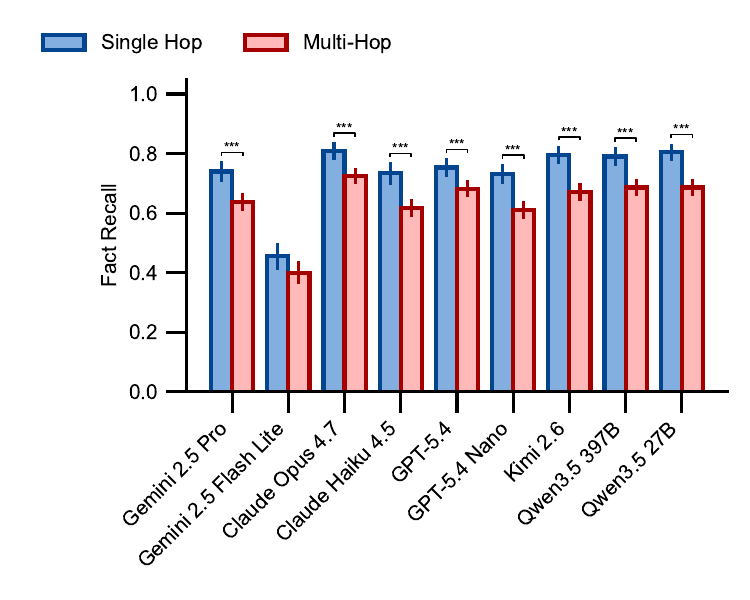}\hfill
    \includegraphics[width=0.49\linewidth]{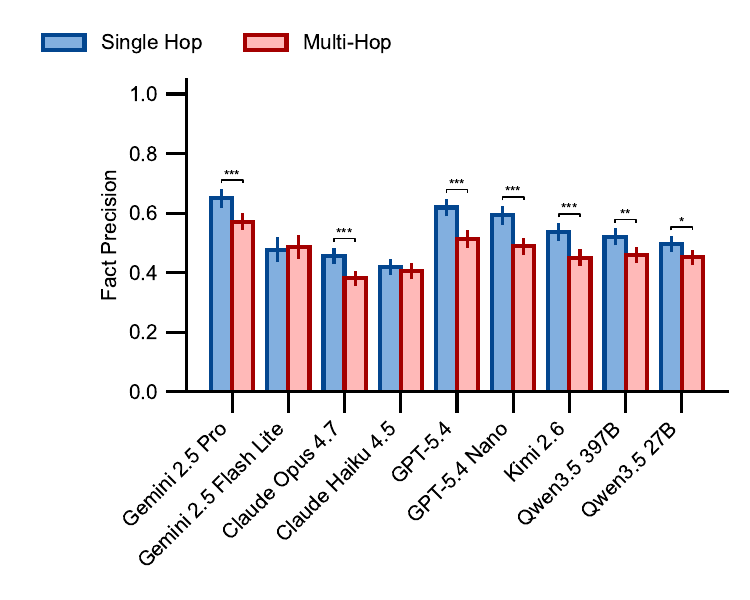}
    \caption{Fact recall and precision for multi- and single-hop questions under \textit{Recent-180K} inference. Significance thresholds: *** $p < 0.001$; ** $p < 0.01$; * $p < 0.05$; ns $=$ not significant ($p \geq 0.05$).}
    \label{sfig:recall-precision-reasoning-recent200}
\end{figure}
\begin{figure}
    \centering
    \includegraphics[width=0.49\linewidth]{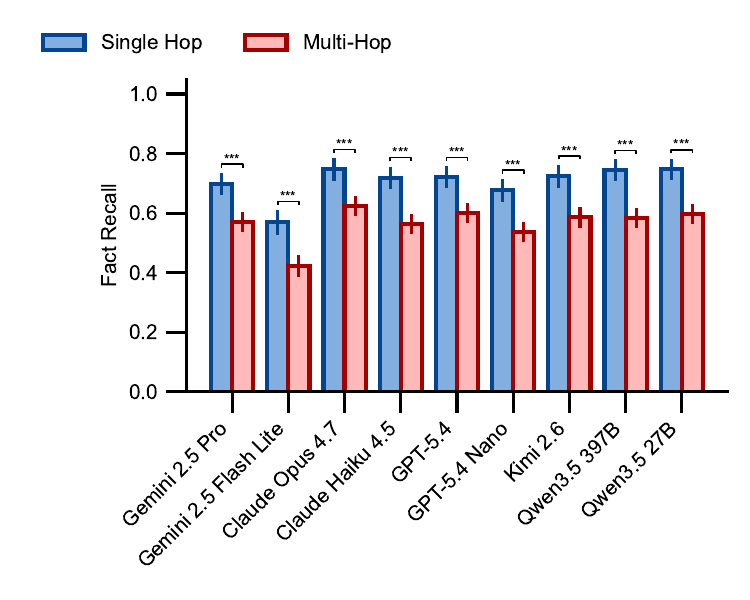}\hfill
    \includegraphics[width=0.49\linewidth]{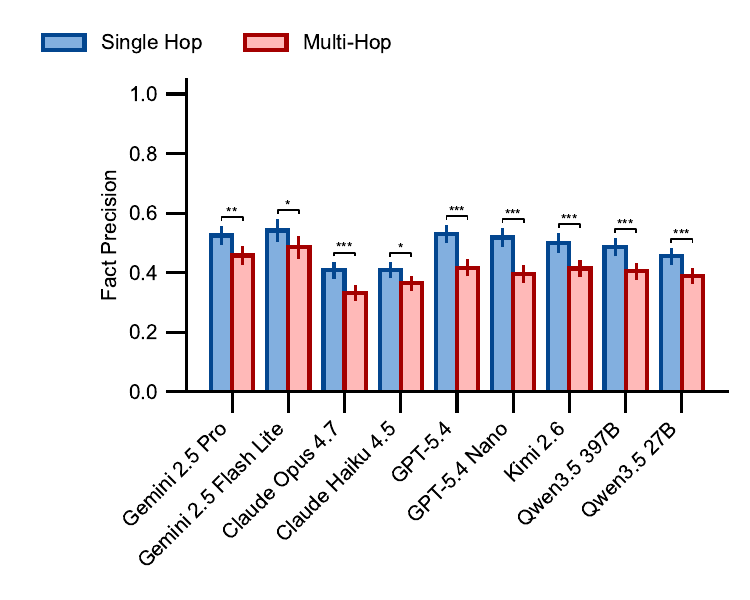}
    \caption{Fact recall and precision for multi- and single-hop questions under \textit{BM25} inference. Significance thresholds: *** $p < 0.001$; ** $p < 0.01$; * $p < 0.05$; ns $=$ not significant ($p \geq 0.05$).}
    \label{sfig:recall-precision-reasoning-bm25}
\end{figure}
\begin{figure}
    \centering
    \includegraphics[width=0.49\linewidth]{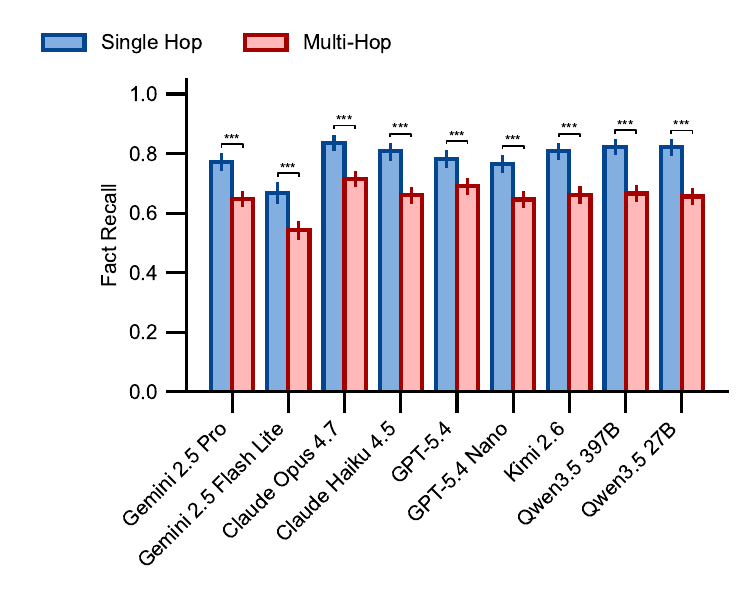}\hfill
    \includegraphics[width=0.49\linewidth]{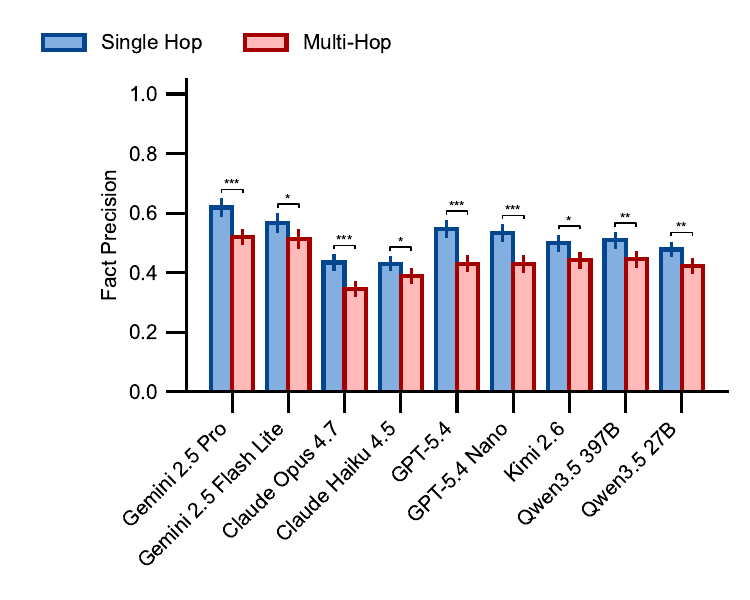}
    \caption{Fact recall and precision for multi- and single-hop questions under \textit{Dense} inference. Significance thresholds: *** $p < 0.001$; ** $p < 0.01$; * $p < 0.05$; ns $=$ not significant ($p \geq 0.05$).}
    \label{sfig:recall-precision-reasoning-dense}
\end{figure}
\begin{figure}
    \centering
    \includegraphics[width=0.49\linewidth]{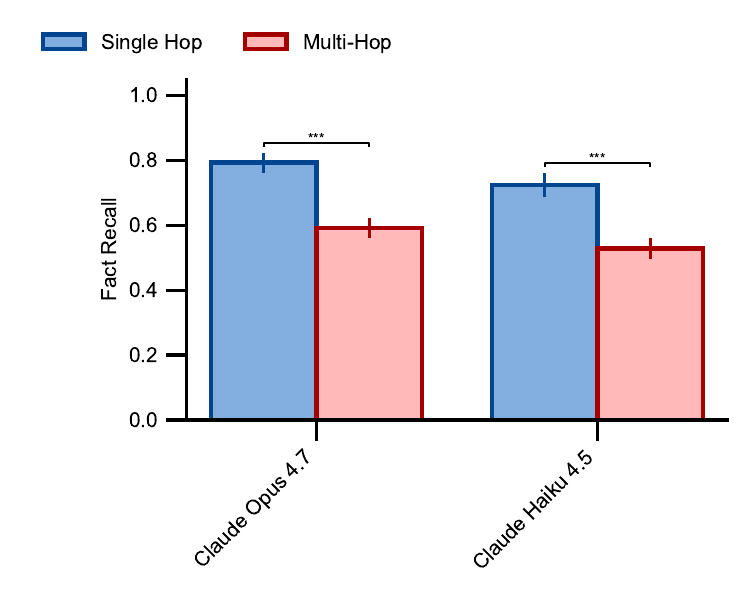}\hfill
    \includegraphics[width=0.49\linewidth]{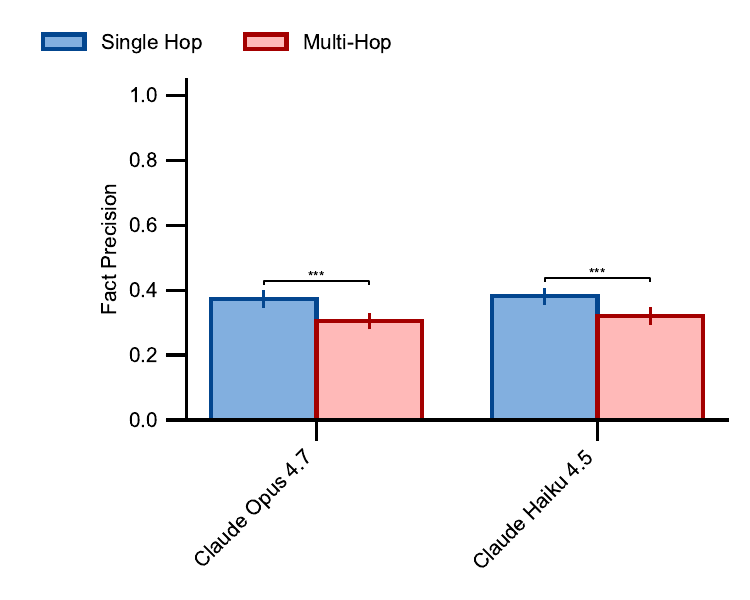}
    \caption{Fact recall and precision for multi- and single-hop questions under \textit{Agent} inference. Significance thresholds: *** $p < 0.001$; ** $p < 0.01$; * $p < 0.05$; ns $=$ not significant ($p \geq 0.05$).}
    \label{sfig:recall-precision-reasoning-agent}
\end{figure}

\begin{figure}
    \centering
    \includegraphics[width=0.49\linewidth]{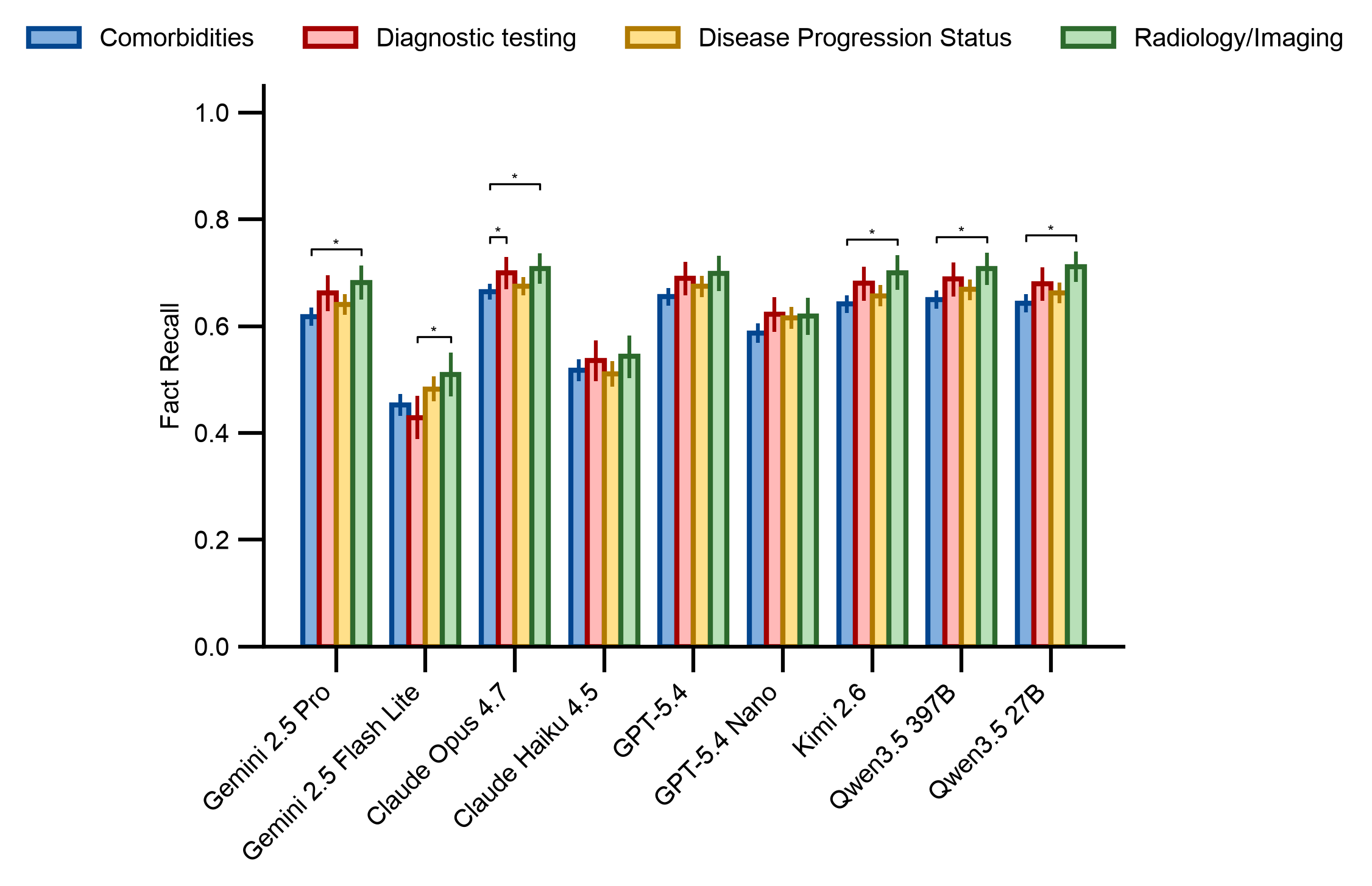}\hfill
    \includegraphics[width=0.49\linewidth]{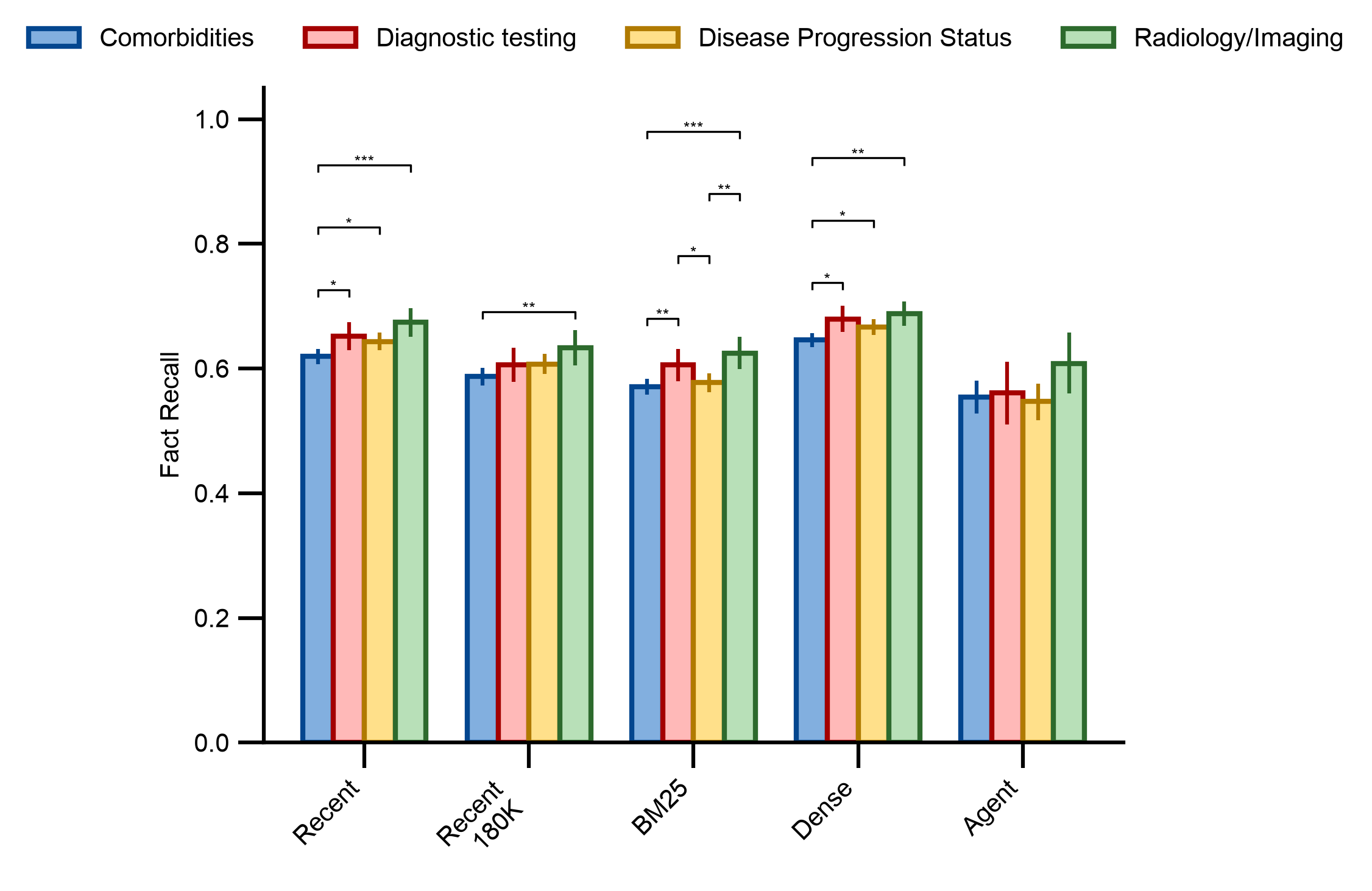}
    \caption{Fact recall for questions related to Comorbidities, Diagnostic testing, Disease Progression Status, and Radiology/Imaging that require multi-hop reasoning across Model and Inference types. Significance thresholds: *** $p < 0.001$; ** $p < 0.01$; * $p < 0.05$; ns $=$ not significant ($p \geq 0.05$).}
    \label{sfig:topic-multi}
\end{figure}
\begin{figure}
    \centering
    \includegraphics[width=0.49\linewidth]{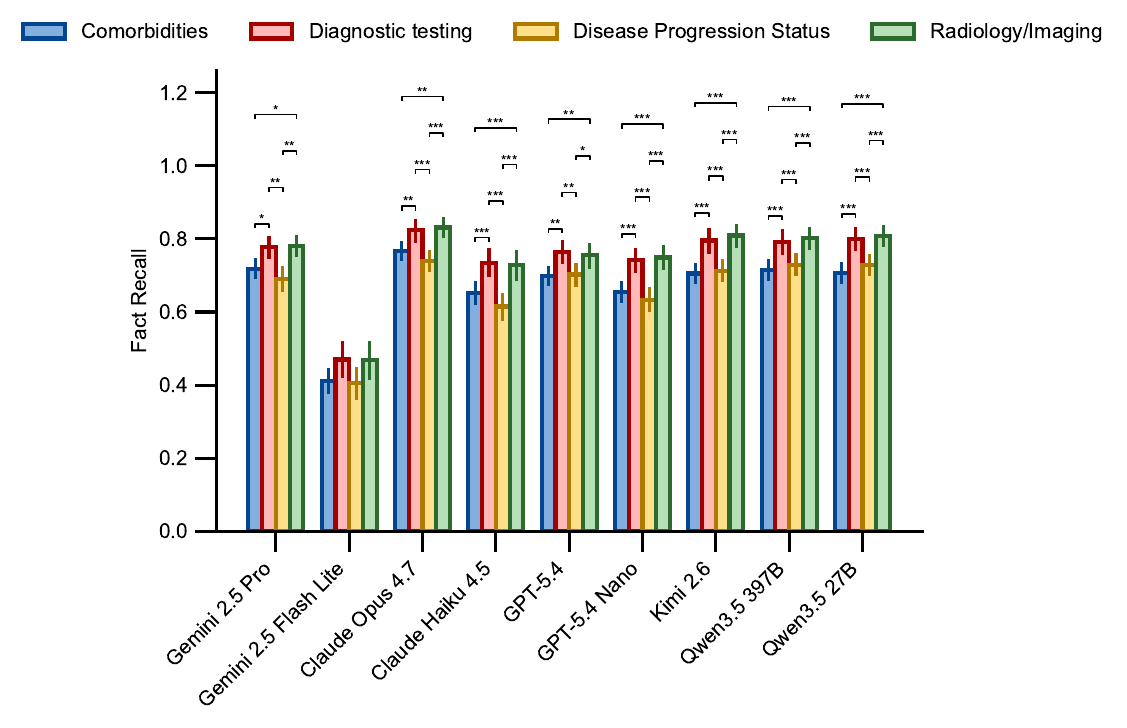}\hfill
    \includegraphics[width=0.49\linewidth]{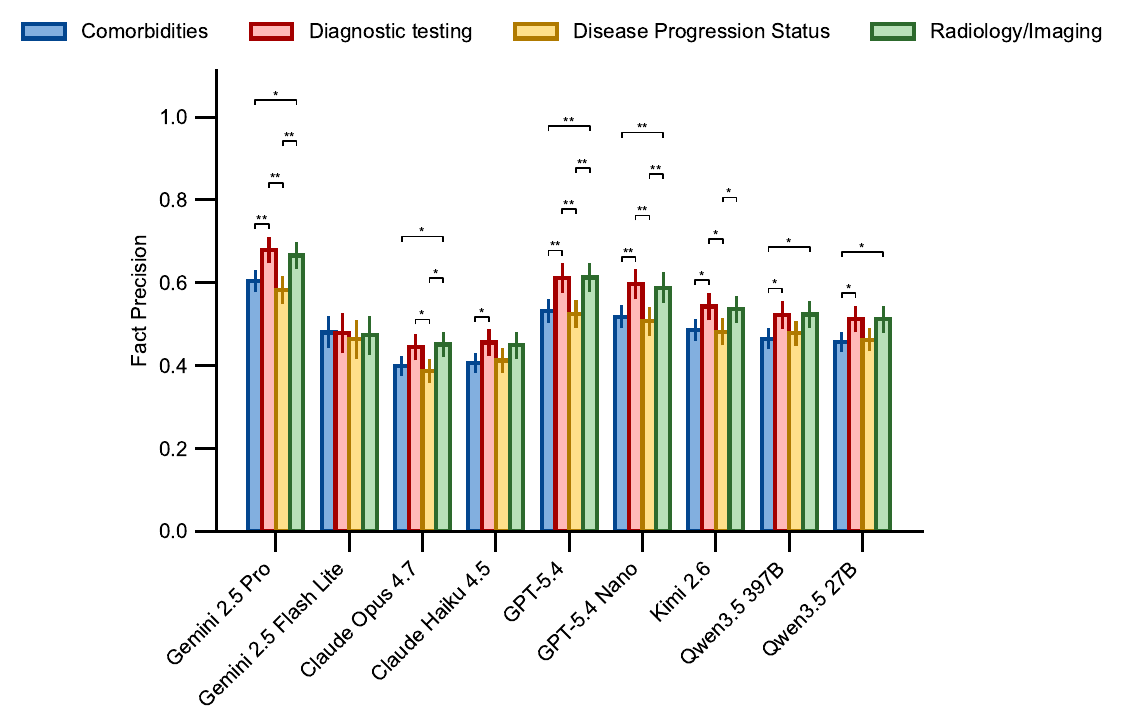}
    \caption{Fact recall and precision for questions related to Comorbidities, Diagnostic testing, Disease Progression Status, and Radiology/Imaging under \textit{Recent} inference. Significance thresholds: *** $p < 0.001$; ** $p < 0.01$; * $p < 0.05$; ns $=$ not significant ($p \geq 0.05$).}
    \label{sfig:recall-precision-topic-recent}
\end{figure}
\begin{figure}
    \centering
    \includegraphics[width=0.49\linewidth]{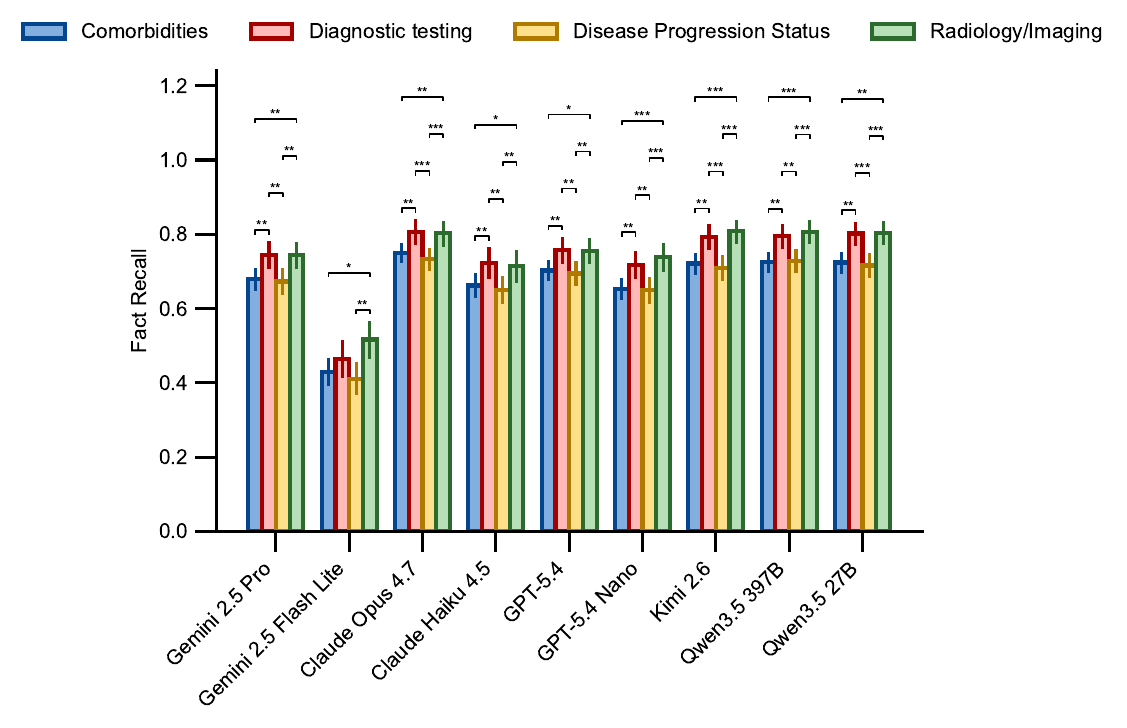}\hfill
    \includegraphics[width=0.49\linewidth]{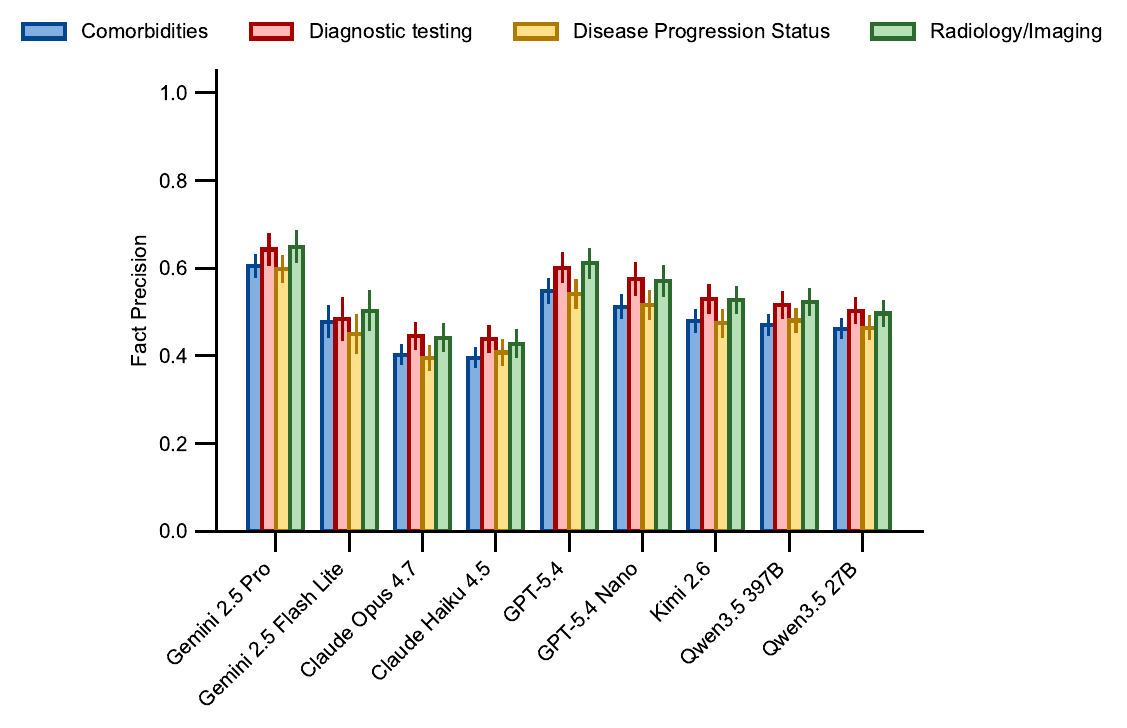}
    \caption{Fact recall and precision for questions related to Comorbidities, Diagnostic testing, Disease Progression Status, and Radiology/Imaging under \textit{Recent-180K} inference. Significance thresholds: *** $p < 0.001$; ** $p < 0.01$; * $p < 0.05$; ns $=$ not significant ($p \geq 0.05$).}
    \label{sfig:recall-precision-topic-recent200}
\end{figure}
\begin{figure}
    \centering
    \includegraphics[width=0.49\linewidth]{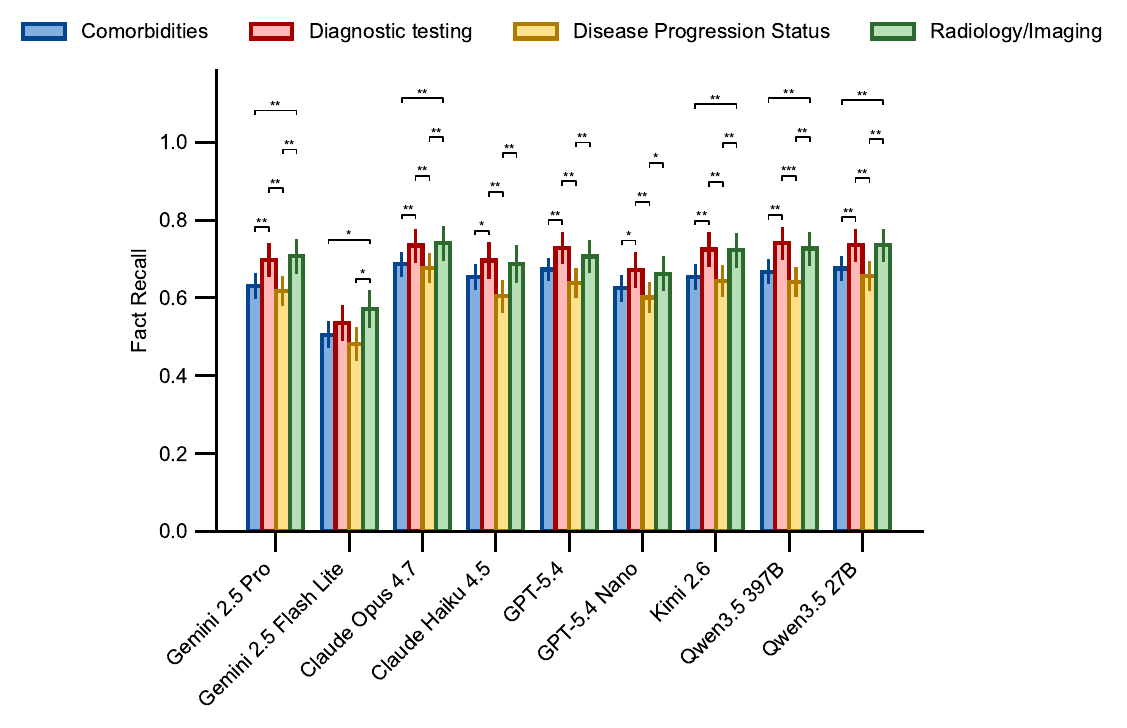}\hfill
    \includegraphics[width=0.49\linewidth]{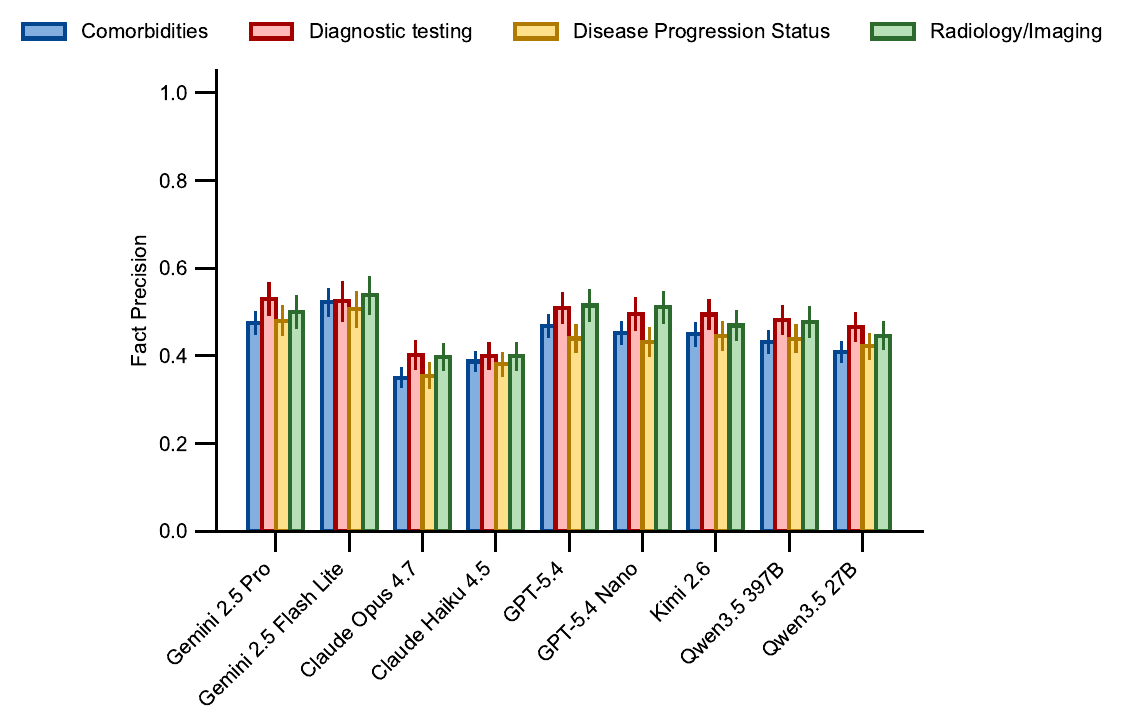}
    \caption{Fact recall and precision for questions related to Comorbidities, Diagnostic testing, Disease Progression Status, and Radiology/Imaging under \textit{BM25} inference. Significance thresholds: *** $p < 0.001$; ** $p < 0.01$; * $p < 0.05$; ns $=$ not significant ($p \geq 0.05$).}
    \label{sfig:recall-precision-topic-bm25}
\end{figure}
\begin{figure}
    \centering
    \includegraphics[width=0.49\linewidth]{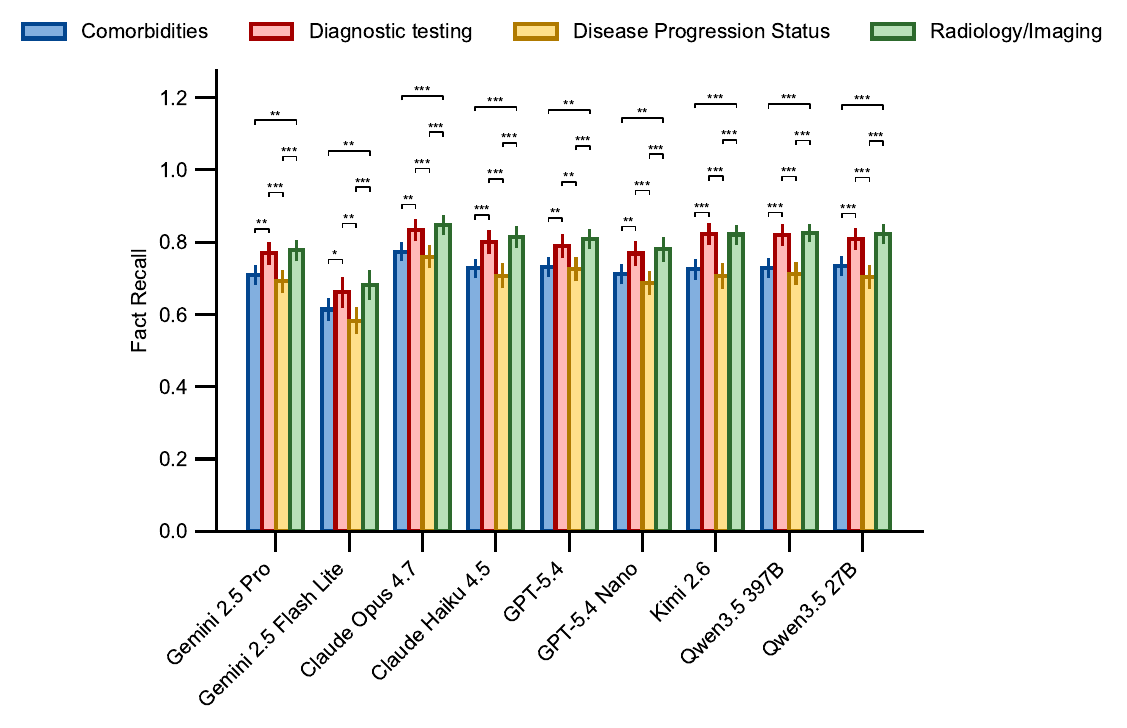}\hfill
    \includegraphics[width=0.49\linewidth]{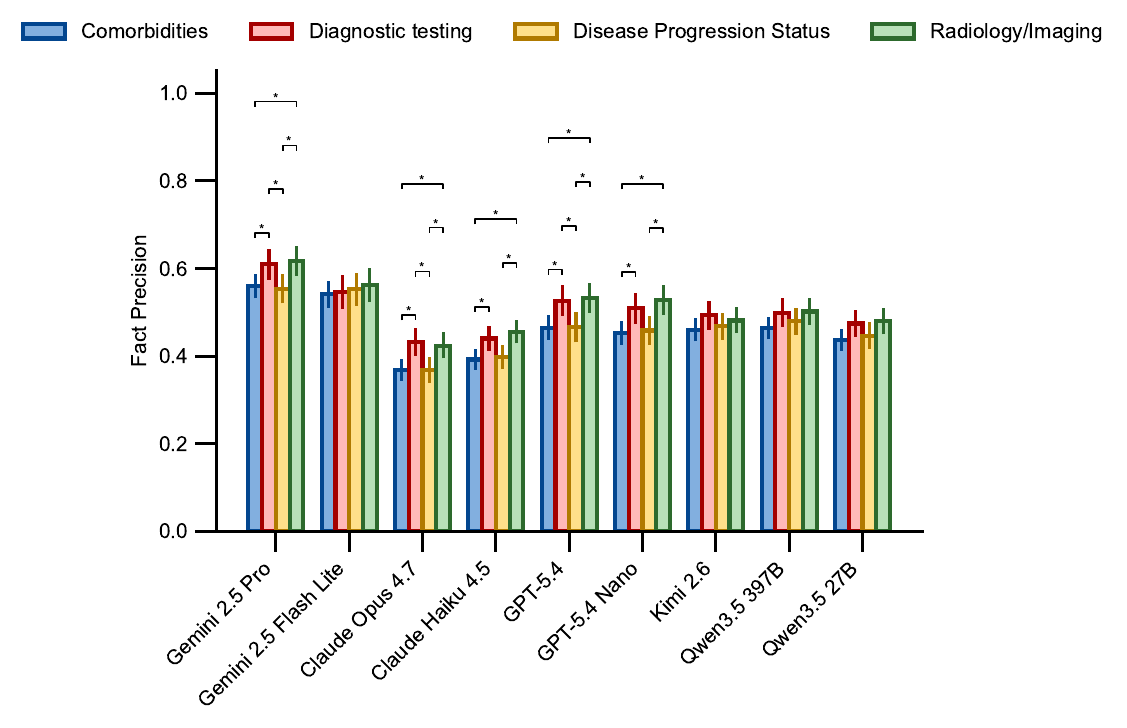}
    \caption{Fact recall and precision for questions related to Comorbidities, Diagnostic testing, Disease Progression Status, and Radiology/Imaging under \textit{Dense} inference. Significance thresholds: *** $p < 0.001$; ** $p < 0.01$; * $p < 0.05$; ns $=$ not significant ($p \geq 0.05$).}
    \label{sfig:recall-precision-topic-dense}
\end{figure}

\begin{figure}
    \centering
    \includegraphics[width=0.49\linewidth]{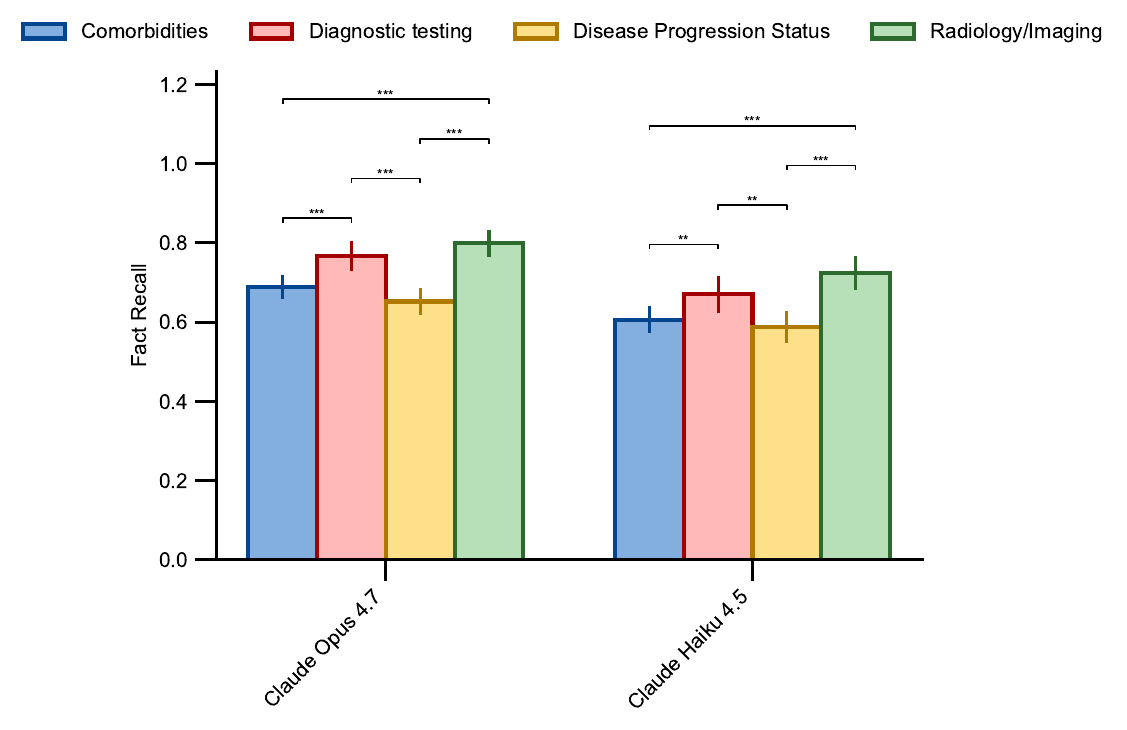}\hfill
    \includegraphics[width=0.49\linewidth]{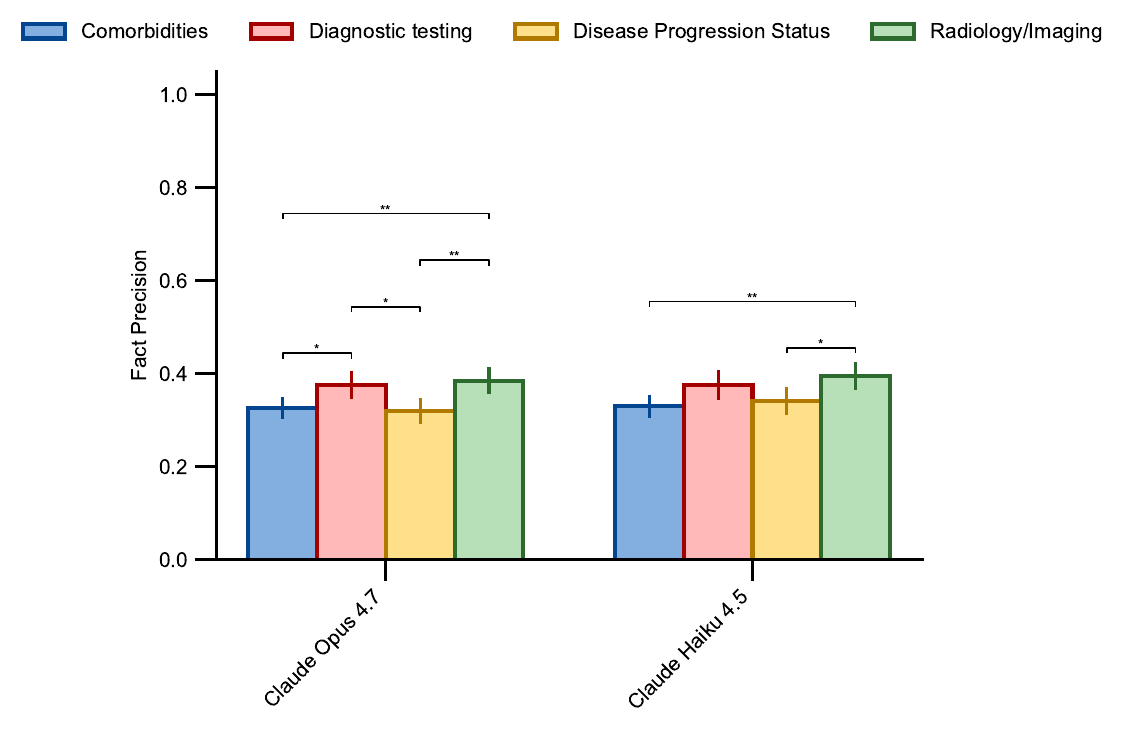}
    \caption{Fact recall and precision for questions related to Comorbidities, Diagnostic testing, Disease Progression Status, and Radiology/Imaging under \textit{Agent} inference. Significance thresholds: *** $p < 0.001$; ** $p < 0.01$; * $p < 0.05$; ns $=$ not significant ($p \geq 0.05$).}
    \label{sfig:recall-precision-topic-agent}
\end{figure}

\begin{figure}
    \centering
    \includegraphics[width=0.49\linewidth]{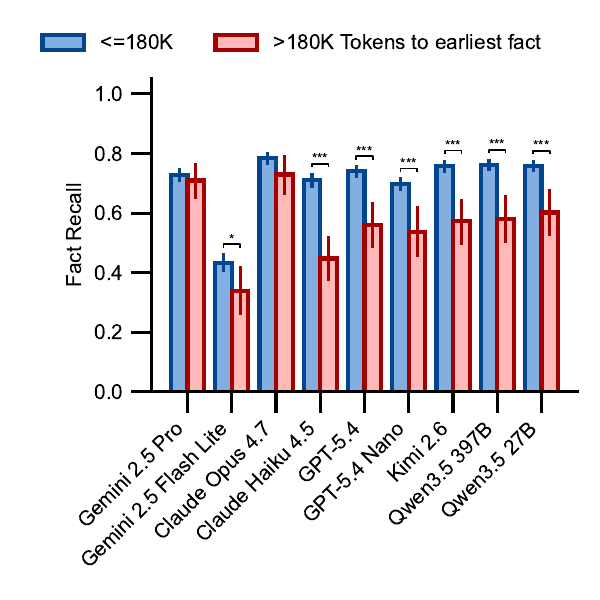}\hfill
    \includegraphics[width=0.49\linewidth]{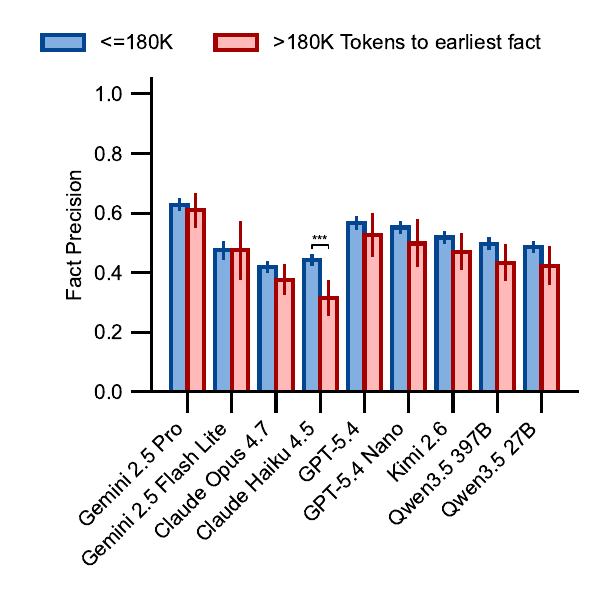}
    \caption{Fact recall and precision for answers with facts $\leq$180K and $>$180K tokens under \textit{Recent} inference. Significance thresholds: *** $p < 0.001$; ** $p < 0.01$; * $p < 0.05$; ns $=$ not significant ($p \geq 0.05$).}
    \label{sfig:recall-precision-temporality-recent}
\end{figure}
\begin{figure}
    \centering
    \includegraphics[width=0.49\linewidth]{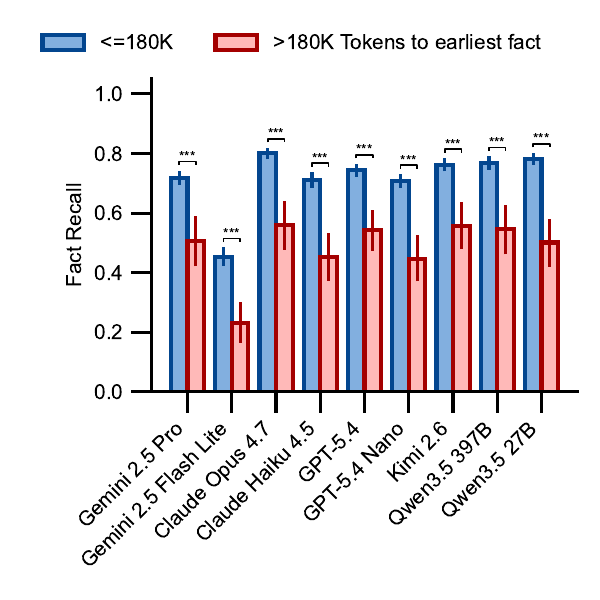}\hfill
    \includegraphics[width=0.49\linewidth]{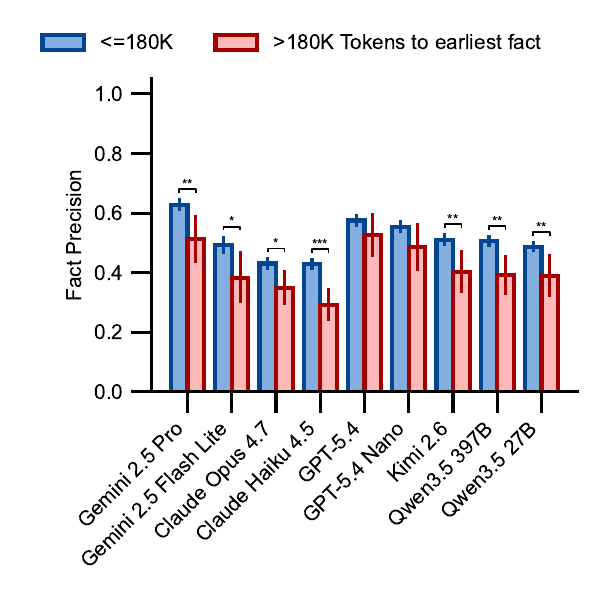}
    \caption{Fact recall and precision for answers with facts $\leq$180K and $>$180K tokens under \textit{Recent-180K} inference. Significance thresholds: *** $p < 0.001$; ** $p < 0.01$; * $p < 0.05$; ns $=$ not significant ($p \geq 0.05$).}
    \label{sfig:recall-precision-temporality-recent200}
\end{figure}
\begin{figure}
    \centering
    \includegraphics[width=0.49\linewidth]{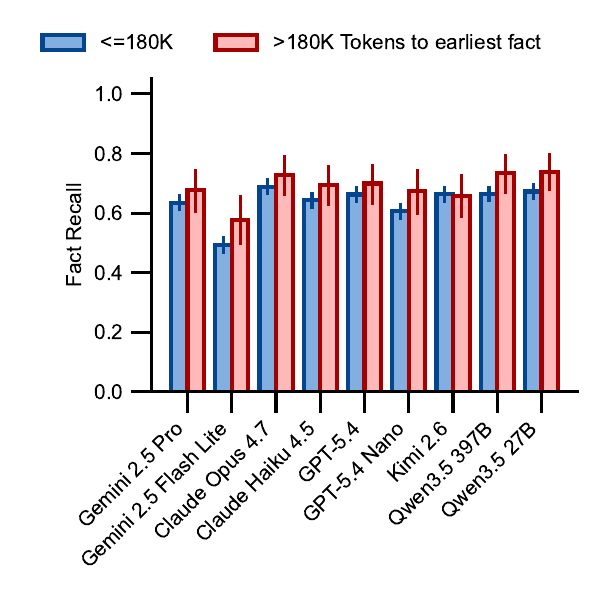}\hfill
    \includegraphics[width=0.49\linewidth]{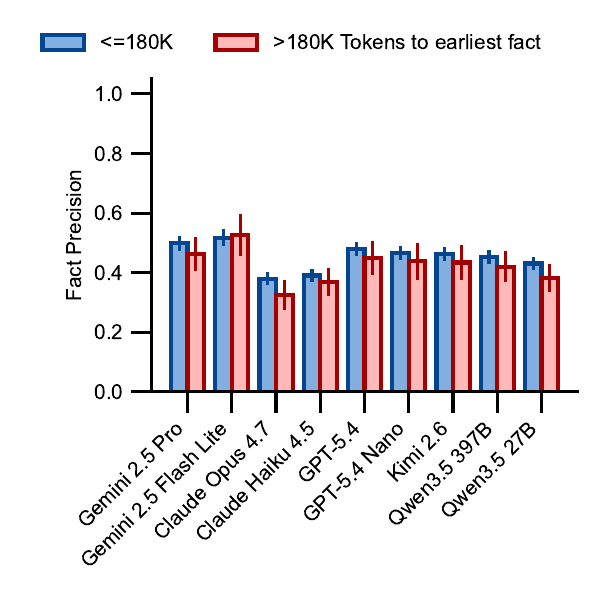}
    \caption{Fact recall and precision for answers with facts $\leq$180K and $>$180K tokens under \textit{BM25} inference. Significance thresholds: *** $p < 0.001$; ** $p < 0.01$; * $p < 0.05$; ns $=$ not significant ($p \geq 0.05$).}
    \label{sfig:recall-precision-temporality-bm25}
\end{figure}
\begin{figure}
    \centering
    \includegraphics[width=0.49\linewidth]{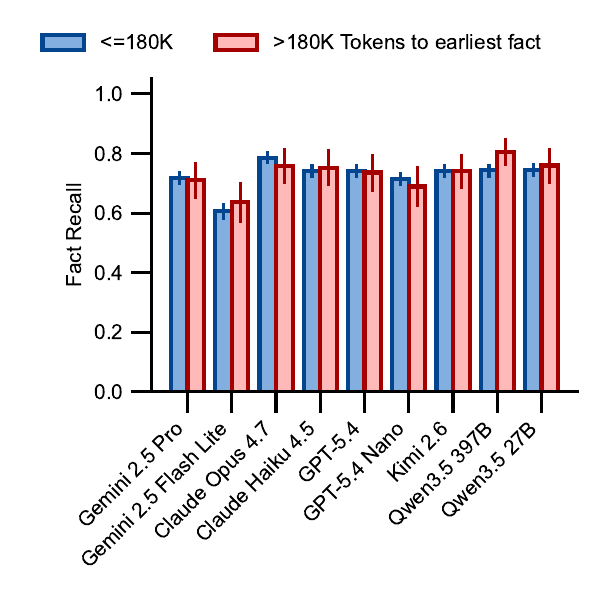}\hfill
    \includegraphics[width=0.49\linewidth]{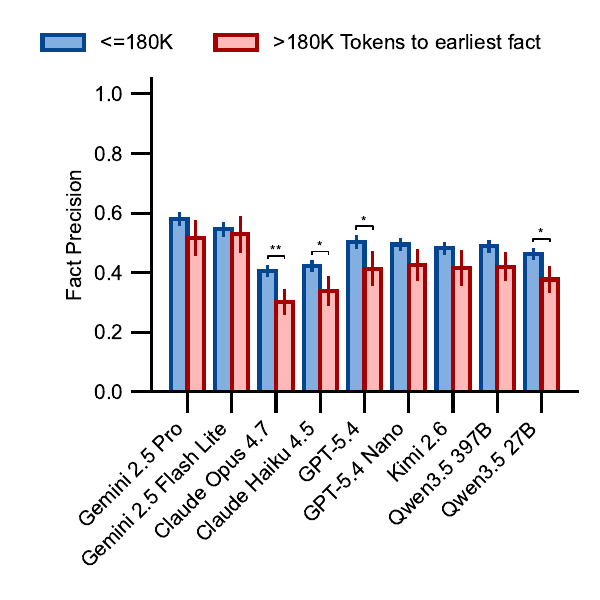}
    \caption{Fact recall and precision for answers with facts $\leq$180K and $>$180K tokens under \textit{Dense} inference. Significance thresholds: *** $p < 0.001$; ** $p < 0.01$; * $p < 0.05$; ns $=$ not significant ($p \geq 0.05$).}
    \label{sfig:recall-precision-temporality-dense}
\end{figure}
\begin{figure}
    \centering
    \includegraphics[width=0.49\linewidth]{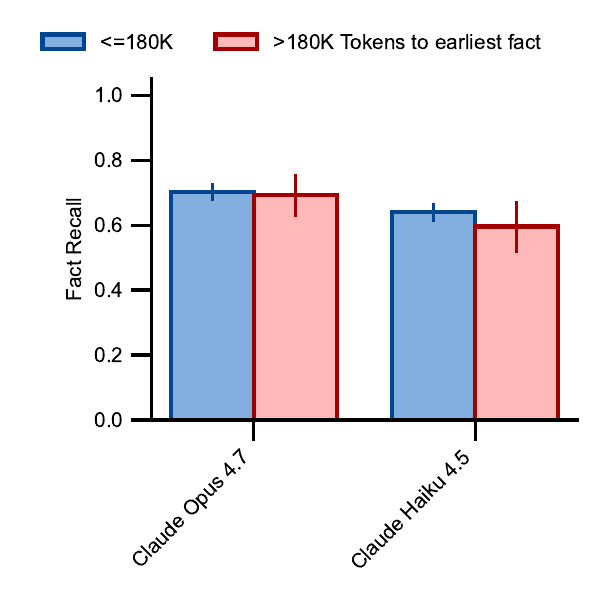}\hfill
    \includegraphics[width=0.49\linewidth]{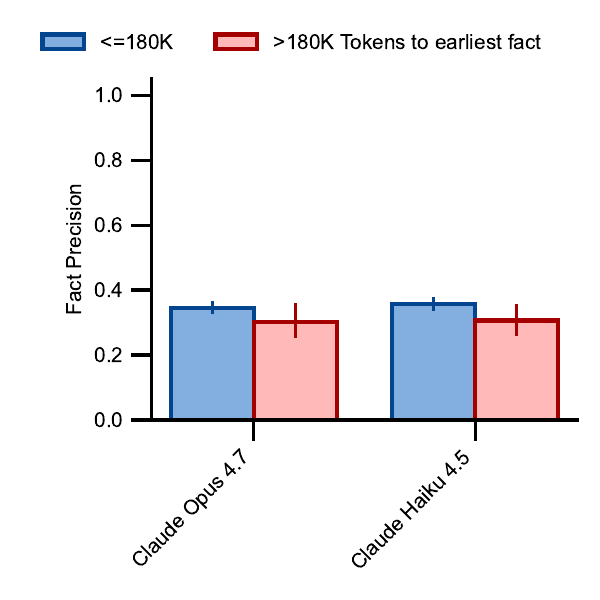}
    \caption{Fact recall and precision for answers with facts $\leq$180K and $>$180K tokens under \textit{Agent} inference. Significance thresholds: *** $p < 0.001$; ** $p < 0.01$; * $p < 0.05$; ns $=$ not significant ($p \geq 0.05$).}
    \label{sfig:recall-precision-temporality-agent}
\end{figure}
\begin{figure}
    \centering
    \includegraphics[width=0.49\linewidth]{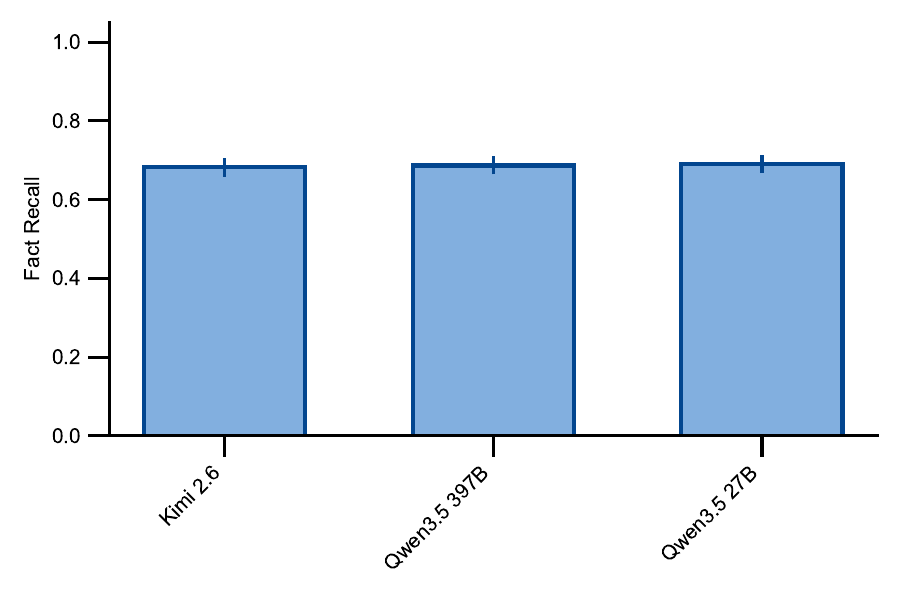}\hfill
    \includegraphics[width=0.49\linewidth]{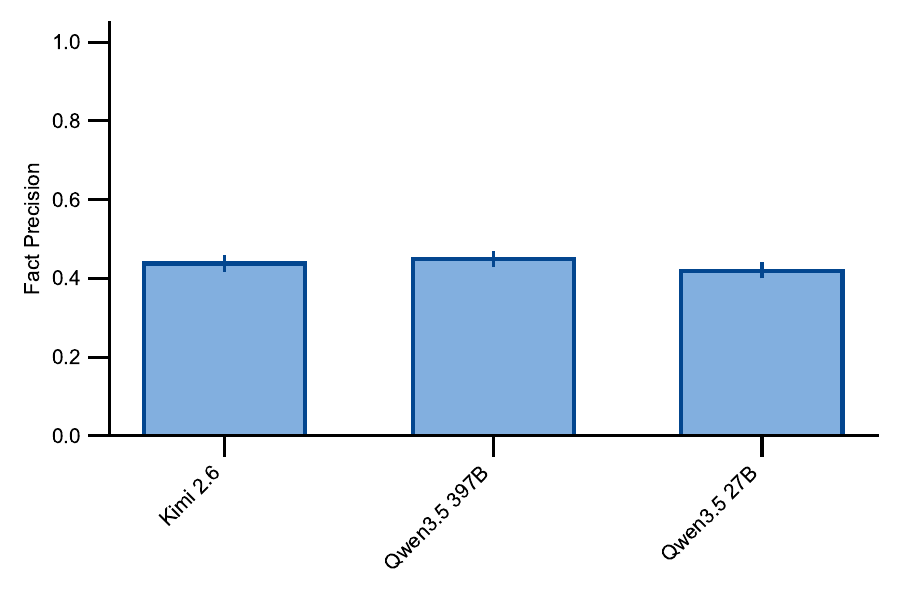}
    \caption{Fact recall and precision for late interaction retrieval for Kimi K2.6, Qwen 3.5 397B, and Qwen 3.5 27B}
    \label{sfig:late}
\end{figure}
\input{SI/FIG/tables/fact_specification}

\input{SI/FIG/tables/fact_noninferiority}
\input{SI/FIG/tables/fact_ks}

%% file: SI/FIG/tables/questiontype_count.tex
\begin{table}[!htbp]
    \centering
    \small
    \setlength{\tabcolsep}{4pt}
    \renewcommand{\arraystretch}{1.05}
    \begin{tabular}{llr}
    \toprule
    \textbf{Cohort} & \textbf{Category} & \textbf{Questions} \\
    \midrule
    \textbf{BRIE} & \textbf{Total questions} & \textbf{508} \\
    BRIE & Single Hop & 276 \\
    BRIE & Multi-Hop & 232 \\
    BRIE & Comorbidities & 296 \\
    BRIE & Diagnostic testing & 191 \\
    BRIE & Disease Progression Status & 188 \\
    BRIE & Radiology/Imaging & 187 \\
    BRIE & Tokens to earliest fact: $\leq$ 180K & 445 \\
    BRIE & Tokens to earliest fact: $>$180K & 63 \\
    \addlinespace[6pt]
    \midrule
    \addlinespace[6pt]
    \textbf{$\text{BRIE}_{\text{unfiltered}}$} & \textbf{Total questions} & \textbf{675} \\
    $\text{BRIE}_{\text{unfiltered}}$ & Single Hop & 369 \\
    $\text{BRIE}_{\text{unfiltered}}$ & Multi-Hop & 306 \\
    $\text{BRIE}_{\text{unfiltered}}$ & Comorbidities & 370 \\
    $\text{BRIE}_{\text{unfiltered}}$ & Diagnostic testing & 234 \\
    $\text{BRIE}_{\text{unfiltered}}$ & Disease Progression Status & 234 \\
    $\text{BRIE}_{\text{unfiltered}}$ & Radiology/Imaging & 225 \\
    $\text{BRIE}_{\text{unfiltered}}$ & Tokens to earliest fact: $\leq$ 180K & 588 \\
    $\text{BRIE}_{\text{unfiltered}}$ & Tokens to earliest fact: $>$ 180K & 87 \\
    \addlinespace[6pt]
    \midrule
    \addlinespace[6pt]
    \textbf{$\text{BRIE}_{\text{new}}$} & \textbf{Total questions} & \textbf{1000} \\
    $\text{BRIE}_{\text{new}}$ & Single Hop & 569 \\
    $\text{BRIE}_{\text{new}}$ & Multi-Hop & 431 \\
    $\text{BRIE}_{\text{new}}$ & Comorbidities & 636 \\
    $\text{BRIE}_{\text{new}}$ & Diagnostic testing & 297 \\
    $\text{BRIE}_{\text{new}}$ & Disease Progression Status & 405 \\
    $\text{BRIE}_{\text{new}}$ & Radiology/Imaging & 277 \\
    $\text{BRIE}_{\text{new}}$ & Tokens to earliest fact: $\leq$ 180K & 828 \\
    $\text{BRIE}_{\text{new}}$ & Tokens to earliest fact: $>$180K & 172 \\
    \bottomrule
    \end{tabular}
    \caption{Question counts for BRIE ($n=508$), $\text{BRIE}_{\text{unfiltered}}$ ($n=675$), and $\text{BRIE}_{\text{new}}$ ($n=1000$).}
    \label{tab:question-type-counts}
\end{table}

%% file: SI/FIG/tables/fact_performance.tex

\begin{table}[h]
    \centering
    \small
    \begin{tabular}{llccccc}
        \hline
        \textbf{Model} & \textbf{Inference} & \textbf{Fact Recall} & \textbf{Lower} & \textbf{Upper} & \textbf{Adj. lower} & \textbf{Adj. upper} \\
        \hline
        Gemini 2.5 Pro & Recent & 0.734 & 0.713 & 0.754 & 0.700 & 0.768 \\
        Gemini 2.5 Pro & Recent 180K & 0.696 & 0.675 & 0.720 & 0.658 & 0.734 \\
        Gemini 2.5 Pro & BM25 & 0.644 & 0.619 & 0.671 & 0.602 & 0.687 \\
        Gemini 2.5 Pro & Dense & 0.725 & 0.705 & 0.745 & 0.692 & 0.759 \\
        Gemini 2.5 Flash Lite & Recent & 0.425 & 0.397 & 0.456 & 0.376 & 0.473 \\
        Gemini 2.5 Flash Lite & Recent 180K & 0.432 & 0.401 & 0.462 & 0.384 & 0.480 \\
        Gemini 2.5 Flash Lite & BM25 & 0.500 & 0.474 & 0.528 & 0.454 & 0.546 \\
        Gemini 2.5 Flash Lite & Dense & 0.617 & 0.592 & 0.642 & 0.577 & 0.658 \\
        Claude Opus 4.7 & Recent & 0.790 & 0.771 & 0.809 & 0.758 & 0.821 \\
        Claude Opus 4.7 & Recent 180K & 0.771 & 0.750 & 0.791 & 0.737 & 0.805 \\
        Claude Opus 4.7 & BM25 & 0.698 & 0.673 & 0.724 & 0.657 & 0.739 \\
        Claude Opus 4.7 & Dense & 0.797 & 0.778 & 0.816 & 0.765 & 0.829 \\
        Claude Opus 4.7 & Agent & 0.712 & 0.690 & 0.735 & 0.674 & 0.751 \\
        Claude Haiku 4.5 & Recent & 0.683 & 0.659 & 0.708 & 0.643 & 0.723 \\
        Claude Haiku 4.5 & Recent 180K & 0.686 & 0.660 & 0.710 & 0.645 & 0.727 \\
        Claude Haiku 4.5 & BM25 & 0.652 & 0.626 & 0.680 & 0.609 & 0.696 \\
        Claude Haiku 4.5 & Dense & 0.749 & 0.729 & 0.770 & 0.714 & 0.784 \\
        Claude Haiku 4.5 & Agent & 0.641 & 0.610 & 0.668 & 0.596 & 0.685 \\
        GPT-5.4 & Recent & 0.720 & 0.699 & 0.743 & 0.685 & 0.755 \\
        GPT-5.4 & Recent 180K & 0.722 & 0.702 & 0.743 & 0.688 & 0.757 \\
        GPT-5.4 & BM25 & 0.671 & 0.647 & 0.697 & 0.631 & 0.712 \\
        GPT-5.4 & Dense & 0.753 & 0.734 & 0.774 & 0.721 & 0.786 \\
        GPT-5.4 Nano & Recent & 0.683 & 0.661 & 0.706 & 0.647 & 0.720 \\
        GPT-5.4 Nano & Recent 180K & 0.679 & 0.657 & 0.703 & 0.641 & 0.717 \\
        GPT-5.4 Nano & BM25 & 0.617 & 0.591 & 0.645 & 0.574 & 0.660 \\
        GPT-5.4 Nano & Dense & 0.712 & 0.691 & 0.734 & 0.676 & 0.748 \\
        Kimi 2.6 & Recent & 0.746 & 0.724 & 0.766 & 0.711 & 0.780 \\
        Kimi 2.6 & Recent 180K & 0.742 & 0.721 & 0.765 & 0.706 & 0.778 \\
        Kimi 2.6 & BM25 & 0.670 & 0.641 & 0.694 & 0.627 & 0.712 \\
        Kimi 2.6 & Dense & 0.752 & 0.731 & 0.773 & 0.719 & 0.786 \\
        Qwen3.5 397B & Recent & 0.747 & 0.726 & 0.769 & 0.713 & 0.782 \\
        Qwen3.5 397B & Recent 180K & 0.746 & 0.724 & 0.767 & 0.710 & 0.781 \\
        Qwen3.5 397B & BM25 & 0.673 & 0.647 & 0.697 & 0.631 & 0.715 \\
        Qwen3.5 397B & Dense & 0.764 & 0.744 & 0.783 & 0.731 & 0.796 \\
        Qwen3.5 27B & Recent & 0.748 & 0.726 & 0.769 & 0.714 & 0.781 \\
        Qwen3.5 27B & Recent 180K & 0.754 & 0.734 & 0.776 & 0.720 & 0.789 \\
        Qwen3.5 27B & BM25 & 0.682 & 0.656 & 0.708 & 0.640 & 0.723 \\
        Qwen3.5 27B & Dense & 0.758 & 0.738 & 0.777 & 0.725 & 0.791 \\
        \hline
    \end{tabular}
    \caption{Fact Recall by model--inference configuration across 508 questions. Means and 95\% confidence intervals use 1,000 paired question bootstrap resamples. Lower/Upper are uncorrected percentile bounds; Adj. lower/upper are simultaneous max-$t$ bounds.}
    \label{tab:fact-recall-bootstrap-model-inference}
\end{table}

\begin{table}[h]
    \centering
    \small
    \begin{tabular}{llccccc}
        \hline
        \textbf{Model} & \textbf{Inference} & \textbf{Fact Precision} & \textbf{Lower} & \textbf{Upper} & \textbf{Adj. lower} & \textbf{Adj. upper} \\
        \hline
        Gemini 2.5 Pro & Recent & 0.637 & 0.616 & 0.659 & 0.605 & 0.669 \\
        Gemini 2.5 Pro & Recent 180K & 0.621 & 0.600 & 0.643 & 0.587 & 0.655 \\
        Gemini 2.5 Pro & BM25 & 0.505 & 0.482 & 0.528 & 0.469 & 0.540 \\
        Gemini 2.5 Pro & Dense & 0.591 & 0.570 & 0.611 & 0.558 & 0.624 \\
        Gemini 2.5 Flash Lite & Recent & 0.484 & 0.453 & 0.514 & 0.437 & 0.530 \\
        Gemini 2.5 Flash Lite & Recent 180K & 0.488 & 0.457 & 0.517 & 0.442 & 0.533 \\
        Gemini 2.5 Flash Lite & BM25 & 0.524 & 0.498 & 0.552 & 0.481 & 0.566 \\
        Gemini 2.5 Flash Lite & Dense & 0.554 & 0.530 & 0.577 & 0.517 & 0.591 \\
        Claude Opus 4.7 & Recent & 0.425 & 0.407 & 0.446 & 0.395 & 0.455 \\
        Claude Opus 4.7 & Recent 180K & 0.428 & 0.409 & 0.449 & 0.397 & 0.458 \\
        Claude Opus 4.7 & BM25 & 0.381 & 0.362 & 0.401 & 0.350 & 0.412 \\
        Claude Opus 4.7 & Dense & 0.406 & 0.386 & 0.425 & 0.375 & 0.436 \\
        Claude Opus 4.7 & Agent & 0.349 & 0.332 & 0.371 & 0.320 & 0.378 \\
        Claude Haiku 4.5 & Recent & 0.437 & 0.419 & 0.458 & 0.408 & 0.467 \\
        Claude Haiku 4.5 & Recent 180K & 0.419 & 0.401 & 0.440 & 0.389 & 0.449 \\
        Claude Haiku 4.5 & BM25 & 0.399 & 0.380 & 0.418 & 0.369 & 0.428 \\
        Claude Haiku 4.5 & Dense & 0.422 & 0.405 & 0.440 & 0.393 & 0.450 \\
        Claude Haiku 4.5 & Agent & 0.361 & 0.343 & 0.382 & 0.331 & 0.392 \\
        GPT-5.4 & Recent & 0.572 & 0.551 & 0.594 & 0.538 & 0.605 \\
        GPT-5.4 & Recent 180K & 0.579 & 0.558 & 0.602 & 0.546 & 0.612 \\
        GPT-5.4 & BM25 & 0.487 & 0.467 & 0.510 & 0.453 & 0.521 \\
        GPT-5.4 & Dense & 0.506 & 0.484 & 0.527 & 0.473 & 0.538 \\
        GPT-5.4 Nano & Recent & 0.552 & 0.529 & 0.576 & 0.518 & 0.586 \\
        GPT-5.4 Nano & Recent 180K & 0.552 & 0.530 & 0.575 & 0.518 & 0.586 \\
        GPT-5.4 Nano & BM25 & 0.470 & 0.450 & 0.492 & 0.436 & 0.505 \\
        GPT-5.4 Nano & Dense & 0.493 & 0.473 & 0.515 & 0.460 & 0.526 \\
        Kimi 2.6 & Recent & 0.523 & 0.502 & 0.544 & 0.490 & 0.555 \\
        Kimi 2.6 & Recent 180K & 0.506 & 0.486 & 0.528 & 0.474 & 0.538 \\
        Kimi 2.6 & BM25 & 0.472 & 0.450 & 0.494 & 0.437 & 0.506 \\
        Kimi 2.6 & Dense & 0.482 & 0.462 & 0.503 & 0.452 & 0.513 \\
        Qwen3.5 397B & Recent & 0.496 & 0.477 & 0.516 & 0.465 & 0.526 \\
        Qwen3.5 397B & Recent 180K & 0.499 & 0.480 & 0.518 & 0.468 & 0.529 \\
        Qwen3.5 397B & BM25 & 0.457 & 0.435 & 0.478 & 0.424 & 0.490 \\
        Qwen3.5 397B & Dense & 0.493 & 0.474 & 0.512 & 0.463 & 0.523 \\
        Qwen3.5 27B & Recent & 0.488 & 0.469 & 0.508 & 0.458 & 0.517 \\
        Qwen3.5 27B & Recent 180K & 0.484 & 0.466 & 0.504 & 0.455 & 0.513 \\
        Qwen3.5 27B & BM25 & 0.434 & 0.415 & 0.454 & 0.403 & 0.465 \\
        Qwen3.5 27B & Dense & 0.464 & 0.446 & 0.483 & 0.435 & 0.493 \\
        \hline
    \end{tabular}
    \caption{Fact Precision by model--inference configuration across 508 questions. Means and 95\% confidence intervals use 1,000 paired question bootstrap resamples. Lower/Upper are uncorrected percentile bounds; Adj. lower/upper are simultaneous max-$t$ bounds}
    \label{tab:fact-precision-bootstrap-model-inference}
\end{table}

%% file: SI/FIG/tables/questiontype_mwu.tex

\begingroup
\small
\setlength{\tabcolsep}{4pt}
\renewcommand{\arraystretch}{1.2}
\setlength{\LTcapwidth}{\linewidth}
\clearpage
\renewcommand{\arraystretch}{1.2}
\begin{longtable}{>{\centering\arraybackslash}p{0.24\linewidth} c r r r}
\toprule
\textbf{Model / inference} & \shortstack{\textbf{Test samples}\\$n_A / n_B$} & $\boldsymbol{U_A}$ & $\boldsymbol{p}$ & $\boldsymbol{p_{\mathrm{BH}}}$ \\
\midrule
\endfirsthead
\multicolumn{5}{l}{\textit{Reasoning comparisons continued}}\\
\toprule
\textbf{Model / inference} & \shortstack{\textbf{Test samples}\\$n_A / n_B$} & $\boldsymbol{U_A}$ & $\boldsymbol{p}$ & $\boldsymbol{p_{\mathrm{BH}}}$ \\
\midrule
\endhead
\midrule\multicolumn{5}{r}{\textit{Continued on next page}}\\\endfoot
\endlastfoot
\multicolumn{5}{l}{\textbf{Family 1: Models}\quad (9 hypotheses)} \\*
\addlinespace[3pt]
Gemini 2.5 Pro & 1104 / 928 & 686,164.5 & $3.70\times10^{-40}$ & {\boldmath\bfseries $4.76\times10^{-40}$} \\
Gemini 2.5 Flash Lite & 1104 / 928 & 592,252.0 & $1.12\times10^{-9}$ & {\boldmath\bfseries $1.12\times10^{-9}$} \\
Claude Opus 4.7 & 1380 / 1160 & 1,123,048.5 & $2.04\times10^{-70}$ & {\boldmath\bfseries $1.84\times10^{-69}$} \\
Claude Haiku 4.5 & 1380 / 1160 & 1,116,796.0 & $6.06\times10^{-67}$ & {\boldmath\bfseries $2.72\times10^{-66}$} \\
GPT-5.4 & 1104 / 928 & 648,879.0 & $1.82\times10^{-25}$ & {\boldmath\bfseries $2.04\times10^{-25}$} \\
GPT-5.4 Nano & 1104 / 928 & 692,827.0 & $4.73\times10^{-43}$ & {\boldmath\bfseries $7.10\times10^{-43}$} \\
Kimi 2.6 & 1104 / 928 & 711,117.5 & $2.77\times10^{-52}$ & {\boldmath\bfseries $4.99\times10^{-52}$} \\
Qwen3.5 397B & 1104 / 928 & 711,824.5 & $1.25\times10^{-52}$ & {\boldmath\bfseries $2.82\times10^{-52}$} \\
Qwen3.5 27B & 1104 / 928 & 714,299.5 & $6.04\times10^{-54}$ & {\boldmath\bfseries $1.81\times10^{-53}$} \\*
\midrule
\multicolumn{5}{l}{\textbf{Family 2: Inference methods}\quad (5 hypotheses)} \\*
\addlinespace[3pt]
Recent & 2484 / 2088 & 3,406,633.5 & $1.13\times10^{-75}$ & {\boldmath\bfseries $1.89\times10^{-75}$} \\
Recent 180K & 2484 / 2088 & 3,345,029.5 & $6.72\times10^{-65}$ & {\boldmath\bfseries $8.40\times10^{-65}$} \\
BM25 & 2484 / 2088 & 3,466,774.0 & $1.24\times10^{-86}$ & {\boldmath\bfseries $3.10\times10^{-86}$} \\
Dense & 2484 / 2088 & 3,604,150.5 & $4.25\times10^{-116}$ & {\boldmath\bfseries $2.12\times10^{-115}$} \\
Agent & 552 / 464 & 189,761.0 & $1.69\times10^{-40}$ & {\boldmath\bfseries $1.69\times10^{-40}$} \\*
\bottomrule
\caption{Two-sided Mann--Whitney $U$ tests of fact recall in BRIE. The comparison is single-hop ($A$) versus multi-hop ($B$) questions. Model comparisons pool inference methods; inference comparisons pool models. Benjamini--Hochberg correction is separate within each displayed family. Bold adjusted $p$ denotes $p_{\mathrm{BH}}<0.05$. $n_A,n_B$: tested score counts; $U_A$: statistic for group $A$.}\label{tab:question-type-reasoning}\\
\end{longtable}

\clearpage
\renewcommand{\arraystretch}{1.1}
\begin{longtable}{>{\centering\arraybackslash}p{0.24\linewidth} c c r r r}
\toprule
\textbf{Model / inference} & \textbf{Pair} & \shortstack{\textbf{Test samples}\\$n_A / n_B$} & $\boldsymbol{U_A}$ & $\boldsymbol{p}$ & $\boldsymbol{p_{\mathrm{BH}}}$ \\
\midrule
\endfirsthead
\multicolumn{6}{l}{\textit{Topic comparisons continued}}\\
\toprule
\textbf{Model / inference} & \textbf{Pair} & \shortstack{\textbf{Test samples}\\$n_A / n_B$} & $\boldsymbol{U_A}$ & $\boldsymbol{p}$ & $\boldsymbol{p_{\mathrm{BH}}}$ \\
\midrule
\endhead
\midrule\multicolumn{6}{r}{\textit{Continued on next page}}\\\endfoot
\endlastfoot
\multicolumn{6}{l}{\textbf{Family 3: Models}\quad (54 hypotheses)} \\*
\addlinespace[3pt]
Gemini 2.5 Pro & C vs. D & 1184 / 764 & 380,923.0 & $3.21\times10^{-9}$ & {\boldmath\bfseries $5.58\times10^{-9}$} \\
Gemini 2.5 Pro & C vs. P & 1184 / 752 & 463,767.5 & 0.1201 & 0.1541 \\
Gemini 2.5 Pro & C vs. I & 1184 / 748 & 371,722.0 & $2.21\times10^{-9}$ & {\boldmath\bfseries $4.12\times10^{-9}$} \\
Gemini 2.5 Pro & D vs. P & 764 / 752 & 345,163.5 & $8.32\times10^{-12}$ & {\boldmath\bfseries $1.87\times10^{-11}$} \\
Gemini 2.5 Pro & D vs. I & 764 / 748 & 286,885.5 & 0.8913 & 0.9081 \\
Gemini 2.5 Pro & P vs. I & 752 / 748 & 223,007.0 & $3.02\times10^{-12}$ & {\boldmath\bfseries $7.42\times10^{-12}$} \\
Gemini 2.5 Flash Lite & C vs. D & 1184 / 764 & 413,551.5 & 0.0014 & {\boldmath\bfseries 0.0020} \\
Gemini 2.5 Flash Lite & C vs. P & 1184 / 752 & 463,656.0 & 0.1227 & 0.1541 \\
Gemini 2.5 Flash Lite & C vs. I & 1184 / 748 & 384,932.5 & $1.18\times10^{-6}$ & {\boldmath\bfseries $1.87\times10^{-6}$} \\
Gemini 2.5 Flash Lite & D vs. P & 764 / 752 & 323,426.0 & $2.09\times10^{-5}$ & {\boldmath\bfseries $3.23\times10^{-5}$} \\
Gemini 2.5 Flash Lite & D vs. I & 764 / 748 & 274,394.5 & 0.1796 & 0.2155 \\
Gemini 2.5 Flash Lite & P vs. I & 752 / 748 & 232,545.5 & $5.83\times10^{-9}$ & {\boldmath\bfseries $9.84\times10^{-9}$} \\
Claude Opus 4.7 & C vs. D & 1480 / 955 & 585,972.0 & $4.89\times10^{-13}$ & {\boldmath\bfseries $1.32\times10^{-12}$} \\
Claude Opus 4.7 & C vs. P & 1480 / 940 & 744,125.5 & 0.0036 & {\boldmath\bfseries 0.0052} \\
Claude Opus 4.7 & C vs. I & 1480 / 935 & 563,008.5 & $4.81\times10^{-15}$ & {\boldmath\bfseries $1.85\times10^{-14}$} \\
Claude Opus 4.7 & D vs. P & 955 / 940 & 558,306.0 & $1.38\times10^{-20}$ & {\boldmath\bfseries $2.48\times10^{-19}$} \\
Claude Opus 4.7 & D vs. I & 955 / 935 & 440,900.0 & 0.6295 & 0.6799 \\
Claude Opus 4.7 & P vs. I & 940 / 935 & 324,996.5 & $4.94\times10^{-23}$ & {\boldmath\bfseries $2.67\times10^{-21}$} \\
Claude Haiku 4.5 & C vs. D & 1480 / 955 & 590,198.0 & $4.34\times10^{-12}$ & {\boldmath\bfseries $1.02\times10^{-11}$} \\
Claude Haiku 4.5 & C vs. P & 1480 / 940 & 740,662.5 & 0.0070 & {\boldmath\bfseries 0.0099} \\
Claude Haiku 4.5 & C vs. I & 1480 / 935 & 564,917.5 & $1.85\times10^{-14}$ & {\boldmath\bfseries $6.67\times10^{-14}$} \\
Claude Haiku 4.5 & D vs. P & 955 / 940 & 553,218.0 & $1.19\times10^{-18}$ & {\boldmath\bfseries $1.28\times10^{-17}$} \\
Claude Haiku 4.5 & D vs. I & 955 / 935 & 439,739.5 & 0.5651 & 0.6227 \\
Claude Haiku 4.5 & P vs. I & 940 / 935 & 328,469.0 & $1.66\times10^{-21}$ & {\boldmath\bfseries $4.49\times10^{-20}$} \\
GPT-5.4 & C vs. D & 1184 / 764 & 378,854.5 & $1.06\times10^{-9}$ & {\boldmath\bfseries $2.12\times10^{-9}$} \\
GPT-5.4 & C vs. P & 1184 / 752 & 458,319.0 & 0.2716 & 0.3188 \\
GPT-5.4 & C vs. I & 1184 / 748 & 378,518.0 & $6.06\times10^{-8}$ & {\boldmath\bfseries $9.92\times10^{-8}$} \\*
\newpage
\multicolumn{6}{l}{\textbf{Family 3: Models (continued)}\quad (54 hypotheses)} \\*
\addlinespace[3pt]
GPT-5.4 & D vs. P & 764 / 752 & 343,795.5 & $2.43\times10^{-11}$ & {\boldmath\bfseries $5.25\times10^{-11}$} \\
GPT-5.4 & D vs. I & 764 / 748 & 291,385.0 & 0.5006 & 0.5632 \\
GPT-5.4 & P vs. I & 752 / 748 & 230,699.0 & $1.34\times10^{-9}$ & {\boldmath\bfseries $2.59\times10^{-9}$} \\
GPT-5.4 Nano & C vs. D & 1184 / 764 & 380,722.5 & $3.00\times10^{-9}$ & {\boldmath\bfseries $5.40\times10^{-9}$} \\
GPT-5.4 Nano & C vs. P & 1184 / 752 & 466,792.5 & 0.0709 & 0.0933 \\
GPT-5.4 Nano & C vs. I & 1184 / 748 & 364,153.0 & $3.61\times10^{-11}$ & {\boldmath\bfseries $7.49\times10^{-11}$} \\
GPT-5.4 Nano & D vs. P & 764 / 752 & 348,812.0 & $4.12\times10^{-13}$ & {\boldmath\bfseries $1.17\times10^{-12}$} \\
GPT-5.4 Nano & D vs. I & 764 / 748 & 279,737.0 & 0.4753 & 0.5461 \\
GPT-5.4 Nano & P vs. I & 752 / 748 & 215,549.0 & $3.59\times10^{-15}$ & {\boldmath\bfseries $1.62\times10^{-14}$} \\
Kimi 2.6 & C vs. D & 1184 / 764 & 358,140.0 & $4.44\times10^{-15}$ & {\boldmath\bfseries $1.85\times10^{-14}$} \\
Kimi 2.6 & C vs. P & 1184 / 752 & 461,523.0 & 0.1712 & 0.2101 \\
Kimi 2.6 & C vs. I & 1184 / 748 & 346,717.0 & $4.52\times10^{-16}$ & {\boldmath\bfseries $2.22\times10^{-15}$} \\
Kimi 2.6 & D vs. P & 764 / 752 & 358,743.5 & $2.54\times10^{-17}$ & {\boldmath\bfseries $1.37\times10^{-16}$} \\
Kimi 2.6 & D vs. I & 764 / 748 & 284,172.0 & 0.8506 & 0.8960 \\
Kimi 2.6 & P vs. I & 752 / 748 & 208,686.5 & $2.47\times10^{-18}$ & {\boldmath\bfseries $2.23\times10^{-17}$} \\
Qwen3.5 397B & C vs. D & 1184 / 764 & 362,989.0 & $1.04\times10^{-13}$ & {\boldmath\bfseries $3.50\times10^{-13}$} \\
Qwen3.5 397B & C vs. P & 1184 / 752 & 467,941.5 & 0.0566 & 0.0764 \\
Qwen3.5 397B & C vs. I & 1184 / 748 & 355,862.5 & $2.06\times10^{-13}$ & {\boldmath\bfseries $6.17\times10^{-13}$} \\
Qwen3.5 397B & D vs. P & 764 / 752 & 359,364.5 & $1.35\times10^{-17}$ & {\boldmath\bfseries $9.08\times10^{-17}$} \\
Qwen3.5 397B & D vs. I & 764 / 748 & 287,174.5 & 0.8628 & 0.8960 \\
Qwen3.5 397B & P vs. I & 752 / 748 & 210,611.0 & $1.97\times10^{-17}$ & {\boldmath\bfseries $1.18\times10^{-16}$} \\
Qwen3.5 27B & C vs. D & 1184 / 764 & 367,364.0 & $1.47\times10^{-12}$ & {\boldmath\bfseries $3.79\times10^{-12}$} \\
Qwen3.5 27B & C vs. P & 1184 / 752 & 469,610.0 & 0.0407 & 0.0564 \\
Qwen3.5 27B & C vs. I & 1184 / 748 & 355,889.5 & $2.05\times10^{-13}$ & {\boldmath\bfseries $6.17\times10^{-13}$} \\
Qwen3.5 27B & D vs. P & 764 / 752 & 359,654.5 & $1.04\times10^{-17}$ & {\boldmath\bfseries $7.99\times10^{-17}$} \\
Qwen3.5 27B & D vs. I & 764 / 748 & 285,446.5 & 0.9723 & 0.9723 \\
Qwen3.5 27B & P vs. I & 752 / 748 & 207,263.5 & $6.04\times10^{-19}$ & {\boldmath\bfseries $8.16\times10^{-18}$} \\*
\newpage
\multicolumn{6}{l}{\textbf{Family 4: Inference methods}\quad (30 hypotheses)} \\*
\addlinespace[3pt]
Recent & C vs. D & 2664 / 1719 & 1,891,500.0 & $1.08\times10^{-22}$ & {\boldmath\bfseries $3.23\times10^{-22}$} \\
Recent & C vs. P & 2664 / 1692 & 2,334,174.0 & 0.0461 & 0.0576 \\
Recent & C vs. I & 2664 / 1683 & 1,833,130.0 & $1.80\times10^{-24}$ & {\boldmath\bfseries $6.74\times10^{-24}$} \\
Recent & D vs. P & 1719 / 1692 & 1,772,442.5 & $8.57\times10^{-29}$ & {\boldmath\bfseries $6.43\times10^{-28}$} \\
Recent & D vs. I & 1719 / 1683 & 1,438,119.0 & 0.7655 & 0.7655 \\
Recent & P vs. I & 1692 / 1683 & 1,099,815.0 & $1.06\times10^{-30}$ & {\boldmath\bfseries $1.06\times10^{-29}$} \\
Recent 180K & C vs. D & 2664 / 1719 & 1,925,115.5 & $2.83\times10^{-19}$ & {\boldmath\bfseries $6.07\times10^{-19}$} \\
Recent 180K & C vs. P & 2664 / 1692 & 2,341,928.5 & 0.0288 & {\boldmath\bfseries 0.0375} \\
Recent 180K & C vs. I & 2664 / 1683 & 1,853,526.5 & $3.07\times10^{-22}$ & {\boldmath\bfseries $7.69\times10^{-22}$} \\
Recent 180K & D vs. P & 1719 / 1692 & 1,751,557.5 & $2.41\times10^{-25}$ & {\boldmath\bfseries $1.20\times10^{-24}$} \\
Recent 180K & D vs. I & 1719 / 1683 & 1,430,372.5 & 0.5670 & 0.6300 \\
Recent 180K & P vs. I & 1692 / 1683 & 1,111,452.5 & $1.19\times10^{-28}$ & {\boldmath\bfseries $7.14\times10^{-28}$} \\
BM25 & C vs. D & 2664 / 1719 & 1,959,456.0 & $4.78\times10^{-16}$ & {\boldmath\bfseries $9.57\times10^{-16}$} \\
BM25 & C vs. P & 2664 / 1692 & 2,377,392.0 & 0.0022 & {\boldmath\bfseries 0.0030} \\
BM25 & C vs. I & 2664 / 1683 & 1,929,590.5 & $7.11\times10^{-15}$ & {\boldmath\bfseries $1.33\times10^{-14}$} \\
BM25 & D vs. P & 1719 / 1692 & 1,740,239.5 & $1.62\times10^{-23}$ & {\boldmath\bfseries $5.39\times10^{-23}$} \\
BM25 & D vs. I & 1719 / 1683 & 1,459,501.5 & 0.6474 & 0.6936 \\
BM25 & P vs. I & 1692 / 1683 & 1,149,036.5 & $1.81\times10^{-22}$ & {\boldmath\bfseries $4.95\times10^{-22}$} \\
Dense & C vs. D & 2664 / 1719 & 1,898,395.0 & $4.74\times10^{-22}$ & {\boldmath\bfseries $1.09\times10^{-21}$} \\
Dense & C vs. P & 2664 / 1692 & 2,394,267.0 & $4.85\times10^{-4}$ & {\boldmath\bfseries $6.93\times10^{-4}$} \\
Dense & C vs. I & 2664 / 1683 & 1,828,060.0 & $4.00\times10^{-25}$ & {\boldmath\bfseries $1.71\times10^{-24}$} \\
Dense & D vs. P & 1719 / 1692 & 1,794,052.5 & $9.65\times10^{-33}$ & {\boldmath\bfseries $1.45\times10^{-31}$} \\
Dense & D vs. I & 1719 / 1683 & 1,435,037.0 & 0.6828 & 0.7064 \\
Dense & P vs. I & 1692 / 1683 & 1,068,399.5 & $9.71\times10^{-37}$ & {\boldmath\bfseries $2.91\times10^{-35}$} \\
Agent & C vs. D & 592 / 382 & 94,911.5 & $1.99\times10^{-5}$ & {\boldmath\bfseries $2.99\times10^{-5}$} \\
Agent & C vs. P & 592 / 376 & 118,732.5 & 0.0786 & 0.0907 \\
Agent & C vs. I & 592 / 374 & 82,972.5 & $3.72\times10^{-11}$ & {\boldmath\bfseries $6.20\times10^{-11}$} \\
Agent & D vs. P & 382 / 376 & 88,106.5 & $5.38\times10^{-8}$ & {\boldmath\bfseries $8.49\times10^{-8}$} \\
Agent & D vs. I & 382 / 374 & 66,050.5 & 0.0682 & 0.0819 \\
Agent & P vs. I & 376 / 374 & 47,501.0 & $9.67\times10^{-15}$ & {\boldmath\bfseries $1.71\times10^{-14}$} \\*
\bottomrule
\caption{Two-sided Mann--Whitney $U$ tests of fact recall in BRIE. Comparisons are all six pairs among Comorbidities (C), Diagnostic testing (D), Disease Progression Status (P), and Radiology/Imaging (I); $A$ and $B$ follow the displayed pair order. Topic categories overlap. Model comparisons pool inference methods; inference comparisons pool models. Benjamini--Hochberg correction is separate within each displayed family. Bold adjusted $p$ denotes $p_{\mathrm{BH}}<0.05$. $n_A,n_B$: tested score counts; $U_A$: statistic for group $A$. }\label{tab:question-type-topic}\\
\end{longtable}

\clearpage
\renewcommand{\arraystretch}{1.2}
\begin{longtable}{>{\centering\arraybackslash}p{0.24\linewidth} c r r r}
\toprule
\textbf{Model / inference} & \shortstack{\textbf{Test samples}\\$n_A / n_B$} & $\boldsymbol{U_A}$ & $\boldsymbol{p}$ & $\boldsymbol{p_{\mathrm{BH}}}$ \\
\midrule
\endfirsthead
\multicolumn{5}{l}{\textit{Time comparisons continued}}\\
\toprule
\textbf{Model / inference} & \shortstack{\textbf{Test samples}\\$n_A / n_B$} & $\boldsymbol{U_A}$ & $\boldsymbol{p}$ & $\boldsymbol{p_{\mathrm{BH}}}$ \\
\midrule
\endhead
\midrule\multicolumn{5}{r}{\textit{Continued on next page}}\\\endfoot
\endlastfoot
\multicolumn{5}{l}{\textbf{Family 5: Models}\quad (9 hypotheses)} \\*
\addlinespace[3pt]
Gemini 2.5 Pro & 1780 / 252 & 252,878.5 & $9.79\times10^{-4}$ & {\boldmath\bfseries 0.0011} \\
Gemini 2.5 Flash Lite & 1780 / 252 & 249,348.5 & 0.0039 & {\boldmath\bfseries 0.0039} \\
Claude Opus 4.7 & 2225 / 315 & 398,887.0 & $5.67\times10^{-5}$ & {\boldmath\bfseries $7.30\times10^{-5}$} \\
Claude Haiku 4.5 & 2225 / 315 & 429,676.5 & $6.06\times10^{-11}$ & {\boldmath\bfseries $5.45\times10^{-10}$} \\
GPT-5.4 & 1780 / 252 & 275,582.0 & $3.25\times10^{-9}$ & {\boldmath\bfseries $9.74\times10^{-9}$} \\
GPT-5.4 Nano & 1780 / 252 & 271,882.5 & $4.25\times10^{-8}$ & {\boldmath\bfseries $7.65\times10^{-8}$} \\
Kimi 2.6 & 1780 / 252 & 279,552.0 & $1.64\times10^{-10}$ & {\boldmath\bfseries $7.39\times10^{-10}$} \\
Qwen3.5 397B & 1780 / 252 & 259,200.0 & $5.40\times10^{-5}$ & {\boldmath\bfseries $7.30\times10^{-5}$} \\
Qwen3.5 27B & 1780 / 252 & 271,664.5 & $4.23\times10^{-8}$ & {\boldmath\bfseries $7.65\times10^{-8}$} \\*
\midrule
\multicolumn{5}{l}{\textbf{Family 6: Inference methods}\quad (5 hypotheses)} \\*
\addlinespace[3pt]
Recent & 4005 / 567 & 1,474,990.5 & $3.46\times10^{-31}$ & {\boldmath\bfseries $8.65\times10^{-31}$} \\
Recent 180K & 4005 / 567 & 1,648,030.0 & $8.21\times10^{-69}$ & {\boldmath\bfseries $4.11\times10^{-68}$} \\
BM25 & 4005 / 567 & 1,106,820.5 & 0.3290 & 0.4113 \\
Dense & 4005 / 567 & 1,156,535.5 & 0.4696 & 0.4696 \\
Agent & 890 / 126 & 59,686.0 & 0.2380 & 0.3967 \\*
\bottomrule
\caption{Two-sided Mann--Whitney $U$ tests of fact recall in BRIE. The comparison is questions with $0\leq t\leq180{,}000$ ($A$) versus $t>180{,}000$ ($B$) tokens to the earliest fact. Model comparisons pool inference methods; inference comparisons pool models. Benjamini--Hochberg correction is separate within each displayed family. Bold adjusted $p$ denotes $p_{\mathrm{BH}}<0.05$. $n_A,n_B$: tested score counts; $U_A$: statistic for group $A$. }\label{tab:question-type-time}\\
\end{longtable}

\endgroup

%% file: SI/FIG/tables/fact_specification.tex

\begin{table}[h]
    \centering
    \small
    \begin{tabular}{llccc}
        \hline
        \textbf{Model} & \textbf{Inference} & $n_{\ne 0}$ & $p$ & \textbf{Adj.} $p$ \\
        \hline
        Gemini 2.5 Pro & Recent & 46 & $3.52\times 10^{-9}$ & $1.34\times 10^{-8}$ \\
        Gemini 2.5 Pro & Recent 180K & 41 & $2.42\times 10^{-8}$ & $2.79\times 10^{-8}$ \\
        Gemini 2.5 Pro & BM25 & 45 & $5.17\times 10^{-9}$ & $1.51\times 10^{-8}$ \\
        Gemini 2.5 Pro & Dense & 46 & $3.52\times 10^{-9}$ & $1.34\times 10^{-8}$ \\
        Gemini 2.5 Flash Lite & Recent & 44 & $7.60\times 10^{-9}$ & $1.52\times 10^{-8}$ \\
        Gemini 2.5 Flash Lite & Recent 180K & 43 & $1.12\times 10^{-8}$ & $1.77\times 10^{-8}$ \\
        Gemini 2.5 Flash Lite & BM25 & 45 & $5.17\times 10^{-9}$ & $1.51\times 10^{-8}$ \\
        Gemini 2.5 Flash Lite & Dense & 44 & $7.60\times 10^{-9}$ & $1.52\times 10^{-8}$ \\
        Claude Opus 4.7 & Recent & 44 & $7.57\times 10^{-9}$ & $1.52\times 10^{-8}$ \\
        Claude Opus 4.7 & Recent 180K & 41 & $2.42\times 10^{-8}$ & $2.79\times 10^{-8}$ \\
        Claude Opus 4.7 & BM25 & 50 & $7.55\times 10^{-10}$ & $1.30\times 10^{-8}$ \\
        Claude Opus 4.7 & Dense & 44 & $7.61\times 10^{-9}$ & $1.52\times 10^{-8}$ \\
        Claude Opus 4.7 & Agent & 43 & $1.12\times 10^{-8}$ & $1.77\times 10^{-8}$ \\
        Claude Haiku 4.5 & Recent & 42 & $1.65\times 10^{-8}$ & $2.16\times 10^{-8}$ \\
        Claude Haiku 4.5 & Recent 180K & 20 & $8.84\times 10^{-5}$ & $8.84\times 10^{-5}$ \\
        Claude Haiku 4.5 & BM25 & 41 & $2.42\times 10^{-8}$ & $2.79\times 10^{-8}$ \\
        Claude Haiku 4.5 & Dense & 37 & $1.14\times 10^{-7}$ & $1.17\times 10^{-7}$ \\
        Claude Haiku 4.5 & Agent & 49 & $1.11\times 10^{-9}$ & $1.30\times 10^{-8}$ \\
        GPT-5.4 & Recent & 42 & $1.65\times 10^{-8}$ & $2.16\times 10^{-8}$ \\
        GPT-5.4 & Recent 180K & 47 & $2.39\times 10^{-9}$ & $1.30\times 10^{-8}$ \\
        GPT-5.4 & BM25 & 42 & $1.64\times 10^{-8}$ & $2.16\times 10^{-8}$ \\
        GPT-5.4 & Dense & 48 & $1.63\times 10^{-9}$ & $1.30\times 10^{-8}$ \\
        GPT-5.4 Nano & Recent & 44 & $7.60\times 10^{-9}$ & $1.52\times 10^{-8}$ \\
        GPT-5.4 Nano & Recent 180K & 41 & $2.42\times 10^{-8}$ & $2.79\times 10^{-8}$ \\
        GPT-5.4 Nano & BM25 & 47 & $2.40\times 10^{-9}$ & $1.30\times 10^{-8}$ \\
        GPT-5.4 Nano & Dense & 43 & $1.12\times 10^{-8}$ & $1.77\times 10^{-8}$ \\
        Kimi 2.6 & Recent & 50 & $7.54\times 10^{-10}$ & $1.30\times 10^{-8}$ \\
        Kimi 2.6 & Recent 180K & 42 & $1.65\times 10^{-8}$ & $2.16\times 10^{-8}$ \\
        Kimi 2.6 & BM25 & 40 & $3.57\times 10^{-8}$ & $3.87\times 10^{-8}$ \\
        Kimi 2.6 & Dense & 43 & $1.12\times 10^{-8}$ & $1.77\times 10^{-8}$ \\
        Qwen3.5 397B & Recent & 45 & $5.18\times 10^{-9}$ & $1.51\times 10^{-8}$ \\
        Qwen3.5 397B & Recent 180K & 40 & $3.55\times 10^{-8}$ & $3.87\times 10^{-8}$ \\
        Qwen3.5 397B & BM25 & 42 & $1.65\times 10^{-8}$ & $2.16\times 10^{-8}$ \\
        Qwen3.5 397B & Dense & 46 & $3.52\times 10^{-9}$ & $1.34\times 10^{-8}$ \\
        Qwen3.5 27B & Recent & 38 & $7.72\times 10^{-8}$ & $8.15\times 10^{-8}$ \\
        Qwen3.5 27B & Recent 180K & 43 & $1.12\times 10^{-8}$ & $1.77\times 10^{-8}$ \\
        Qwen3.5 27B & BM25 & 44 & $7.61\times 10^{-9}$ & $1.52\times 10^{-8}$ \\
        Qwen3.5 27B & Dense & 47 & $2.39\times 10^{-9}$ & $1.30\times 10^{-8}$ \\
        \hline
    \end{tabular}
    \caption{Multiple answer versus single reference fact recall by model--inference configuration. Two-sided Wilcoxon signed-rank tests. Each configuration has $N=87$ paired questions; $n_{\ne 0}$ excludes zero differences. Adjusted $p$-values use Benjamini--Hochberg (BH) correction across the 38 tests in this table. $p$ and Adj.\ $p$ are the raw and corrected asymptotic $p$-values. }
    \label{tab:fact-recall-specification-wilcoxon}
\end{table}

\begin{table}[h]
    \centering
    \small
    \begin{tabular}{llccc}
        \hline
        \textbf{Model} & \textbf{Inference} & $n_{\ne 0}$ & $p$ & \textbf{Adj.} $p$ \\
        \hline
        Gemini 2.5 Pro & Recent & 41 & $2.40\times 10^{-8}$ & $2.61\times 10^{-8}$ \\
        Gemini 2.5 Pro & Recent 180K & 49 & $1.09\times 10^{-9}$ & $1.50\times 10^{-9}$ \\
        Gemini 2.5 Pro & BM25 & 46 & $3.51\times 10^{-9}$ & $4.04\times 10^{-9}$ \\
        Gemini 2.5 Pro & Dense & 47 & $2.38\times 10^{-9}$ & $2.84\times 10^{-9}$ \\
        Gemini 2.5 Flash Lite & Recent & 33 & $5.32\times 10^{-7}$ & $5.46\times 10^{-7}$ \\
        Gemini 2.5 Flash Lite & Recent 180K & 39 & $5.22\times 10^{-8}$ & $5.51\times 10^{-8}$ \\
        Gemini 2.5 Flash Lite & BM25 & 45 & $5.13\times 10^{-9}$ & $5.74\times 10^{-9}$ \\
        Gemini 2.5 Flash Lite & Dense & 48 & $1.62\times 10^{-9}$ & $2.12\times 10^{-9}$ \\
        Claude Opus 4.7 & Recent & 67 & $1.12\times 10^{-12}$ & $4.24\times 10^{-11}$ \\
        Claude Opus 4.7 & Recent 180K & 62 & $7.52\times 10^{-12}$ & $1.43\times 10^{-10}$ \\
        Claude Opus 4.7 & BM25 & 58 & $3.50\times 10^{-11}$ & $1.66\times 10^{-10}$ \\
        Claude Opus 4.7 & Dense & 54 & $1.62\times 10^{-10}$ & $3.25\times 10^{-10}$ \\
        Claude Opus 4.7 & Agent & 47 & $2.39\times 10^{-9}$ & $2.84\times 10^{-9}$ \\
        Claude Haiku 4.5 & Recent & 54 & $1.62\times 10^{-10}$ & $3.25\times 10^{-10}$ \\
        Claude Haiku 4.5 & Recent 180K & 26 & $8.29\times 10^{-6}$ & $8.29\times 10^{-6}$ \\
        Claude Haiku 4.5 & BM25 & 51 & $5.13\times 10^{-10}$ & $7.51\times 10^{-10}$ \\
        Claude Haiku 4.5 & Dense & 58 & $3.50\times 10^{-11}$ & $1.66\times 10^{-10}$ \\
        Claude Haiku 4.5 & Agent & 51 & $5.14\times 10^{-10}$ & $7.51\times 10^{-10}$ \\
        GPT-5.4 & Recent & 47 & $2.37\times 10^{-9}$ & $2.84\times 10^{-9}$ \\
        GPT-5.4 & Recent 180K & 51 & $5.13\times 10^{-10}$ & $7.51\times 10^{-10}$ \\
        GPT-5.4 & BM25 & 52 & $3.49\times 10^{-10}$ & $6.03\times 10^{-10}$ \\
        GPT-5.4 & Dense & 55 & $1.10\times 10^{-10}$ & $2.80\times 10^{-10}$ \\
        GPT-5.4 Nano & Recent & 52 & $3.48\times 10^{-10}$ & $6.03\times 10^{-10}$ \\
        GPT-5.4 Nano & Recent 180K & 57 & $5.07\times 10^{-11}$ & $1.77\times 10^{-10}$ \\
        GPT-5.4 Nano & BM25 & 60 & $1.62\times 10^{-11}$ & $1.66\times 10^{-10}$ \\
        GPT-5.4 Nano & Dense & 57 & $5.13\times 10^{-11}$ & $1.77\times 10^{-10}$ \\
        Kimi 2.6 & Recent & 49 & $1.11\times 10^{-9}$ & $1.50\times 10^{-9}$ \\
        Kimi 2.6 & Recent 180K & 57 & $5.11\times 10^{-11}$ & $1.77\times 10^{-10}$ \\
        Kimi 2.6 & BM25 & 52 & $3.49\times 10^{-10}$ & $6.03\times 10^{-10}$ \\
        Kimi 2.6 & Dense & 59 & $2.39\times 10^{-11}$ & $1.66\times 10^{-10}$ \\
        Qwen3.5 397B & Recent & 55 & $1.10\times 10^{-10}$ & $2.80\times 10^{-10}$ \\
        Qwen3.5 397B & Recent 180K & 51 & $5.14\times 10^{-10}$ & $7.51\times 10^{-10}$ \\
        Qwen3.5 397B & BM25 & 58 & $3.49\times 10^{-11}$ & $1.66\times 10^{-10}$ \\
        Qwen3.5 397B & Dense & 54 & $1.62\times 10^{-10}$ & $3.25\times 10^{-10}$ \\
        Qwen3.5 27B & Recent & 54 & $1.62\times 10^{-10}$ & $3.25\times 10^{-10}$ \\
        Qwen3.5 27B & Recent 180K & 56 & $7.52\times 10^{-11}$ & $2.20\times 10^{-10}$ \\
        Qwen3.5 27B & BM25 & 56 & $7.53\times 10^{-11}$ & $2.20\times 10^{-10}$ \\
        Qwen3.5 27B & Dense & 58 & $3.50\times 10^{-11}$ & $1.66\times 10^{-10}$ \\
        \hline
    \end{tabular}
    \caption{Multiple answer versus single reference fact recall by model--inference configuration. Two-sided Wilcoxon signed-rank tests. Each configuration has $N=87$ paired questions; $n_{\ne 0}$ excludes zero differences. Adjusted $p$-values use Benjamini--Hochberg (BH) correction across the 38 tests in this table. $p$ and Adj.\ $p$ are the raw and corrected asymptotic $p$-values.}
    \label{tab:fact-precision-specification-wilcoxon}
\end{table}

%% file: SI/FIG/tables/fact_noninferiority.tex
\begin{table}[h]
 \begin{tabular}{llcccccc}
        \hline
        \textbf{Category} & \textbf{Cohort} & $\widehat{\Delta}$ & \textbf{Upper} & $t$ & $p$ & \textbf{Adj.} $p$ & \textbf{NI} \\
        \hline
        \multicolumn{8}{l}{\textbf{Overall (2 tests)}} \\
        All questions & New & -0.022 & +0.000 & -5.288 & $7.54\times 10^{-8}$ & $7.54\times 10^{-8}$ & Yes \\
         & Unfiltered & +0.005 & +0.014 & -7.789 & $7.65\times 10^{-14}$ & $1.53\times 10^{-13}$ & Yes \\
        \hline
        \multicolumn{8}{l}{\textbf{Reasoning (4 tests)}} \\
        Single Hop & New & -0.019 & +0.012 & -3.634 & $1.52\times 10^{-4}$ & $1.52\times 10^{-4}$ & Yes \\
         & Unfiltered & -0.003 & +0.010 & -6.460 & $7.06\times 10^{-10}$ & $2.82\times 10^{-9}$ & Yes \\
        \\[-5pt]
        Multiple Hops & New & -0.034 & -0.004 & -4.613 & $2.55\times 10^{-6}$ & $5.10\times 10^{-6}$ & Yes \\
         & Unfiltered & +0.013 & +0.026 & -4.598 & $5.74\times 10^{-6}$ & $7.65\times 10^{-6}$ & Yes \\
        \hline
        \multicolumn{8}{l}{\textbf{Time (4 tests)}} \\
        $\leq 180$K tokens & New & +0.002 & +0.024 & -3.576 & $1.84\times 10^{-4}$ & $3.68\times 10^{-4}$ & Yes \\
         & Unfiltered & +0.010 & +0.019 & -7.414 & $1.07\times 10^{-12}$ & $4.29\times 10^{-12}$ & Yes \\
        \\[-5pt]
        $>180$K tokens & New & -0.095 & -0.020 & -3.192 & $9.37\times 10^{-4}$ & $9.77\times 10^{-4}$ & Yes \\
         & Unfiltered & -0.023 & +0.014 & -3.322 & $9.77\times 10^{-4}$ & $9.77\times 10^{-4}$ & Yes \\
        \hline
        \multicolumn{8}{l}{\textbf{Topic (8 tests)}} \\
        Comorbidities & New & -0.011 & +0.018 & -3.404 & $3.55\times 10^{-4}$ & $3.55\times 10^{-4}$ & Yes \\
         & Unfiltered & +0.004 & +0.016 & -6.254 & $4.55\times 10^{-9}$ & $2.83\times 10^{-8}$ & Yes \\
        \\[-5pt]
        Disease Progression Status & New & -0.020 & +0.013 & -3.502 & $2.55\times 10^{-4}$ & $2.92\times 10^{-4}$ & Yes \\
         & Unfiltered & -0.001 & +0.014 & -5.642 & $2.76\times 10^{-7}$ & $5.51\times 10^{-7}$ & Yes \\
        \\[-5pt]
        Diagnostic testing & New & -0.032 & +0.005 & -3.597 & $1.79\times 10^{-4}$ & $2.38\times 10^{-4}$ & Yes \\
         & Unfiltered & -0.006 & +0.009 & -6.172 & $4.65\times 10^{-8}$ & $1.24\times 10^{-7}$ & Yes \\
        \\[-5pt]
        Radiology/Imaging & New & -0.029 & +0.007 & -3.651 & $1.46\times 10^{-4}$ & $2.34\times 10^{-4}$ & Yes \\
         & Unfiltered & -0.002 & +0.010 & -6.829 & $7.08\times 10^{-9}$ & $2.83\times 10^{-8}$ & Yes \\
        \hline
    \end{tabular}
    \caption{Fact recall inferiority testing for $\mathrm{BRIE}_{\mathrm{new}}$ and $\mathrm{BRIE}_{\mathrm{unfiltered}}$ for Qwen3.5 27B, Recent inference. $\widehat{\Delta}$ is the mean recall difference (comparison cohort minus BRIE), on the 0--1 scale. The absolute margin is $\delta=0.05$ (5 percentage points). One-sided Welch tests evaluate $H_0:\Delta\geq\delta$ against $H_1:\Delta<\delta$. Upper is the unadjusted one-sided 95\% confidence bound for $\Delta$. Adj.\ $p$ uses Benjamini--Hochberg correction jointly across both comparison cohorts within each family: overall (2 tests), reasoning (4), time (4), and topics (8). NI is Yes when adjusted $p\leq0.05$; -- means non-inferiority was not established. }
    \label{tab:fact-recall-not-easier}
\end{table}

%% file: SI/FIG/tables/fact_ks.tex

\begin{table}[h]
    \centering
    \small
    \begin{tabular}{lccccc}
        \hline
        \textbf{Category} & $n_{\mathrm{BRIE}}$ & $n_{\mathrm{new}}$ & \textbf{KS} $D$ & $p$ & \textbf{Adj.} $p$ \\
        \hline
        \multicolumn{6}{l}{\textbf{Reasoning (2 tests)}} \\
        Single Hop & 276 & 569 & 0.064 & 0.4029 & 0.4029 \\
        Multiple Hops & 232 & 431 & 0.081 & 0.2592 & 0.4029 \\
        \hline
        \multicolumn{6}{l}{\textbf{Time (2 tests)}} \\
        $\leq 180$K tokens & 445 & 828 & 0.024 & 0.9937 & 0.9937 \\
        $>180$K tokens & 63 & 172 & 0.184 & 0.0762 & 0.1525 \\
        \hline
        \multicolumn{6}{l}{\textbf{Topic (4 tests)}} \\
        Comorbidities & 296 & 636 & 0.053 & 0.5909 & 0.5909 \\
        Disease Progression Status & 188 & 405 & 0.095 & 0.1857 & 0.5909 \\
        Diagnostic testing & 191 & 297 & 0.078 & 0.4436 & 0.5909 \\
        Radiology/Imaging & 187 & 277 & 0.088 & 0.3300 & 0.5909 \\
        \hline
    \end{tabular}
    \caption{Fact Recall: two-sided two-sample Kolmogorov--Smirnov tests comparing $\mathrm{BRIE}$ with $\mathrm{BRIE}_{\mathrm{new}}$ for Qwen3.5 27B, Recent inference. $D$ is the maximum absolute difference between empirical cumulative distributions. Adj.\ $p$ applies Benjamini--Hochberg (BH) correction separately within each plot: two reasoning tests, two time tests, and four topic tests (three correction families)}
    \label{tab:fact-recall-specification-ks}
\end{table}

%% file: MAIN/refs.bib
@article{fleming_medalign_2024,
	title = {{MedAlign}: {A} {Clinician}-{Generated} {Dataset} for {Instruction} {Following} with {Electronic} {Medical} {Records}},
	volume = {38},
	copyright = {Copyright (c) 2024 Association for the Advancement of Artificial Intelligence},
	issn = {2374-3468},
	shorttitle = {{MedAlign}},
	url = {https://ojs.aaai.org/index.php/AAAI/article/view/30205},
	doi = {10.1609/aaai.v38i20.30205},
	language = {en},
	number = {20},
	urldate = {2025-07-07},
	journal = {Proceedings of the AAAI Conference on Artificial Intelligence},
	author = {Fleming, Scott L. and Lozano, Alejandro and Haberkorn, William J. and Jindal, Jenelle A. and Reis, Eduardo and Thapa, Rahul and Blankemeier, Louis and Genkins, Julian Z. and Steinberg, Ethan and Nayak, Ashwin and Patel, Birju and Chiang, Chia-Chun and Callahan, Alison and Huo, Zepeng and Gatidis, Sergios and Adams, Scott and Fayanju, Oluseyi and Shah, Shreya J. and Savage, Thomas and Goh, Ethan and Chaudhari, Akshay S. and Aghaeepour, Nima and Sharp, Christopher and Pfeffer, Michael A. and Liang, Percy and Chen, Jonathan H. and Morse, Keith E. and Brunskill, Emma P. and Fries, Jason A. and Shah, Nigam H.},
	month = mar,
	year = {2024},
	note = {Number: 20},
	pages = {22021--22030},
}

@article{cui_timer_2025,
  author  = {Cui, Hejie and Unell, Alyssa and Chen, Bowen and Fries, Jason Alan and Alsentzer, Emily and Koyejo, Sanmi and Shah, Nigam H.},
  title   = {{TIMER}: temporal instruction modeling and evaluation for longitudinal clinical records},
  journal = {npj Digital Medicine},
  year    = {2025},
  volume  = {8},
  number  = {1},
  pages   = {577},
  doi     = {10.1038/s41746-025-01965-9},
  url     = {https://doi.org/10.1038/s41746-025-01965-9}
}

@article{liu_lost_2023,
    title = "Lost in the Middle: How Language Models Use Long Contexts",
    author = "Liu, Nelson F.  and
      Lin, Kevin  and
      Hewitt, John  and
      Paranjape, Ashwin  and
      Bevilacqua, Michele  and
      Petroni, Fabio  and
      Liang, Percy",
    journal = "Transactions of the Association for Computational Linguistics",
    volume = "12",
    year = "2024",
    address = "Cambridge, MA",
    publisher = "MIT Press",
    url = "https://aclanthology.org/2024.tacl-1.9/",
    doi = "10.1162/tacl_a_00638",
    pages = "157--173"
}

@misc{li_r2med_2025,
	  title={R2MED: A Benchmark for Reasoning-Driven Medical Retrieval}, 
      author={Xiangxu Zhang and Lei Li and Xiao Zhou and Zheng Liu},
      year={2026},
      eprint={2505.14558},
      archivePrefix={arXiv},
      primaryClass={cs.IR},
      url={https://arxiv.org/abs/2505.14558}, 
}

@article{chung_verifact_2025,
	author = {Philip Chung  and Akshay Swaminathan  and Alex J. Goodell  and Yeasul Kim  and S. Momsen Reincke  and Lichy Han  and Ben Deverett  and Mohammad Amin Sadeghi  and Abdel-Badih Ariss  and Marc Ghanem  and David Seong  and Andrew A. Lee  and Caitlin E. Coombes  and Brad Bradshaw  and Mahir A. Sufian  and Hyo Jung Hong  and Teresa P. Nguyen  and Mohammad R. Rasouli  and Komal Kamra  and Mark A. Burbridge  and James C. McAvoy  and Roya Saffary  and Stephen P. Ma  and Dev Dash  and James Xie  and Ellen Y. Wang  and Clifford A. Schmiesing  and Nigam Shah  and Nima Aghaeepour },
title = {Verifying Facts in Patient Care Documents Generated by Large Language Models Using Electronic Health Records},
journal = {NEJM AI},
volume = {3},
number = {1},
pages = {AIdbp2500418},
year = {2026},
doi = {10.1056/AIdbp2500418},

URL = {https://ai.nejm.org/doi/full/10.1056/AIdbp2500418},
eprint = {https://ai.nejm.org/doi/pdf/10.1056/AIdbp2500418}
}

@misc{kweon_ehrnoteqa_2024,
	title = {{EHRNoteQA}: {An} {LLM} {Benchmark} for {Real}-{World} {Clinical} {Practice} {Using} {Discharge} {Summaries}},
	shorttitle = {{EHRNoteQA}},
	url = {http://arxiv.org/abs/2402.16040},
	doi = {10.48550/arXiv.2402.16040},
	urldate = {2025-07-15},
	publisher = {arXiv},
	author = {Kweon, Sunjun and Kim, Jiyoun and Kwak, Heeyoung and Cha, Dongchul and Yoon, Hangyul and Kim, Kwanghyun and Yang, Jeewon and Won, Seunghyun and Choi, Edward},
	month = nov,
	year = {2024},
	note = {arXiv:2402.16040 [cs]},
}

@inproceedings{raman_its_2024,
	title = {Its {All} {Relative}! – {A} {Synthetic} {Query} {Generation} {Approach} for {Improving} {Zero}-{Shot} {Relevance} {Prediction}},
	url = {https://aclanthology.org/2024.findings-naacl.107.pdf},
	booktitle = {Findings of the {Association} for {Computational} {Linguistics}: {NAACL} 2024},
	author = {Raman, Karthik and Bendersky, Michael and Chaudhary, Aditi},
	year = {2024},
}

@misc{chiang_chatbot_2024,
	 title={Chatbot Arena: An Open Platform for Evaluating LLMs by Human Preference}, 
      author={Wei-Lin Chiang and Lianmin Zheng and Ying Sheng and Anastasios Nikolas Angelopoulos and Tianle Li and Dacheng Li and Hao Zhang and Banghua Zhu and Michael Jordan and Joseph E. Gonzalez and Ion Stoica},
      year={2024},
      eprint={2403.04132},
      archivePrefix={arXiv},
      primaryClass={cs.AI},
      url={https://arxiv.org/abs/2403.04132}, 
}

@inproceedings{munnangi_factehr_2024,
  title = 	 {Fact{EHR}: A Dataset for Evaluating Factuality in Clinical Notes Using {LLM}s},
  author =       {Munnangi, Monica and Swaminathan, Akshay and Fries, Jason Alan and Jindal, Jenelle A and Narayanan, Sanjana and Lopez, Ivan and Tu, Lucia and Chung, Philip and Omiye, Jesutofunmi and Kashyap, Mehr and Shah, Nigam},
  booktitle = 	 {Proceedings of the 10th Machine Learning for Healthcare Conference},
  year = 	 {2025},
  editor = 	 {Agrawal, Monica and Deshpande, Kaivalya and Engelhard, Matthew and Joshi, Shalmali and Tang, Shengpu and Urteaga, Iñigo},
  volume = 	 {298},
  series = 	 {Proceedings of Machine Learning Research},
  month = 	 {15--16 Aug},
  publisher =    {PMLR},
  url = 	 {https://proceedings.mlr.press/v298/munnangi25a.html}
}

@book{russell_philosophy_2009,
	address = {Abingdon, Oxon},
	series = {Routledge {Classics}},
	title = {The philosophy of logical atomism},
	isbn = {978-0-415-47461-0},
	language = {en},
	publisher = {Routledge},
	author = {Russell, Bertrand},
	year = {2009},
}

@misc{noauthor_chop_nodate,
        author={Diaz, Naomi},
        month = jul,
	year = {2025},
	title = {{CHOP} creates {AI} agent for {Epic} - {Becker}'s {Hospital} {Review} {\textbar} {Healthcare} {News} \& {Analysis}},
	url = {https://www.beckershospitalreview.com/healthcare-information-technology/ehrs/chop-develops-ai-agent-for-epic/},
	urldate = {2025-09-06},
}

@misc{lynn_deep_2025,
	title = {A {Deep} {Dive} {Into} the {Announcements} at {Epic} {UGM} 2025 {\textbar} {Healthcare} {IT} {Today}},
	url = {https://www.healthcareittoday.com/2025/08/21/a-deep-dive-into-the-announcements-at-epic-ugm-2025/},
	language = {en-US},
	urldate = {2025-09-06},
	author = {Lynn, John},
	month = aug,
	year = {2025},
}

@inproceedings{ronaghi2026clinicallygroundedprivacyevaluation,
title={Clinically Grounded Privacy Evaluation of Medical {LM}s},
author={Sasha Ronaghi and Sana Tonekaboni and Lena Stempfle and Vivian Utti and Jordan Li Cahoon and Nathaniel Hendrix and Marzyeh Ghassemi and Emily Alsentzer},
booktitle={LLM/VLM Deployment Opportunities and Risks in Healthcare},
year={2026},
url={https://openreview.net/forum?id=6dNFO7csJS}
}

@inproceedings{livebench,
  title={LiveBench: A Challenging, Contamination-Free {LLM} Benchmark},
  author={Colin White and Samuel Dooley and Manley Roberts and Arka Pal and Benjamin Feuer and Siddhartha Jain and Ravid Shwartz-Ziv and Neel Jain and Khalid Saifullah and Sreemanti Dey and Shubh-Agrawal and Sandeep Singh Sandha and Siddartha Venkat Naidu and Chinmay Hegde and Yann LeCun and Tom Goldstein and Willie Neiswanger and Micah Goldblum},
  booktitle={The Thirteenth International Conference on Learning Representations},
  year={2025},
}

@misc{tiktoken,
  author = {OpenAI},
  title = {tiktoken: Fast BPE tokeniser for use with OpenAI's models},
  year = {2022},
  publisher = {GitHub},
  journal = {GitHub repository},
  howpublished = {\url{https://github.com/openai/tiktoken}},
}

@misc{armitage_clinicians_2025,
	title = {Clinicians can ‘chat’ with medical records through new {AI} software, {ChatEHR}},
	url = {https://med.stanford.edu/news/all-news/2025/06/chatehr.html},
	language = {en-US},
	urldate = {2025-09-06},
	journal = {News Center},
	author = {Armitage, Hanae},
	month = jun,
	year = {2025},
	note = {Section: Artificial Intelligence (AI)},
}

@techreport{comanici2025gemini25,
  title        = {{Gemini 2.5: Pushing the Frontier with Advanced Reasoning,
                   Multimodality, Long Context, and Next Generation Agentic
                   Capabilities}},
  author       = {Comanici, Gheorghe and Bieber, Eric and Schaekermann, Mike
                   and Pasupat, Ice and Sachdeva, Noveen and Dhillon, Inderjit
                   and Blistein, Marcel and Ram, Ori and Zhang, Dan and
                   Rosen, Evan and Marris, Luke and Petulla, Sam and
                   Gaffney, Colin and Aharoni, Asaf and Lintz, Nathan and
                   {Cardal Pais}, Tiago and Jacobsson, Henrik and
                   Szpektor, Idan and Jiang, Nan-Jiang and Haridasan, Krishna
                   and Omran, Ahmed and Saunshi, Nikunj and others},
  institution  = {Google DeepMind},
  year         = {2025},
  month        = {July},
  note         = {arXiv:2507.06261},
  url          = {https://arxiv.org/abs/2507.06261}
}

@misc{google2026gemini25family,
  title        = {{We're Expanding Our Gemini 2.5 Family of Models}},
  author       = {{Google}},
  year         = {2026},
  month        = {January},
  url          = {https://blog.google/products/gemini/gemini-2-5-model-family-expands/},
  note         = {Accessed June 2026}
}

@misc{singh2026openaigpt5card,
      title={OpenAI GPT-5 System Card}, 
      author={Aaditya Singh and Adam Fry and Adam Perelman and Adam Tart and Adi Ganesh and Ahmed El-Kishky and Aidan McLaughlin and Aiden Low and AJ Ostrow and Akhila Ananthram and Akshay Nathan and Alan Luo and Alec Helyar and Aleksander Madry and Aleksandr Efremov and Aleksandra Spyra and Alex Baker-Whitcomb and Alex Beutel and Alex Karpenko and Alex Makelov and Alex Neitz and Alex Wei and Alexandra Barr and Alexandre Kirchmeyer and Alexey Ivanov and Alexi Christakis and Alistair Gillespie and Allison Tam and Ally Bennett and Alvin Wan and Alyssa Huang and Amy McDonald Sandjideh and Amy Yang and Ananya Kumar and Andre Saraiva and Andrea Vallone and Andrei Gheorghe and Andres Garcia Garcia and Andrew Braunstein and Andrew Liu and Andrew Schmidt and Andrey Mereskin and Andrey Mishchenko and Andy Applebaum and Andy Rogerson and Ann Rajan and Annie Wei and Anoop Kotha and Anubha Srivastava and Anushree Agrawal and Arun Vijayvergiya and Ashley Tyra and Ashvin Nair and Avi Nayak and Ben Eggers and Bessie Ji and Beth Hoover and Bill Chen and Blair Chen and Boaz Barak and Borys Minaiev and Botao Hao and Bowen Baker and Brad Lightcap and Brandon McKinzie and Brandon Wang and Brendan Quinn and Brian Fioca and Brian Hsu and Brian Yang and Brian Yu and Brian Zhang and Brittany Brenner and Callie Riggins Zetino and Cameron Raymond and Camillo Lugaresi and Carolina Paz and Cary Hudson and Cedric Whitney and Chak Li and Charles Chen and Charlotte Cole and Chelsea Voss and Chen Ding and Chen Shen and Chengdu Huang and Chris Colby and Chris Hallacy and Chris Koch and Chris Lu and Christina Kaplan and Christina Kim and CJ Minott-Henriques and Cliff Frey and Cody Yu and Coley Czarnecki and Colin Reid and Colin Wei and Cory Decareaux and Cristina Scheau and Cyril Zhang and Cyrus Forbes and Da Tang and Dakota Goldberg and Dan Roberts and Dana Palmie and Daniel Kappler and Daniel Levine and Daniel Wright and Dave Leo and David Lin and David Robinson and Declan Grabb and Derek Chen and Derek Lim and Derek Salama and Dibya Bhattacharjee and Dimitris Tsipras and Dinghua Li and Dingli Yu and DJ Strouse and Drew Williams and Dylan Hunn and Ed Bayes and Edwin Arbus and Ekin Akyurek and Elaine Ya Le and Elana Widmann and Eli Yani and Elizabeth Proehl and Enis Sert and Enoch Cheung and Eri Schwartz and Eric Han and Eric Jiang and Eric Mitchell and Eric Sigler and Eric Wallace and Erik Ritter and Erin Kavanaugh and Evan Mays and Evgenii Nikishin and Fangyuan Li and Felipe Petroski Such and Filipe de Avila Belbute Peres and Filippo Raso and Florent Bekerman and Foivos Tsimpourlas and Fotis Chantzis and Francis Song and Francis Zhang and Gaby Raila and Garrett McGrath and Gary Briggs and Gary Yang and Giambattista Parascandolo and Gildas Chabot and Grace Kim and Grace Zhao and Gregory Valiant and Guillaume Leclerc and Hadi Salman and Hanson Wang and Hao Sheng and Haoming Jiang and Haoyu Wang and Haozhun Jin and Harshit Sikchi and Heather Schmidt and Henry Aspegren and Honglin Chen and Huida Qiu and Hunter Lightman and Ian Covert and Ian Kivlichan and Ian Silber and Ian Sohl and Ibrahim Hammoud and Ignasi Clavera and Ikai Lan and Ilge Akkaya and Ilya Kostrikov and Irina Kofman and Isak Etinger and Ishaan Singal and Jackie Hehir and Jacob Huh and Jacqueline Pan and Jake Wilczynski and Jakub Pachocki and James Lee and James Quinn and Jamie Kiros and Janvi Kalra and Jasmyn Samaroo and Jason Wang and Jason Wolfe and Jay Chen and Jay Wang and Jean Harb and Jeffrey Han and Jeffrey Wang and Jennifer Zhao and Jeremy Chen and Jerene Yang and Jerry Tworek and Jesse Chand and Jessica Landon and Jessica Liang and Ji Lin and Jiancheng Liu and Jianfeng Wang and Jie Tang and Jihan Yin and Joanne Jang and Joel Morris and Joey Flynn and Johannes Ferstad and Johannes Heidecke and John Fishbein and John Hallman and Jonah Grant and Jonathan Chien and Jonathan Gordon and Jongsoo Park and Jordan Liss and Jos Kraaijeveld and Joseph Guay and Joseph Mo and Josh Lawson and Josh McGrath and Joshua Vendrow and Joy Jiao and Julian Lee and Julie Steele and Julie Wang and Junhua Mao and Kai Chen and Kai Hayashi and Kai Xiao and Kamyar Salahi and Kan Wu and Karan Sekhri and Karan Sharma and Karan Singhal and Karen Li and Kenny Nguyen and Keren Gu-Lemberg and Kevin King and Kevin Liu and Kevin Stone and Kevin Yu and Kristen Ying and Kristian Georgiev and Kristie Lim and Kushal Tirumala and Kyle Miller and Lama Ahmad and Larry Lv and Laura Clare and Laurance Fauconnet and Lauren Itow and Lauren Yang and Laurentia Romaniuk and Leah Anise and Lee Byron and Leher Pathak and Leon Maksin and Leyan Lo and Leyton Ho and Li Jing and Liang Wu and Liang Xiong and Lien Mamitsuka and Lin Yang and Lindsay McCallum and Lindsey Held and Liz Bourgeois and Logan Engstrom and Lorenz Kuhn and Louis Feuvrier and Lu Zhang and Lucas Switzer and Lukas Kondraciuk and Lukasz Kaiser and Manas Joglekar and Mandeep Singh and Mandip Shah and Manuka Stratta and Marcus Williams and Mark Chen and Mark Sun and Marselus Cayton and Martin Li and Marvin Zhang and Marwan Aljubeh and Matt Nichols and Matthew Haines and Max Schwarzer and Mayank Gupta and Meghan Shah and Melody Y. Guan and Melody Huang and Meng Dong and Mengqing Wang and Mia Glaese and Micah Carroll and Michael Lampe and Michael Malek and Michael Sharman and Michael Zhang and Michele Wang and Michelle Pokrass and Mihai Florian and Mikhail Pavlov and Miles Wang and Ming Chen and Mingxuan Wang and Minnia Feng and Mo Bavarian and Molly Lin and Moose Abdool and Mostafa Rohaninejad and Nacho Soto and Natalie Staudacher and Natan LaFontaine and Nathan Marwell and Nelson Liu and Nick Preston and Nick Turley and Nicklas Ansman and Nicole Blades and Nikil Pancha and Nikita Mikhaylin and Niko Felix and Nikunj Handa and Nishant Rai and Nitish Keskar and Noam Brown and Ofir Nachum and Oleg Boiko and Oleg Murk and Olivia Watkins and Oona Gleeson and Pamela Mishkin and Patryk Lesiewicz and Paul Baltescu and Pavel Belov and Peter Zhokhov and Philip Pronin and Phillip Guo and Phoebe Thacker and Qi Liu and Qiming Yuan and Qinghua Liu and Rachel Dias and Rachel Puckett and Rahul Arora and Ravi Teja Mullapudi and Raz Gaon and Reah Miyara and Rennie Song and Rishabh Aggarwal and RJ Marsan and Robel Yemiru and Robert Xiong and Rohan Kshirsagar and Rohan Nuttall and Roman Tsiupa and Ronen Eldan and Rose Wang and Roshan James and Roy Ziv and Rui Shu and Ruslan Nigmatullin and Saachi Jain and Saam Talaie and Sam Altman and Sam Arnesen and Sam Toizer and Sam Toyer and Samuel Miserendino and Sandhini Agarwal and Sarah Yoo and Savannah Heon and Scott Ethersmith and Sean Grove and Sean Taylor and Sebastien Bubeck and Sever Banesiu and Shaokyi Amdo and Shengjia Zhao and Sherwin Wu and Shibani Santurkar and Shiyu Zhao and Shraman Ray Chaudhuri and Shreyas Krishnaswamy and Shuaiqi and Xia and Shuyang Cheng and Shyamal Anadkat and Simón Posada Fishman and Simon Tobin and Siyuan Fu and Somay Jain and Song Mei and Sonya Egoian and Spencer Kim and Spug Golden and SQ Mah and Steph Lin and Stephen Imm and Steve Sharpe and Steve Yadlowsky and Sulman Choudhry and Sungwon Eum and Suvansh Sanjeev and Tabarak Khan and Tal Stramer and Tao Wang and Tao Xin and Tarun Gogineni and Taya Christianson and Ted Sanders and Tejal Patwardhan and Thomas Degry and Thomas Shadwell and Tianfu Fu and Tianshi Gao and Timur Garipov and Tina Sriskandarajah and Toki Sherbakov and Tomek Korbak and Tomer Kaftan and Tomo Hiratsuka and Tongzhou Wang and Tony Song and Tony Zhao and Troy Peterson and Val Kharitonov and Victoria Chernova and Vineet Kosaraju and Vishal Kuo and Vitchyr Pong and Vivek Verma and Vlad Petrov and Wanning Jiang and Weixing Zhang and Wenda Zhou and Wenlei Xie and Wenting Zhan and Wes McCabe and Will DePue and Will Ellsworth and Wulfie Bain and Wyatt Thompson and Xiangning Chen and Xiangyu Qi and Xin Xiang and Xinwei Shi and Yann Dubois and Yaodong Yu and Yara Khakbaz and Yifan Wu and Yilei Qian and Yin Tat Lee and Yinbo Chen and Yizhen Zhang and Yizhong Xiong and Yonglong Tian and Young Cha and Yu Bai and Yu Yang and Yuan Yuan and Yuanzhi Li and Yufeng Zhang and Yuguang Yang and Yujia Jin and Yun Jiang and Yunyun Wang and Yushi Wang and Yutian Liu and Zach Stubenvoll and Zehao Dou and Zheng Wu and Zhigang Wang},
      year={2026},
      eprint={2601.03267},
      archivePrefix={arXiv},
      primaryClass={cs.CL},
      url={https://arxiv.org/abs/2601.03267}, 
}

@misc{openai2026gpt54blog,
  title        = {{Introducing GPT-5.4}},
  author       = {{OpenAI}},
  year         = {2026},
  month        = {March},
  url          = {https://openai.com/index/introducing-gpt-5-4/},
  note         = {Accessed June 2026}
}

@misc{openai2026gpt54nanoblog,
  title        = {{Introducing GPT‑5.4 mini and nano}},
  author       = {{OpenAI}},
  year         = {2026},
  month        = {March},
  url          = {https://openai.com/index/introducing-gpt-5-4-mini-and-nano/},
  note         = {Accessed June 2026}
}

@misc{anthropic2026claudeopus47blog,
  title        = {{Introducing Claude Opus 4.7}},
  author       = {{Anthropic}},
  year         = {2026},
  month        = {April},
  url          = {https://www.anthropic.com/news/claude-opus-4-7},
  note         = {Accessed June 2026}
}

@misc{anthropic2025claudehaiku45blog,
  title        = {{Introducing Claude Haiku 4.5}},
  author       = {{Anthropic}},
  year         = {2025},
  month        = {October},
  url          = {https://www.anthropic.com/news/claude-haiku-4-5},
  note         = {Accessed June 2026}
}

@misc{kimiTeam2025kimik2,
      title={Kimi K2: Open Agentic Intelligence}, 
      author={Kimi Team and Yifan Bai and Yiping Bao and Y. Charles and Cheng Chen and Guanduo Chen and Haiting Chen and Huarong Chen and Jiahao Chen and Ningxin Chen and Ruijue Chen and Yanru Chen and Yuankun Chen and Yutian Chen and Zhuofu Chen and Jialei Cui and Hao Ding and Mengnan Dong and Angang Du and Chenzhuang Du and Dikang Du and Yulun Du and Yu Fan and Yichen Feng and Kelin Fu and Bofei Gao and Chenxiao Gao and Hongcheng Gao and Peizhong Gao and Tong Gao and Yuyao Ge and Shangyi Geng and Qizheng Gu and Xinran Gu and Longyu Guan and Haiqing Guo and Jianhang Guo and Xiaoru Hao and Tianhong He and Weiran He and Wenyang He and Yunjia He and Chao Hong and Hao Hu and Yangyang Hu and Zhenxing Hu and Weixiao Huang and Zhiqi Huang and Zihao Huang and Tao Jiang and Zhejun Jiang and Xinyi Jin and Yongsheng Kang and Guokun Lai and Cheng Li and Fang Li and Haoyang Li and Ming Li and Wentao Li and Yang Li and Yanhao Li and Yiwei Li and Zhaowei Li and Zheming Li and Hongzhan Lin and Xiaohan Lin and Zongyu Lin and Chengyin Liu and Chenyu Liu and Hongzhang Liu and Jingyuan Liu and Junqi Liu and Liang Liu and Shaowei Liu and T. Y. Liu and Tianwei Liu and Weizhou Liu and Yangyang Liu and Yibo Liu and Yiping Liu and Yue Liu and Zhengying Liu and Enzhe Lu and Haoyu Lu and Lijun Lu and Yashuo Luo and Shengling Ma and Xinyu Ma and Yingwei Ma and Shaoguang Mao and Jie Mei and Xin Men and Yibo Miao and Siyuan Pan and Yebo Peng and Ruoyu Qin and Zeyu Qin and Bowen Qu and Zeyu Shang and Lidong Shi and Shengyuan Shi and Feifan Song and Jianlin Su and Zhengyuan Su and Lin Sui and Xinjie Sun and Flood Sung and Yunpeng Tai and Heyi Tang and Jiawen Tao and Qifeng Teng and Chaoran Tian and Chensi Wang and Dinglu Wang and Feng Wang and Hailong Wang and Haiming Wang and Jianzhou Wang and Jiaxing Wang and Jinhong Wang and Shengjie Wang and Shuyi Wang and Si Wang and Xinyuan Wang and Yao Wang and Yejie Wang and Yiqin Wang and Yuxin Wang and Yuzhi Wang and Zhaoji Wang and Zhengtao Wang and Zhengtao Wang and Zhexu Wang and Chu Wei and Qianqian Wei and Haoning Wu and Wenhao Wu and Xingzhe Wu and Yuxin Wu and Chenjun Xiao and Jin Xie and Xiaotong Xie and Weimin Xiong and Boyu Xu and Jinjing Xu and L. H. Xu and Lin Xu and Suting Xu and Weixin Xu and Xinran Xu and Yangchuan Xu and Ziyao Xu and Jing Xu and Jing Xu and Junjie Yan and Yuzi Yan and Hao Yang and Xiaofei Yang and Yi Yang and Ying Yang and Zhen Yang and Zhilin Yang and Zonghan Yang and Haotian Yao and Xingcheng Yao and Wenjie Ye and Zhuorui Ye and Bohong Yin and Longhui Yu and Enming Yuan and Hongbang Yuan and Mengjie Yuan and Siyu Yuan and Haobing Zhan and Dehao Zhang and Hao Zhang and Wanlu Zhang and Xiaobin Zhang and Yadong Zhang and Yangkun Zhang and Yichi Zhang and Yizhi Zhang and Yongting Zhang and Yu Zhang and Yutao Zhang and Yutong Zhang and Zheng Zhang and Haotian Zhao and Yikai Zhao and Zijia Zhao and Huabin Zheng and Shaojie Zheng and Longguang Zhong and Jianren Zhou and Xinyu Zhou and Zaida Zhou and Jinguo Zhu and Zhen Zhu and Weiyu Zhuang and Xinxing Zu},
      year={2026},
      eprint={2507.20534},
      archivePrefix={arXiv},
      primaryClass={cs.LG},
      url={https://arxiv.org/abs/2507.20534}, 
}

@misc{moonshot2026kimik26,
  title        = {Kimi K2.6: Advancing Open-Source Coding},
  author       = {Moonshot AI},
  year         = {2026},
  month        = {April},
  url          = {https://www.kimi.com/blog/kimi-k2-6},
  note         = {Released April 20, 2026. Model card references arXiv:2602.02276.
                  Weights: https://huggingface.co/moonshotai/Kimi-K2.6}
}

@misc{qwen3.5,
    title  = {{Qwen3.5}: Towards Native Multimodal Agents},
    author = {{Qwen Team}},
    month  = {February},
    year   = {2026},
    url    = {https://qwen.ai/blog?id=qwen3.5}
}

@misc{octenTeam2025rteb,
  title        = {{Octen Series: Optimizing Embedding Models to \#1 on RTEB
                   Leaderboard}},
  author       = {{Octen Team}},
  year         = {2025},
  url          = {https://octen-team.github.io/octen_blog/posts/octen-rteb-first-place/},
  note         = {Blog post. Models available at https://huggingface.co/bflhc
                  (Octen-Embedding-8B, 4B, 0.6B). Accessed June 2026.}
}

@misc{octen2025embedding8b,
  title        = {{Octen-Embedding-8B}},
  author       = {{Octen Team}},
  year         = {2025},
  howpublished = {Hugging Face Model Card},
  url          = {https://huggingface.co/Octen/Octen-Embedding-8B},
  note         = {Fine-tuned from Qwen3-Embedding-8B; ranks \#1 on RTEB
                  (Mean Task score 0.8045). Accessed June 2026.}
}

@inproceedings{robertson1994bm25,
  author       = {Robertson, Stephen E. and Walker, Steve},
  title        = {{Some Simple Effective Approximations to the 2-Poisson Model
                   for Probabilistic Weighted Retrieval}},
  booktitle    = {Proceedings of the 17th Annual International {ACM} {SIGIR}
                   Conference on Research and Development in Information
                   Retrieval},
  series       = {SIGIR~'94},
  pages        = {232--241},
  year         = {1994},
  publisher    = {ACM/Springer},
  address      = {Dublin, Ireland}
}

@article{robertson2009bm25beyond,
  author       = {Robertson, Stephen E. and Zaragoza, Hugo},
  title        = {{The Probabilistic Relevance Framework: {BM25} and Beyond}},
  journal      = {Foundations and Trends in Information Retrieval},
  volume       = {3},
  number       = {4},
  pages        = {333--389},
  year         = {2009},
  publisher    = {Now Publishers},
  doi          = {10.1561/1500000019}
}

@inproceedings{robertson1994okapitrec3,
  author       = {Robertson, Stephen E. and Walker, Steve and
                   Jones, Susan and Hancock-Beaulieu, Micheline M. and
                   Gatford, Mike},
  title        = {{Okapi at {TREC}-3}},
  booktitle    = {Proceedings of the Third Text {RE}trieval Conference
                   ({TREC}-3)},
  year         = {1994},
  publisher    = {NIST},
  address      = {Gaithersburg, MD, USA},
  note         = {NIST Special Publication 500-225}
}

@inproceedings{yang2018hotpotqa,
  title={{HotpotQA}: A Dataset for Diverse, Explainable Multi-hop Question Answering},
  author={Yang, Zhilin and Qi, Peng and Zhang, Saizheng and Bengio, Yoshua and Cohen, William W. and Salakhutdinov, Ruslan and Manning, Christopher D.},
  booktitle={Conference on Empirical Methods in Natural Language Processing ({EMNLP})},
  year={2018}
}

@article{grolleau2026medfacteval,
  author    = {Grolleau, F. and Alsentzer, E. and Keyes, T. and Chung, P. and Swaminathan, A. and Aali, A. and Hom, J. and Huynh, T. and Lew, T. and Liang, A. and Chu, W. and Steele, N. and Lin, C. and Yang, J. and Black, K. and Ma, S. and Haredasht, F. N. and Shah, N. H. and Schulman, K. and Chen, J. H.},
  title     = {{MedFactEval and MedAgentBrief: A Framework and Workflow for Generating and Evaluating Factual Clinical Summaries}},
  journal   = {Pacific Symposium on Biocomputing},
  year      = {2026},
  volume    = {31},
  pages     = {388--399},
  doi       = {10.1142/9789819824755_0027},
  pmid      = {41758155},
  pmcid     = {PMC13182771},
}

@misc{cinarkoras2026configurableclinicalinformationextraction,
      title={Configurable Clinical Information Extraction with Agentic RAG: What Works, What Breaks, and Why}, 
      author={Osman Alperen Çinar-Koraş and Marie Bauer and Sameh Khattab and Merlin Engelke and Moon Kim and Stephan Settelmeier and Shigeyasu Sugawara and Fabian Freisleben and Felix Nensa and Jens Kleesiek},
      year={2026},
      eprint={2606.19602},
      archivePrefix={arXiv},
      primaryClass={cs.AI},
      url={https://arxiv.org/abs/2606.19602}, 
}

@misc{niu2026aipatient,
  title={AIPatient Arena: EHR-grounded evaluation of large language models in end-to-end clinical consultation workflows},
  author={Niu, Jiahui and Yu, Huizi and Wang, Wenkong and Dai, Guangxin and He, Jingxian and Li, Xiang and Liang, Zhiying and Lin, Xinxin and So, Kent CY and Yan, Bryan YP and Wing, Yun Kwok and Xing, Yanqiu and Ma, Xin and Fan, Lizhou},
  year={2026},
  eprint={2606.17474},
  archivePrefix={arXiv},
  primaryClass={cs.CL},
  url={https://arxiv.org/abs/2606.17474}
}

@misc{ucsf_ars_brim_2026,
  author       = {{UCSF Academic Research Services}},
  title        = {Introducing {BRIM} at {UCSF}: Accelerate Chart Abstraction with {AI} (Pilot Now Open)},
  url= {https://ars.ucsf.edu/news/introducing-brim-ucsf-accelerate-chart-abstraction-ai-pilot-now-open},
  year         = {2026},
  month        = apr,
  day          = {9},
  note         = {Accessed: 2026-06-25}
}

@article{lopez2025clinical,
  title   = {Clinical entity augmented retrieval for clinical information extraction},
  author  = {Lopez, Ivan and Swaminathan, Akshay and Vedula, Karthik and others},
  journal = {npj Digital Medicine},
  volume  = {8},
  number  = {1},
  pages   = {45},
  year    = {2025},
  doi     = {10.1038/s41746-024-01377-1},
  url     = {https://doi.org/10.1038/s41746-024-01377-1}
}

@inproceedings{lewis2020retrieval,
  title     = {Retrieval-Augmented Generation for Knowledge-Intensive {NLP} Tasks},
  author    = {Lewis, Patrick and Perez, Ethan and Piktus, Aleksandra and Petroni, Fabio and Karpukhin, Vladimir and Goyal, Naman and K{\"u}ttler, Heinrich and Lewis, Mike and Yih, Wen-tau and Rockt{\"a}schel, Tim and Riedel, Sebastian and Kiela, Douwe},
  booktitle = {Advances in Neural Information Processing Systems},
  volume    = {33},
  pages     = {9459--9474},
  year      = {2020},
  url       = {https://proceedings.neurips.cc/paper/2020/file/6b493230205f780e1bc26945df7481e5-Paper.pdf}
}

@article{qu2026trace,
  title        = {{TRACE}: Temporal Reasoning via Agentic Context Evolution
                  for Streaming Electronic Health Records (EHRs)},
  author       = {Qu, Zhan and F{\"a}rber, Michael},
  journal      = {arXiv preprint arXiv:2602.12833},
  year         = {2026},
  eprint       = {2602.12833},
  archivePrefix = {arXiv},
  primaryClass = {cs.CL},
  url          = {https://arxiv.org/abs/2602.12833}
}

@article{zhang2024agentic,
  title        = {Agentic Information Retrieval},
  author       = {Zhang, Weinan and Liao, Junwei and Li, Ning and
                  Du, Kounianhua and Lin, Jianghao},
  journal      = {arXiv preprint arXiv:2410.09713},
  year         = {2024},
  eprint       = {2410.09713},
  archivePrefix = {arXiv},
  primaryClass = {cs.IR},
  url          = {https://arxiv.org/abs/2410.09713}
}

@article{singhal2023llmclinical,
  title   = {Large language models encode clinical knowledge},
  author  = {Singhal, Karan and Azizi, Shekoofeh and Tu, Tao and Mahdavi, S. Sara
             and Wei, Jason and Chung, Hyung Won and Scales, Nathan and Tanwani, Ajay
             and Cole-Lewis, Heather and Pfohl, Stephen and Payne, Perry
             and Seneviratne, Martin and Gamble, Paul and Kelly, Chris and Babiker, Abubakr
             and Sch{\"a}rli, Nathanael and Chowdhery, Aakanksha and Mansfield, Philip
             and Demner-Fushman, Dina and Ag{\"u}era y Arcas, Blaise and Webster, Dale
             and Corrado, Greg S. and Matias, Yossi and Chou, Katherine and Gottweis, Juraj
             and Tomasev, Nenad and Liu, Yun and Rajkomar, Alvin and Barral, Joelle
             and Semturs, Christopher and Karthikesalingam, Alan and Natarajan, Vivek},
  journal = {Nature},
  volume  = {620},
  number  = {7972},
  pages   = {172--180},
  year    = {2023},
  doi     = {10.1038/s41586-023-06291-2}
}

@inproceedings{yang2025ehrstruct,
title={EHRStruct: A Comprehensive Benchmark Framework for Evaluating Large Language Models on Structured Electronic Health Record Tasks},
  author={Yang, Xiao and Zhao, Xuejiao and Shen, Zhiqi},
  booktitle={Proceedings of the AAAI Conference on Artificial Intelligence},
  volume={40},
  number={40},
  pages={34340--34348},
  year={2026}
}

@inproceedings{yan2026livemedbench,
author = {Yan, Zhiling and Song, Dingjie and Fang, Zhe and Ji, Yisheng and Li, Xiang and Li, Quanzheng and Sun, Lichao},
title = {LiveMedBench: A Contamination-Limited Medical Benchmark for LLMs with Automated Rubric Evaluation},
year = {2026},
isbn = {9798400722592},
publisher = {Association for Computing Machinery},
address = {New York, NY, USA},
url = {https://doi-org.stanford.idm.oclc.org/10.1145/3770855.3817579},
doi = {10.1145/3770855.3817579},
booktitle = {Proceedings of the 32nd ACM SIGKDD Conference on Knowledge Discovery and Data Mining V.2},
pages = {10162–10173},
numpages = {12},
location = {Republic of Korea},
series = {KDD '26}
}

@misc{cahoon2026clinicalnotebloatreduction,
      title={Clinical Note Bloat Reduction for Efficient LLM Use}, 
      author={Jordan L. Cahoon and Chloe Stanwyck and Asad Aali and Rachel Madding and Emma Sun and Yixing Jiang and Renumathy Dhanasekaran and Emily Alsentzer},
      year={2026},
      eprint={2604.16364},
      archivePrefix={arXiv},
      primaryClass={cs.CY},
      url={https://arxiv.org/abs/2604.16364}, 
}

@article{hirschtick_copy_paste_2006,
  author  = {Hirschtick, Robert E.},
  title   = {A piece of my mind. Copy-and-Paste},
  journal = {JAMA},
  year    = {2006},
  volume  = {295},
  number  = {20},
  pages   = {2335--2336}
}

@article{weis_copy_2014,
	title = {Copy, {Paste}, and {Cloned} {Notes} in {Electronic} {Health} {Records}},
	volume = {145},
	issn = {0012-3692},
	url = {https://www.sciencedirect.com/science/article/pii/S0012369215343786},
	doi = {10.1378/chest.13-0886},
	number = {3},
	urldate = {2026-01-20},
	journal = {Chest},
	author = {Weis, Justin M. and Levy, Paul C.},
	month = mar,
	year = {2014},
	pages = {632--638},
}

@article{kanithi2024medic,
  title={{MEDIC}: Comprehensive Evaluation of Leading Indicators for {LLM} Safety and Utility in Clinical Applications},
author={Praveenkumar Kanithi and Clement Christophe and Marco AF Pimentel and Tathagata Raha and Prateek Munjal and Nada Saadi and Hamza A Javed and Svetlana Maslenkova and Nasir Hayat and Ronnie Rajan and Shadab Khan},
journal={Transactions on Machine Learning Research},
issn={2835-8856},
year={2026},
url={https://openreview.net/forum?id=pDQe9Icwb6},
note={}
}

@misc{ravichandran2025healthbench,
  title         = {{OpenAI's HealthBench} in Action: Evaluating an {LLM}-Based Medical Assistant on Realistic Clinical Queries},
  author        = {Ravichandran, Sandhanakrishnan and Kumar, Shivesh and Corga Da Silva, Rog{\'e}rio and Romano, Miguel and Berkels, Reinhard and van der Heijden, Michiel and Fail, Olivier and Gnanapragasam, Valentine Emmanuel},
  year          = {2025},
  eprint        = {2509.02594},
  archivePrefix = {arXiv},
  primaryClass  = {cs.CL},
  url           = {https://arxiv.org/abs/2509.02594}
}

@article{artsi2025workflows,
  title   = {Large language models in real-world clinical workflows: a systematic review of applications and implementation},
  author  = {Artsi, Yaara and Sorin, Vera and Glicksberg, Benjamin S. and Korfiatis, Panagiotis and Nadkarni, Girish N. and Klang, Eyal},
  journal = {Frontiers in Digital Health},
  volume  = {7},
  pages   = {1659134},
  year    = {2025},
  doi     = {10.3389/fdgth.2025.1659134}
}

@article{pan2025beyond,
  author  = {Pan, Jiazhen and Jian, Bailiang and Hager, Paul and Zhang, Yundi and Liu, Che and Jungmann, Friedrike and Li, Hongwei Bran and Canisius, Julian and You, Chenyu and Wu, Junde and Zhu, Jiayuan and Liu, Fenglin and Liu, Yuyuan and Bubeck, Niklas and Knolle, Moritz and Chen, Chen and Wachinger, Christian and Gong, Zhenyu and Ouyang, Cheng and Kaissis, Georgios and Wiestler, Benedikt and Rueckert, Daniel},
  title   = {Addressing benchmarking gaps in large language models for health and medicine with dynamic red-teaming},
  journal = {Nature Health},
  year    = {2026},
  month   = jul,
  doi     = {10.1038/s44360-026-00152-8},
  url     = {https://doi.org/10.1038/s44360-026-00152-8}
}

@misc{dsouza2025automating,
  title         = {Automating Benchmark Design},
  author        = {Dsouza, Amanda and Vishwakarma, Harit and Qi, Zhengyang and Bauer, Justin and Pham, Derek and Walshe, Thomas and Parchami, Armin and Sala, Frederic and Varma, Paroma},
  year          = {2025},
  eprint        = {2510.25039},
  archivePrefix = {arXiv},
  primaryClass  = {cs.LG},
  url           = {https://arxiv.org/abs/2510.25039}
}

@inproceedings{akhtar2026plateau,
  title     = {When {AI} Benchmarks Plateau: A Systematic Study of Benchmark Saturation},
  author    = {Akhtar, Mubashara and Reuel, Anka and Soni, Prajna and Ahuja, Sanchit and Ammanamanchi, Pawan Sasanka and Rawal, Ruchit and Zouhar, Vil{\'e}m and Yadav, Srishti and Whitehouse, Chenxi and Ki, Dayeon and Mickel, Jennifer and Choshen, Leshem and {\v{S}}uppa, Marek and Batzner, Jan and Chim, Jenny and Sania, Jeba and Long, Yanan and Rahmani, Hossein A. and Knight, Christina and Nan, Yiyang and Raj, Jyoutir and Fan, Yu and Singh, Shubham and Sahoo, Subramanyam and Habba, Eliya and Gohar, Usman and Pawar, Siddhesh and Scholz, Robert and Subramonian, Arjun and Ni, Jingwei and Kochenderfer, Mykel and Koyejo, Sanmi and Sachan, Mrinmaya and Biderman, Stella and Talat, Zeerak and Ghosh, Avijit and Solaiman, Irene},
booktitle={Forty-third International Conference on Machine Learning},
year={2026},
url={https://openreview.net/forum?id=YC1Otscjbs}
}

@online{ruder2024evolving,
  title   = {The Evolving Landscape of {LLM} Evaluation},
  author  = {Ruder, Sebastian},
  year    = {2024},
  month   = may,
  url     = {https://newsletter.ruder.io/p/the-evolving-landscape-of-llm-evaluation},
  note    = {Accessed: 2026-06-27}
}

@article{shool2025systematic,
  title   = {A systematic review of large language model ({LLM}) evaluations in clinical medicine},
  author  = {Shool, Sina and Adimi, Sara and Saboori Amleshi, Reza and Bitaraf, Ehsan and Golpira, Reza and Tara, Mahmood},
  journal = {BMC Medical Informatics and Decision Making},
  volume  = {25},
  number  = {1},
  pages   = {117},
  year    = {2025},
  doi     = {10.1186/s12911-025-02954-4}
}

@article{artsi2025challenges,
  title   = {Challenges of Implementing {LLMs} in Clinical Practice: Perspectives},
  author  = {Artsi, Yaara and Sorin, Vera and Glicksberg, Benjamin S. and Korfiatis, Panagiotis and Freeman, Robert and Nadkarni, Girish N. and Klang, Eyal},
  journal = {Journal of Clinical Medicine},
  volume  = {14},
  number  = {17},
  pages   = {6169},
  year    = {2025},
  doi     = {10.3390/jcm14176169}
}

@article{ebbers2022impact,
  title   = {The Impact of Structured and Standardized Documentation on Documentation Quality: A Multicenter, Retrospective Study},
  author  = {Ebbers, Tom and Kool, Rudolf B. and Smeele, Ludi E. and Dirven, Richard and den Besten, Catharina A. and Karssemakers, Luc H. E. and Verhoeven, Tim and Herruer, Jasper M. and van den Broek, Guido B. and Takes, Robert P.},
  journal = {Journal of Medical Systems},
  volume  = {46},
  number  = {7},
  pages   = {46},
  year    = {2022},
  doi     = {10.1007/s10916-022-01837-9},
  pmid    = {35618978},
  pmcid   = {PMC9135789}
}

@inproceedings{ahsan2024retrieving,
   author  = {Ahsan, Hiba and McInerney, Denis Jered and Kim, Jisoo and Potter, Christopher and Young, Geoffrey and Amir, Silvio and Wallace, Byron C},
  title   = {Retrieving Evidence from {EHRs} with {LLMs}: Possibilities and Challenges},
  journal = {Proceedings of Machine Learning Research},
  year    = {2024},
  month   = jun,
  volume  = {248},
  pages   = {489--505},
  pmid    = {39224857},
  pmcid   = {PMC11368037},
  url     = {https://pubmed.ncbi.nlm.nih.gov/39224857/}
}

@inproceedings{nahum2025llms,
  title     = {Are {LLM}s Better than Reported? Detecting Label Errors and Mitigating Their Effect on Model Performance},
  author    = {Nahum, Omer and Calderon, Nitay and Keller, Orgad and Szpektor, Idan and Reichart, Roi},
  booktitle = {Proceedings of the 2025 Conference on Empirical Methods in Natural Language Processing (EMNLP)},
  pages     = {26782--26809},
  year      = {2025},
  month     = nov,
  address   = {Suzhou, China},
  publisher = {Association for Computational Linguistics},
  doi       = {10.18653/v1/2025.emnlp-main.1360},
  url       = {https://aclanthology.org/2025.emnlp-main.1360/}
}

@inproceedings{pandit2025medhallucomprehensivebenchmarkdetecting,
      title = "{M}ed{H}allu: A Comprehensive Benchmark for Detecting Medical Hallucinations in Large Language Models",
    author = "Pandit, Shrey  and
      Xu, Jiawei  and
      Hong, Junyuan  and
      Wang, Zhangyang  and
      Chen, Tianlong  and
      Xu, Kaidi  and
      Ding, Ying",
    editor = "Christodoulopoulos, Christos  and
      Chakraborty, Tanmoy  and
      Rose, Carolyn  and
      Peng, Violet",
    booktitle = "Proceedings of the 2025 Conference on Empirical Methods in Natural Language Processing",
    month = nov,
    year = "2025",
    address = "Suzhou, China",
    publisher = "Association for Computational Linguistics",
    url = "https://aclanthology.org/2025.emnlp-main.143/",
    doi = "10.18653/v1/2025.emnlp-main.143",
    pages = "2858--2873",
    ISBN = "979-8-89176-332-6"
}

@article{asgari2025framework,
  title   = {A framework to assess clinical safety and hallucination rates of {LLMs} for medical text summarisation},
  author  = {Asgari, E. and Monta{\~n}a-Brown, N. and Dubois, M. and others},
  journal = {npj Digital Medicine},
  volume  = {8},
  number  = {1},
  pages   = {274},
  year    = {2025},
  doi     = {10.1038/s41746-025-01670-7},
  url     = {https://doi.org/10.1038/s41746-025-01670-7}
}

@article{shah2024accuracy,
  title   = {Accuracy, Consistency, and Hallucination of Large Language Models When Analyzing Unstructured Clinical Notes in Electronic Medical Records},
  author  = {Shah, Savyasachi V.},
  journal = {JAMA Network Open},
  volume  = {7},
  number  = {8},
  pages   = {e2425953},
  year    = {2024},
  doi     = {10.1001/jamanetworkopen.2024.25953},
  url     = {https://jamanetwork.com/journals/jamanetworkopen/fullarticle/2822301}
}

@inproceedings{wornow2025contextcluesevaluatinglong,
      title={Context Clues: Evaluating Long Context Models for Clinical Prediction Tasks on {EHR} Data},
author={Michael Wornow and Suhana Bedi and Miguel Angel Fuentes Hernandez and Ethan Steinberg and Jason Alan Fries and Christopher Re and Sanmi Koyejo and Nigam Shah},
booktitle={The Thirteenth International Conference on Learning Representations},
year={2025},
url={https://openreview.net/forum?id=zg3ec1TdAP}
}

@misc{chen2025buildingehrfoundationmodel,
      title={Building the EHR Foundation Model via Next Event Prediction}, 
      author={Zekai Chen and Arda Pekis and Kevin Brown},
      year={2025},
      eprint={2509.25591},
      archivePrefix={arXiv},
      primaryClass={cs.AI},
      url={https://arxiv.org/abs/2509.25591}, 
}

@article{griot2025implementation,
  title   = {Implementation of large language models in electronic health records},
  author  = {Griot, M. and Vanderdonckt, J. and Yuksel, D.},
  journal = {PLOS Digital Health},
  volume  = {4},
  number  = {12},
  pages   = {e0001141},
  year    = {2025},
  doi     = {10.1371/journal.pdig.0001141},
  pmid    = {41417848},
  pmcid   = {PMC12716761}
}

@misc{taveekitworachai2026robustnessanswerformatsmedical,
      title={On the Robustness of Answer Formats in Medical Reasoning Models}, 
      author={Pittawat Taveekitworachai and Natpatchara Pongjirapat and Krittaphas Chaisutyakorn and Piyalitt Ittichaiwong and Tossaporn Saengja and Kunat Pipatanakul},
      year={2026},
      eprint={2509.20866},
      archivePrefix={arXiv},
      primaryClass={cs.CL},
      url={https://arxiv.org/abs/2509.20866}, 
}

@inproceedings{hosseini2024benchmarklongformmedicalquestion,
      title={A Benchmark for Long-Form Medical Question Answering},
author={Pedram Hosseini and Jessica M. Sin and Bing Ren and Bryceton G. Thomas and Elnaz Nouri and Ali Farahanchi and Saeed Hassanpour},
booktitle={Advancements In Medical Foundation Models: Explainability, Robustness, Security, and Beyond},
year={2024},
url={https://openreview.net/forum?id=8Qba6OeW9a}
}

@inproceedings{zheng2023judging,
author = {Zheng, Lianmin and Chiang, Wei-Lin and Sheng, Ying and Zhuang, Siyuan and Wu, Zhanghao and Zhuang, Yonghao and Lin, Zi and Li, Zhuohan and Li, Dacheng and Xing, Eric P. and Zhang, Hao and Gonzalez, Joseph E. and Stoica, Ion},
title = {Judging LLM-as-a-judge with MT-bench and Chatbot Arena},
year = {2023},
publisher = {Curran Associates Inc.},
address = {Red Hook, NY, USA},
booktitle = {Proceedings of the 37th International Conference on Neural Information Processing Systems},
articleno = {2020},
numpages = {29},
location = {New Orleans, LA, USA},
series = {NIPS '23}
}

@article{Bedi2026,
  title     = {Holistic evaluation of large language models for medical tasks},
  author    = {Bedi, Suhana and Fuentes Hernandez, Miguel Angel and Unell, Alyssa and Cui, Hejie and Wornow, Michael and Swaminathan, Akshay and Kashyap, Mehr and Chung, Philip and Nateghi, Fateme and Jain, Shrey and Oez, Mert and Qiu, Hao and Schettini, Leonardo and Yim, Wen-wai and Lungren, Matthew and Daneshjou, Roxana and Chen, Jonathan and Alsentzer, Emily and Morse, Keith and Ravi, Nirmal and Aghaeepour, Nima and Kennedy, Vanessa and Koyejo, Sanmi and Horvitz, Eric and Chen, James H. and others},
  journal   = {Nature Medicine},
  year      = {2026},
  doi       = {10.1038/s41591-025-04151-2},
  url       = {https://doi.org/10.1038/s41591-025-04151-2}
}

@misc{saab2024capabilities,
  title={Capabilities of Gemini Models in Medicine},
  author={Saab, Khaled and Tu, Tao and Weng, Wei-Hung and Tanno, Ryutaro and Stutz, David and Wulczyn, Ellery and others},
  year={2024},
  eprint={2404.18416},
  archivePrefix={arXiv},
  primaryClass={cs.AI}
}

@misc{sourty2026denseonlateon,
  title={DenseOn with the LateOn: Open State-of-the-Art Single and Multi-Vector Models},
  author={Sourty, Raphael and Chaffin, Antoine and Weller, Orion and Demoura, Paulo and Chatelain, Amelie},
  year={2026},
  howpublished={\url{https://huggingface.co/blog/lightonai/denseon-lateon}},
}

@misc{moll2026agenticclinicalreasoninglongitudinal,
      title={Agentic clinical reasoning over longitudinal myeloma records: a retrospective evaluation against expert consensus}, 
      author={Johannes Moll and Jannik Lübberstedt and Christoph Nuernbergk and Jacob Stroh and Luisa Mertens and Anna Purcarea and Christopher Zirn and Zeineb Benchaaben and Fabian Drexel and Hartmut Häntze and Anirudh Narayanan and Friedrich Puttkammer and Andrei Zhukov and Jacqueline Lammert and Sebastian Ziegelmayer and Markus Graf and Marion Högner and Marcus Makowski and Florian Bassermann and Lisa C. Adams and Jiazhen Pan and Daniel Rueckert and Krischan Braitsch and Keno K. Bressem},
      year={2026},
      eprint={2604.24473},
      archivePrefix={arXiv},
      primaryClass={cs.AI},
      url={https://arxiv.org/abs/2604.24473}, 
}

@misc{gao2026scout,
  title         = {A Randomized Controlled Trial and Pilot of Scout: an LLM-Based EHR Search and Synthesis Platform},
  author        = {Michael Gao and Suresh Balu and William Knechtle and Kartik Pejavara and William Jeck and Matthew Ellis and Jason Thieling and Blake Cameron and Jason Tatreau and Tareq Aljurf and Henry Foote and Michael Revoir and Marshall Nichols and Matthew Gardner and William Ratliff and Bradley Hintze and Angelo Milazzo and Sreekanth Vemulapalli},
  year          = {2026},
  eprint        = {2604.26953},
  archivePrefix = {arXiv},
  primaryClass  = {cs.IR},
  doi           = {10.48550/arXiv.2604.26953},
  url           = {https://arxiv.org/abs/2604.26953}
}

@misc{shah2026adoption,
  title         = {Adoption and Use of LLMs at an Academic Medical Center},
  author        = {Nigam H. Shah and Nerissa Ambers and Abby Pandya and Timothy Keyes and Juan M. Banda and Srikar Nallan and Carlene Lugtu and Artem A. Trotsyuk and Suhana Bedi and Alyssa Unell and Miguel Fuentes and Francois Grolleau and Sneha S. Jain and Jonathan Chen and Devdutta Dash and Danton Char and Aditya Sharma and Duncan McElfresh and Patrick Scully and Vishanthan Kumar and Clancy Dennis and Connor OBrien and Satchi Mouniswamy and Elvis Jones and Krishna Jasti and Gunavathi Mannika Lakshmanan and Sree Ram Akula and Varun Kumar Singh and Ramesh Rajmanickam and Sudhir Sinha and Vicky Zhou and Xu Wang and Bilal Mawji and Joshua Ge and Wencheng Li and Travis Lyons and Jarrod Helzer and Vikas Kakkar and Ramesh Powar and Darren Batara and Cheryl Cordova and William Frederick III and Olivia Tang and Phoebe Morgan and April S. Liang and Stephen P. Ma and Shivam Vedak and Dong-han Yao and Akshay Swaminathan and Mehr Kashyap and Brian Ng and Jamie Hellman and Nikesh Kotecha and Christopher Sharp and Gretchen Brown and Christian Lindmark and Anurang Revri and Michael A. Pfeffer},
  year          = {2026},
  eprint        = {2602.00074},
  archivePrefix = {arXiv},
  primaryClass  = {cs.CY},
  doi           = {10.48550/arXiv.2602.00074},
  url           = {https://arxiv.org/abs/2602.00074}
}

@article{Carrell2013Hiding,
  author  = {Carrell, David and Malin, Bradley and Aberdeen, John and Bayer, Samuel and Clark, Cheryl and Wellner, Ben and Hirschman, Lynette},
  title   = {Hiding in plain sight: use of realistic surrogates to reduce exposure of protected health information in clinical text},
  journal = {Journal of the American Medical Informatics Association},
  year    = {2013},
  volume  = {20},
  number  = {2},
  pages   = {342--348},
  doi     = {10.1136/amiajnl-2012-001034},
  pmid    = {22771529},
  pmcid   = {PMC3638183}
}

@online{openai2026healthcare,
  author  = {{OpenAI}},
  title   = {Healthcare Organizations Can Now Connect {EHR} and Additional Industry Data to {ChatGPT}},
  year    = {2026},
  month   = sep,
  day     = {1},
  url     = {https://openai.com/index/chatgpt-connects-health-records-and-healthcare-sources/},
  urldate = {2026-09-02}
}

@article{rajpurkar2025clinical,
  author  = {Rajpurkar, Pranav and Topol, Eric J.},
  title   = {A clinical certification pathway for generalist medical {AI} systems},
  journal = {The Lancet},
  year    = {2025},
  volume  = {405},
  number  = {10472},
  pages   = {20},
  doi     = {10.1016/S0140-6736(24)02797-1},
  url     = {https://doi.org/10.1016/S0140-6736(24)02797-1}
}

@article{bressman2026software,
  author  = {Bressman, Eric and Shachar, Carmel and Stern, Ariel D. and Mehrotra, Ateev},
  title   = {Software as a Medical Practitioner---Is It Time to License Artificial Intelligence?},
  journal = {JAMA Internal Medicine},
  year    = {2026},
  volume  = {186},
  number  = {1},
  pages   = {5--6},
  doi     = {10.1001/jamainternmed.2025.6132},
  url     = {https://doi.org/10.1001/jamainternmed.2025.6132}
}

@article{rteb2025,
  author = {Liu, Frank and Enevoldsen, Kenneth and Solomatin, Roman and Chung, Isaac and Aarsen, Tom and Fődi, Zoltán},
  title = {Introducing RTEB: A New Standard for Retrieval Evaluation},
  year = {2025},
}
